\documentclass{article} %
\usepackage{iclr2025_conference,times}

\usepackage{amsmath,amsfonts,bm}

\def\eqref#1{equation~\ref{#1}}

\def\1{\bm{1}}

\DeclareMathAlphabet{\mathsfit}{\encodingdefault}{\sfdefault}{m}{sl}
\SetMathAlphabet{\mathsfit}{bold}{\encodingdefault}{\sfdefault}{bx}{n}

\usepackage{hyperref}
\usepackage{url}
\usepackage{todonotes}

\usepackage{placeins}
\usepackage{xspace}
\usepackage[capitalize,noabbrev]{cleveref}

\usepackage[hide=true,setmargin=true,marginparwidth=1in]{marginalia}

\usepackage{inputenc}[utf8]
\usepackage{booktabs} %
\usepackage{graphicx}
\usepackage{subcaption}
\usepackage{float}
\graphicspath{{../}}

\title{Mechanistic Unlearning: \\
Robust Knowledge Unlearning and Editing \\via Mechanistic Localization}

\author{Phillip Guo$^1$\thanks{Equal contribution, determined by coin flip.}\; \quad Aaquib Syed$^{1*}$ \quad Abhay Sheshadri$^2$ \quad Aidan Ewart$^3$ \quad Gintare Karolina Dziugaite$^4$\\
$^1$University of Maryland $^2$Georgia Institute of Technology $^3$University
of Bristol\\
$^4$Google DeepMind \\
\texttt{\{phguo,asyed04\}@umd.edu}\\
\texttt{gkdz@google.com}
}

\newcommand{\LT}{FLU\xspace}
\newcommand{\OT}{OT\xspace}
\newcommand{\Sportunlearn}{Sports-Unlearning\xspace}
\newcommand{\sportunlearn}{sports-unlearning\xspace}
\newcommand{\sportathleteedit}{sports-athlete-editing\xspace}
\newcommand{\Sportathleteedit}{Sports-Athlete-Editing\xspace}
\newcommand{\sportedit}{full-sports-editing\xspace}
\newcommand{\Sportedit}{Full-Sports-Editing\xspace}
\newcommand{\counterfactedit}{counterfact-editing\xspace}
\newcommand{\CounterFactedit}{CounterFact-Editing\xspace}
\newcommand{\counterfactsequentialedit}{sequential-counterfact-editing\xspace}
\newcommand{\CounterFactsequentialedit}{Sequential-CounterFact-Editing\xspace}

\iclrfinalcopy %
\begin{document}

\maketitle

\begin{abstract}
Methods for knowledge editing and unlearning in large language models seek to edit or remove undesirable knowledge or capabilities without compromising general language modeling performance. This work investigates how mechanistic interpretability---which, in part, aims to identify model components (circuits) associated to specific interpretable mechanisms that make up a model capability---can improve the precision and effectiveness of editing and unlearning. 
We find a stark difference in unlearning and edit robustness when training components localized by different methods. We highlight an important distinction between methods that localize components based primarily on preserving outputs, and those finding high level mechanisms with predictable intermediate states.
In particular, localizing edits/unlearning to components associated with the \textit{lookup-table mechanism} for factual recall 1) leads to more robust edits/unlearning across different input/output formats, and 2) resists attempts to relearn the unwanted information, while also reducing unintended side effects compared to baselines, on both a sports facts dataset and the CounterFact dataset across multiple models.
We also find that certain localized edits disrupt the latent knowledge in the model more than any other baselines, making unlearning more robust to various attacks.
\end{abstract}

\section{Introduction}

Large language models (LLMs) often learn to encode undesirable knowledge. The possibility of selectively editing or unlearning this type of knowledge is viewed as paramount for ensuring accuracy, fairness, and control of AI. Yet, editing and unlearning of knowledge from these models remains challenging.

Common editing and unlearning methods often come at the cost of affecting other general or tangential knowledge or capabilities within the model. Moreover, the edits achieved through these methods may not be robust -- e.g., slight variations in the prompt formulation can often still elicit the original fact or capability, or the original answers are still present/extractable given white-box access. 

Some recent work has explored editing or unlearning techniques that rely on mechanistic interpretability methods attempting to trace which components of a network store specific facts \citep{meng2023locating}. These methods, such as causal tracing or attribution patching, focus on measuring how output or task accuracy is affected when clean/corrupted input is patched into specific components. 

We coin a new term to categorize localizations which measure causal effects of components on only the output: \textit{Output-Tracing} localizations. The effectiveness of output-tracing (\OT) techniques like Causal Tracing for editing has been questioned by \citet{hase2023does}. Our research confirms these doubts, finding that localized editing and unlearning of facts based on several existing OT methods often perform equal to or worse than simply updating the entire model. This is particularly evident when evaluating the robustness of edits against prompt variations and relearning, and when probing for remaining latent knowledge. %

Another style of interpretability techniques first breaks down computations into high-level mechanisms with predictable intermediate states. Based on such work by \citet{nanda2023factfinding,geva2023dissecting}, we link certain MLP layers to a fact lookup (\LT) mechanism for facts used in our analysis, that enrich the latent stream with subject attributes but don't directly write to the output.
For unlearning and edits of these facts, we only modify components that implement the \LT mechanism.
More broadly, we refer to editing and unlearning that acts on components of the model identified by mechanistic intermediate component analysis as \emph{mechanistic unlearning}.
We demonstrate that \LT \emph{mechanistic unlearning} leads to better trade-offs between edits/unlearning and maintaining performance on general language modelling capabilities, compared to edits done using \OT or without any localization. 
Further, it exhibits improved robustness to re-learning and alternative prompting, and we demonstrate that the latent knowledge is also perturbed.

\ \\ 

\vspace{-20pt}
\paragraph{Summary of Contributions}

\begin{itemize}

    \item  We perform a rigorous evaluation of several standard editing approaches on factual recall tasks, and we identify mechanisms for factual lookup and attribute extraction on Gemma-7B, Gemma-2-9B, and Llama-3-8b. We demonstrate that gradient-based editing localized on the factual lookup mechanism is more robust than \OT localizations and baselines across multiple datasets, models, and evaluations.

    \item We demonstrate that it is more difficult to elicit the  forgotten ground truth answers using alternative prompting with \LT localizations. We also demonstrate slower or no relearning of the ground truth answers, retraining edited models on half of the edited set and evaluating them on the other half of the edited set. %

    \item We analyze intermediate representations using probing, and provide further evidence that editing with \LT localization modifies the internal latent information to reflect the desired edited answer more than other localizations and baselines. We also analyze the weights that are modified for each localization, and find that \OT techniques and baselines modify the attribute extraction mechanisms more than the fact lookup mechanism.
    
    \item
    We show that editing and unlearning localized on these mechanisms is more parameter efficient, by controlling for the sizes of edits made to the model with weight masking.
    
\end{itemize}

\begin{figure}[t] \centering
\includegraphics[width=.85\textwidth]{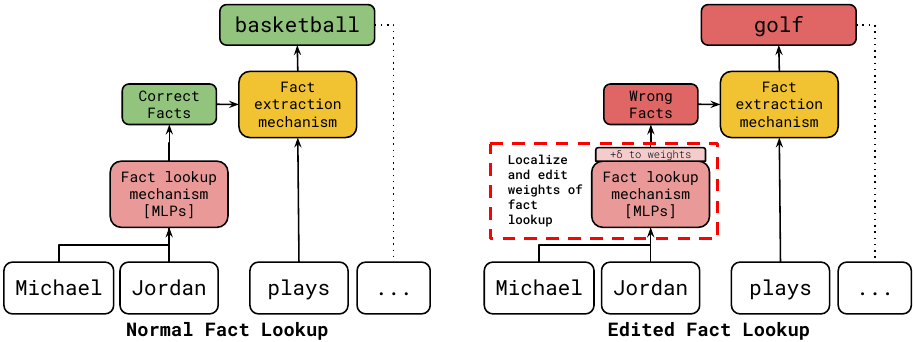}
\caption{High level depiction of \textit{mechanistic unlearning}. We localize components responsible for fact extraction/enrichment and modify their weights to change the associations, in order to target internal latent representations rather than targeting the output. Graph inspired by \citet{nanda2023factfinding}.}
\label{fig:high-level}
\end{figure}
\newpage
\subsection{Related Work}
\textbf{Mechanistic Interpretability} is a subfield of AI interpretability, aiming to understand the internal processes of AI models by attributing them to subnetworks (called circuits) within the model \citep{olah2020zoom}. We focus on the factual recall interpretability literature \citep{nanda2023factfinding, geva2023dissecting, chughtai2024summing, yu2023characterizing}, which studies methods that aim to discover mechanisms for the retrieval and formatted extraction of factual information. %

\textbf{Output tracing} methods aim to automatically find causally important subnetworks of components for a task. Causal Tracing \citep{meng2023locating} and Automated circuit discovery (ACDC) \citep{conmy2023automated} utilize repeated activation patching to attempt to find the subnetworks that are most critical for the model's output on that task. Efficient methods such as attribution patching \citep{neelattribution} and edge attribution patching \citep{syed2023attribution} are linear approximations of activation patching for discovering important components quickly.

\vspace{-10pt}
\paragraph{Fact Editing and Machine Unlearning} seek to modify pre-trained models to eliminate or alter learned knowledge such as capabilities or facts.  
Some prior approaches focus on identifying and removing specific individual training data points, aiming to obtain a model that is ``similar'' to one that had never trained on these data points \citep{cao2015towards,xu2023machine}.
One formalization of unlearning to match a retrained-from-scratch model is due to \citet{ginart2019making}, and is closely inspired by differential privacy \citep{dwork2014algorithmic}.
Fact editing focuses on overwriting factual information while preserving overall language generation ability. \citet{meng2023locating} attempts to identify MLP modules that are most responsible for factual predictions via Causal Tracing and then applies a rank-one transformation upon these modules to replace factual associations. %

In the context of LLMs and safety, techniques such as Helpful-Harmless RLHF \citep{bai2022training} and Representation Misdirection for Unlearning \citep{li2024wmdp} aim to suppress dangerous knowledge or harmful tendencies in LLMs. \citet{li2024circuit,zou2023representationengineeringtopdownapproach,zou2024improvingalignmentrobustnesscircuit} approach unlearning and dangerous knowledge suppression from a top-down feature view, reading or suppressing linear features related to memorized, harmful, and undesired concepts. 
A related line of work on safety proposes methods making it difficult to modify open models for use on harmful domains \citep{tamirisa2024tamperresistantsafeguardsopenweightllms,deng2024sophon,henderson2023self}, including through adversarial relearning. 

\paragraph{Failures of Unlearning and Editing} have been shown for both localized and nonlocalized methods. \citet{patil2023sensitive} extract correct answers to edited facts from the intermediate residual stream and through prompt rephrasing. 
\citet{yong2024lowresource} show that low-resource languages jailbreak models output unsafe content, and \citet{lo2024large, lermen2023lora, deeb2024unlearningmethodsremoveinformation} demonstrate that relearning with a small amount of compute/data causes models to regain undesirable knowledge/tendencies.
Even without explicit finetuning, \citet{xhonneux2024context} show that in-context learning alone suffices to reintroduce undesirable knowledge despite the model being designed to refuse to output such knowledge.
\citet{lee2024mechanistic} shows that even after alignment techniques are applied to make models nontoxic, toxicity representations are still present, just not triggered - they argue that this is a reason that models lack robustness and can still be jailbroken to trigger this unwanted behavior.

\citet{hong2024intrinsicevaluationunlearningusing} evaluates unlearning by measuring residual knowledge left in internal activations, and demonstrate that current approaches fail to remove this residual knowledge and thus can be exploited. They attempt unlearning by targeting the MLPs these residual knowledge traces reside in, but fail to find a non-oracle unlearning approach that successfully removes residual information.

\newpage
\section{Methods}
Our experiments are designed to test the effectiveness of localization for editing of facts.
In this section we describe the tasks used and the localization and editing methods evaluated.

\subsection{Editing Tasks}
\label{sec:tasks}
We focus on editing subsets of two datasets: (1) Sports Facts dataset from \citet{nanda2023factfinding}, which contains subject-sport relations across three sports categories for 1567 athletes, and (2) the CounterFact dataset from \citet{meng2023locating}.

\paragraph{Sports Facts: \Sportathleteedit, \Sportedit, and \Sportunlearn tasks} In the Sports Facts dataset, we edit two general groups of factual associations. 
For the first editing task, we edit factual associations  for a constant set of randomly selected athletes belonging to any of the three sport categories. 
We test editing these sets of associations by replacing their correct sports with one of the other two incorrect sports (with equal probability). 
To increase the comprehensiveness of our evaluation, we run experiments with different \emph{forget set} sizes: 16 athletes and 64 athletes. We refer to this task as \textbf{\Sportathleteedit}. These chosen forget sets are constant between all localizations.
For the second editing task, we unlearn all athlete-sport associations for athletes who play one sport. In this case, we establish a forget set consisting of all the athletes who play one sport (basketball, baseball, or football), and we edit the association by replacing the athlete's correct sport with golf. 
For comprehensiveness, we vary the sport that the forget set is constructed from. We refer to this task as \textbf{\Sportedit}.
Finally, we also design an unlearning task, \textbf{\Sportunlearn}, where the goal is to unlearn factual associations for all athletes in one of the sports. 

\paragraph{\CounterFactedit and \CounterFactsequentialedit task} In the CounterFact dataset, following \citet{geva2023dissecting}, we first filter the dataset for facts which our models assign higher than 50\% probability to the right answer, which varies per model. 
The goal of our \textbf{\CounterFactedit} task is to edit a constant set of facts, replacing the correct answers with an alternative false target, with the retain set being the rest of the non-forget facts. We vary forget set sizes to be of 16 and 64 facts. In \textbf{\CounterFactsequentialedit} task, we edit a total of 64 facts by sequentially editing four randomly selected subsets of 16 facts. We test sequential editing here because facts from CounterFact can reside in different parts of the model, so we wish to test if we can exploit different localizations for different facts. These chosen forget sets are constant between all localizations.

\paragraph{Models} We implement editing on the Gemma-7B LLM, the Gemma-2-9B LLM, and the Llama-3-8b LLM. We don't use the Pythia-2.8B \citep{biderman2023pythiasuiteanalyzinglarge} and GPT-2 models tested in the previous fact interpretability literature because our larger models have stronger general capabilities which we can measure for side effects, and also because our larger models can provide factual knowledge in more input/output formats for more robustness evaluations. All graphs presented in the main text and appendix, unless otherwise specified, are averaged over all three model types.

\subsection{Localization Methods and Baselines}
\label{sec:localization}

Given a model $M:X \mapsto L$ mapping sequence of tokens $X$ to logits $L \in \mathbb{R}^{|V|}$ over vocabulary $V$, we consider $M$ to be a directed acyclic graph $(C, E)$ with $C$ being a set of model components and $E$ being edges between components. Adopting notation from \citet{elhage2021mathematical}, we consider the query, key, value, and output weights of each head along with the input, gate, and output projection weights of each MLP as components. %

We are interested in finding $S:C\xrightarrow{} \mathbb{R}$, a mapping of components to their importance in a given task. A localization is a set of components $C_{\tau}:=\{c : c \in C, |S(c)| > \tau\}$, where $\tau$ is a threshold. In practice, we fix $\tau$ such that $C_{\tau}$ contains the same number of parameters in \OT, \LT, and random localizations. We use these efficient localization methods for finding these mappings: 
\paragraph{Output Tracing (\OT) localization: Causal Tracing and Attribution Patching} First, we test Causal Tracing, a method for finding components with high direct causal importance for factual associations \citep{meng2023locating}.
We also use Attribution Patching \citep{neelattribution} as a fast and acceptably accurate approximation of causal tracing to automatically localize over components with high direct and indirect importance. We additionally consider the versions of these localizations with only MLPs (\textit{Causal Tracing MLPs} and \textit{Attribution Patching MLPs}). %

We hypothesize that these output-based techniques will prioritize the shared extraction components and other mechanisms for reformatting predictions over the more diffuse \LT components, and thus appear more precise yet leave the underlying latent information present in the model. This might decrease robustness under alternative extraction methods, thus motivating non-\OT-based localization, described next. We discuss the precise components/mechanisms highlighted by \OT localizations in \cref{app:ot_components}.
\paragraph{Fact Lookup (\LT) localization:} Next, we use manually derived localizations for MLP layers. 
For Sports Facts, our localization is inspired by \citet{nanda2023factfinding}, who identified components in Pythia 2.8B responsible for \emph{token concatenation, fact lookup}, and \emph{attribute extraction}. Their work, along with \citet{geva2023dissecting}, demonstrates that the fact-lookup stage enriches the latent representation of the subject (athlete) with information about their corresponding sport, while the attribute extraction stage extracts the latent sport information and formats it in the final token position. We replicate a key result of their work in our three models by training logistic regression models (probes) to predict the correct sport using the internal activations of the model taken from different layers. We consider the FLU localization to be the layers at which the probe accuracies rapidly increase as these correspond to layers where representations of the athletes are being enriched to encode the correct sport.

For CounterFact, we cannot use the probe technique since unlike Sports Facts, the correct answers for the dataset do not fall into a few categories that we can train a probe for. Instead, we first use path patching from \citet{goldowskydill2023localizingmodelbehaviorpath} to measure the direct importance of attention heads and MLPs on the final logit difference between the correct answer and a sample incorrect answer. Path patching is a technique that finds the effect of corrupting a single path from a sender component to a receiver component on the final logits of the model. Components that change the output significantly without being mediated by other components do so by directly affecting the logits, and thus we consider them to be responsible for extracting the facts encoded in the representation into the answer logits. We thus refer to these components as the “attribute extraction mechanism” \citet{geva2023dissecting}. We use path patching again, this time patching paths between MLPs and the attribute extraction mechanism to find the components with the highest contribution to the logit difference as mediated through the extraction mechanisms. Such components enrich the token representations with the appropriate facts to then be extracted, and thus we use them as our \LT localization. More details about the manual analysis for both datasets is outlined in \cref{app:gemma_mi}.

Importantly, \LT differs from \OT techniques by focusing on the causal effects of ablations upon intermediate representations used by the factual recall mechanism, not just the effects on the
output. We hypothesize that the optimal location for robust editing is in the fact lookup stage rather
than in the attribute extraction stage,as adversaries could potentially devise alternative extraction methods if knowledge remains in the latent stream. Therefore, our model edits are focused exclusively on the fact lookup MLPs.
\paragraph{Baselines: Random-MLPs, Random, All-MLPs, and Nonlocalized} We additionally consider four baselines: one corresponding to $C_{\tau} = C$ (i.e., no localization, optimizing all the components of the model), another that randomly chooses components, another that trains all MLP components, and another trains a random selection of MLP components. We test the last MLP baselines to determine if our mechanistically localized MLPs are uniquely important - we want to know if the same unlearning performance can be achieved with just the heuristic that training only MLPs improves robustness, or if mechanistic understanding of the role of the component is crucial. 

In \cref{sec:mechanism_modification_analysis}, we analyze the proportions of each mechanism (the extraction heads, extraction MLPs, and fact lookup MLPs, by parameter count) that are present in each localization. 

The main text focuses on comparing \LT, Causal Tracing, and Nonlocalized, while the appendix has the same figures with all above localizations included, with the same conclusions in every case.

\subsection{Parameter Update Method}
\label{sec:unlearningalgs}

Once we have a localization $C_{\tau}$, we run one of the unlearning or editing methods, restricting weight updates to only components in $C_{\tau}$. We update weights using gradient descent on a combination of loss functions.

\paragraph{Localized Fine-Tuning} Following work by \citet{lee2023surgical} and \citet{panigrahi2023taskspecific}, we fine-tune the parameters within the localized components. 
For editing, we use a loss function $L = \lambda_1 L_{\text{injection}} + \lambda_2 L_{\text{retain}} + \lambda_3 L_{\text{SFT}}$, where $L_{\text{injection}}$ is a cross-entropy loss on the forget facts maximizing the probability of the alternative new false target. $L_{\text{retain}}$ is a cross-entropy loss on a train split of the remaining facts, and $L_{\text{SFT}}$ is a cross-entropy loss on the Pile dataset \citep{gao2020pile}. We sweep over learning rates and injection loss $\lambda$s for three representative localizations in \cref{app:hyperparams} .

\section{Editing Evaluation}

In this section, we show the results of model editing with localization: we test localization techniques from \cref{sec:localization}, and edit these localized components using fine-tuning (\cref{sec:unlearningalgs}). 
Here we focus on four main editing tasks: \textbf{\sportathleteedit}, \textbf{\sportedit}, \textbf{\counterfactedit}, and \textbf{\counterfactsequentialedit}. We present augmenting results for \textbf{\sportunlearn} in \cref{app:sports-unlearning-results}.

All the editing tasks are assessed based on prompt-completion based and adversarial relearning evaluations.

\subsection{Prompting-based Evaluation}\label{sec:prompting}

Our prompting-based evaluation assesses an editing method's ability to forget or edit specific information while retaining unrelated knowledge.
This is measured by evaluating how the model post-editing completes the prompts coming from the forget set: we report how accurately it recalls the undesired forgotten answer (\textit{forget accuracy}), and how accurately it recalls the new desired edited answer (\textit{edit accuracy}). In addition, we also measure the accuracy on facts not in the forget set (\textit{maintain accuracy}). In cases where a positive result is lower accuracy, we use the term \textit{error} to denote 1 - accuracy (e.g. \textit{forget error} = 1 - \textit{forget accuracy}). Thus, well edited models should decrease forget accuracy, increase forget error, and increase edit accuracy. Results of these standard evaluations are reported in \cref{app:standard_prompting}.

Inspired by \citet{patil2023sensitive} and \citet{lynch2024methods}, to ensure the editing process has not overfit to the specific format of the original prompts, we incorporate a robustness check using a multiple-choice question format (MCQ accuracy).  This helps determine to what extent the model edited the information, and whether it can still access and utilize that knowledge when prompted differently.
In this MCQ evaluation, the prompt also includes some in-context examples of answering multiple choice questions correctly.
On the forget set, we refer to the accuracy of the model answering with the ground truth as the \textit{MCQ Forget Accuracy} (stronger methods should decrease \textit{MCQ Forget Accuracy}), and the accuracy of the model answering with the new edited answer's choice as the \textit{MCQ Edit Accuracy} (stronger methods should increase \textit{MCQ Edit Accuracy}).

Finally, we also evaluate the models' accuracy post-editing on MMLU \citep{hendrycks2021measuring} as a proxy for general language understanding to measure any unintended side effects.

\subsubsection{Sports Tasks}
\paragraph{Sports Prompting}
For the sports dataset, following \citet{nanda2023factfinding}, we first evaluate the accuracy of our models to complete the prompt, ``Fact: [athlete] plays the sport of'', with a one-shot example of Tiger Woods playing golf given first. Note that this is the same prompt used for the editing loss in the first place. %
For the MCQ evaluation, we use choices of all four sports (football, baseball, basketball, and golf).

We average accuracies over all models, for \Sportathleteedit we average over editing both 16 and 64 facts, and for \Sportedit we average over editing Basketball, Baseball, and Football.

\paragraph{Sports Results}
\begin{figure}
    \centering
    \begin{subfigure}[b]{0.4\textwidth}
        \centering
        \includegraphics[height=4.5cm,keepaspectratio]{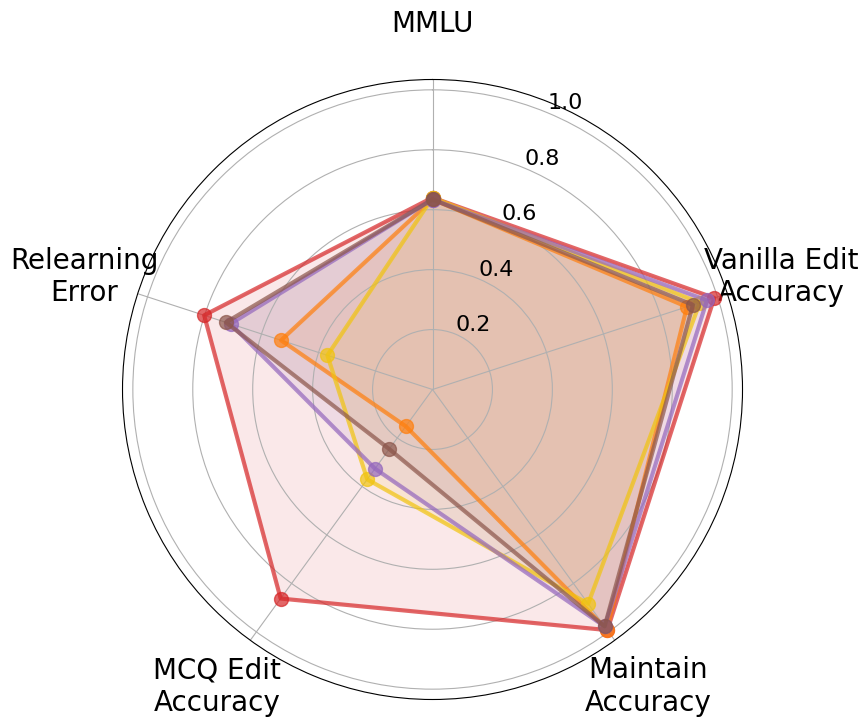}
        \caption{\Sportathleteedit}
        \label{fig:sportathlete}
    \end{subfigure}
    \hfill
    \begin{subfigure}[b]{0.58\textwidth}
        \centering
        \includegraphics[height=4.5cm,keepaspectratio]{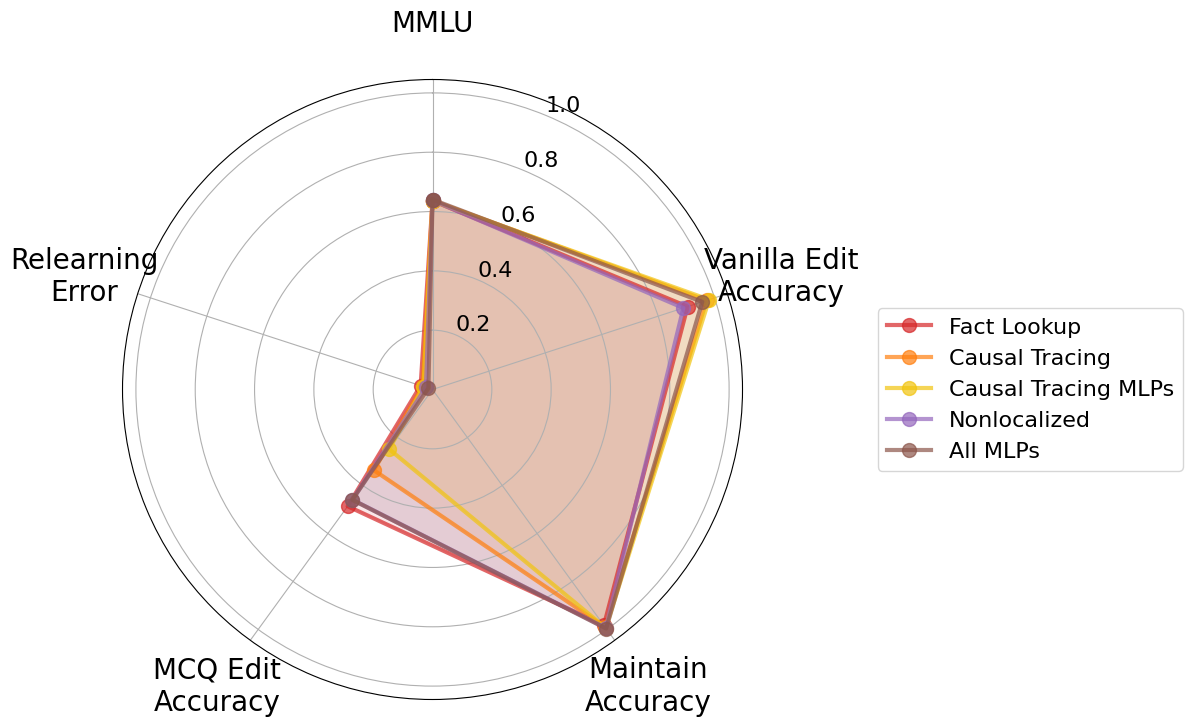}
        \caption{\Sportedit}
        \label{fig:sportedit}
    \end{subfigure}
\caption{Spider plots illustrating the advantages of \LT for editing Sports across adversarial prompting and relearning evaluations, averaged over all three model types. \textbf{(Left)} The \Sportathleteedit plot shows that \LT localization leads to editing that is the most robust against MCQ prompting and relearning. \textbf{(Right)} The plot shows that most localizations perform approximately equivalently in the \Sportedit task, with \LT localization slightly better for MCQ.}
    \label{fig:sport_spiderplots}
\end{figure}

\begin{figure}
    \centering
    \begin{subfigure}[b]{0.4\textwidth}
        \centering
        \includegraphics[height=4.75cm,keepaspectratio]{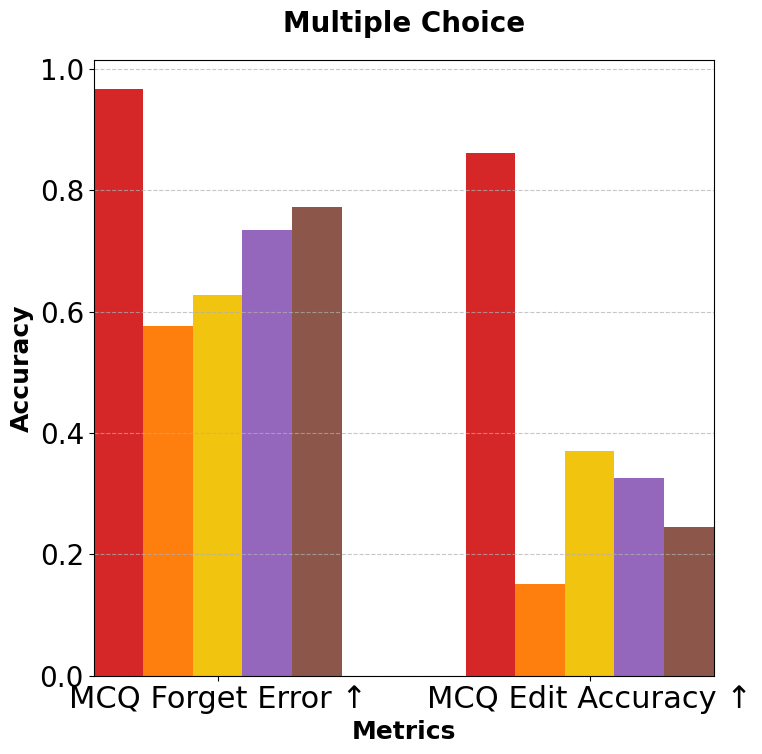}
        \caption{\Sportathleteedit}
        \label{fig:sportathlete_bar}
    \end{subfigure}
    \hfill
    \begin{subfigure}[b]{0.58\textwidth}
        \centering
        \includegraphics[height=4.75cm,keepaspectratio]{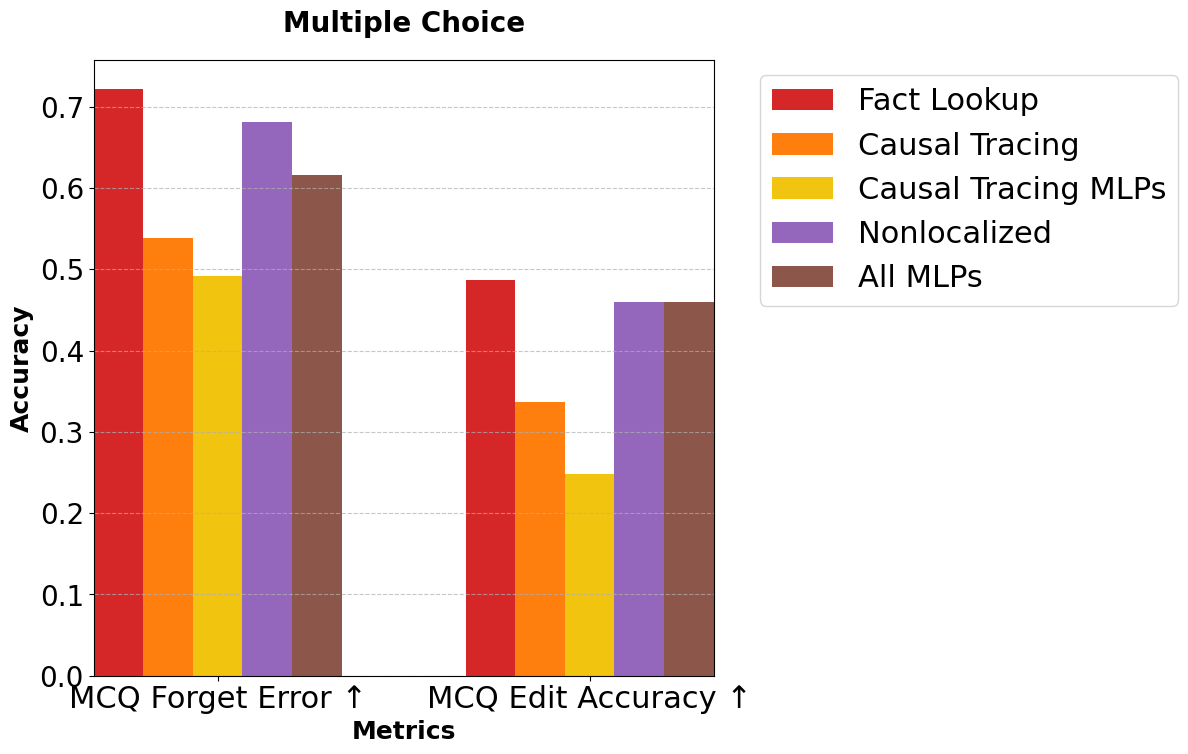}
        \caption{\Sportedit}
        \label{fig:sportedit_bar}
    \end{subfigure}
\caption{Bar charts showing results of MCQ evaluations, reporting both the forget error and edit accuracy when prompted with MCQ, averaged over three model types. For both \textbf{(a)} \Sportathleteedit and \textbf{(b)} \Sportedit, \LT localization answers with the original answer the least (MCQ Forget Error) and answers with the edited answer most accurately (MCQ Edit Accuracy).}
    \label{fig:sport_barcharts}
\end{figure}

Our analysis reveals that editing employing \LT localization exhibits superior performance in forgetting the original information and adopting the edited information, across different prompt formats.
As explained in \cref{sec:prompting}, better editing should result in higher \textit{MCQ Forget Error}, and higher \textit{MCQ Edit Accuracy}.
\Cref{fig:sport_barcharts} shows that \LT localized models are the best on both fronts. The difference is especially significant in \cref{fig:sport_barcharts}a, where only \LT model edits generalize meaningfully to MCQ, exceeding other localization methods by more than 40\% in MCQ Edit Accuracy.

A comprehensive comparison of all localization methods with multiple-choice prompting is available in \cref{app:adv_prompting}, further supporting our findings.

\subsubsection{CounterFact Tasks}

\paragraph{CounterFact Prompting}
For CounterFact, we create an MCQ evaluation with four choices for every question, randomly ordering the true answer, the injected false answer, and two other question-specific LLM-generated incorrect answers. 
We also consider the original robustness and side effect evaluations from the \citet{meng2023locating} dataset: the Paraphrase and Neighborhood facts accuracies. Answers of edited facts are meant to generalize to the Paraphrase evaluation, which phrases the fact in a different but equivalent way, so we report the \textit{Paraphrase Edit Accuracy} (stronger methods should increase Paraphrase edit accuracy). Editing should not generalize to the Neighborhood evaluation, which presents similar but unrelated facts to the forget set, so we report the \textit{Neighborhood Edit Error} which is lower if models incorrectly report the edited answer in these unrelated facts.

We again also use an MCQ evaluation, where the choices consist of the true answer, the injected false answer, and two other question-specific LLM-generated incorrect answers. We note that Paraphrase and Neighborhood evaluations ask for the answer in the same original format, so they are more in-distribution than MCQ. We average accuracies over all models, and for \CounterFactedit we average over editing both 16 and 64 facts.

\begin{figure}[t]
    \centering
    \begin{subfigure}[b]{0.41\textwidth}
        \centering
        \includegraphics[height=4.25cm,keepaspectratio]{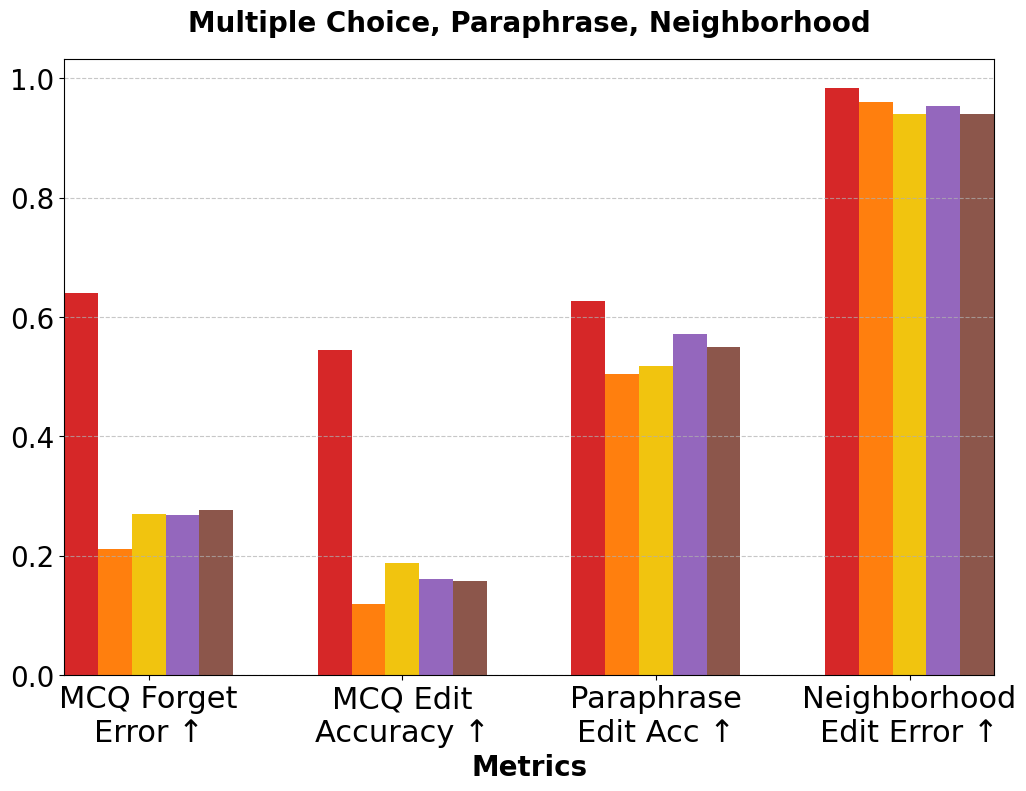}
        \caption{\CounterFactedit}
        \label{fig:counterfact_bar}
    \end{subfigure}
    \hfill
    \begin{subfigure}[b]{0.58\textwidth}
        \centering
        \includegraphics[height=4.25cm,keepaspectratio]{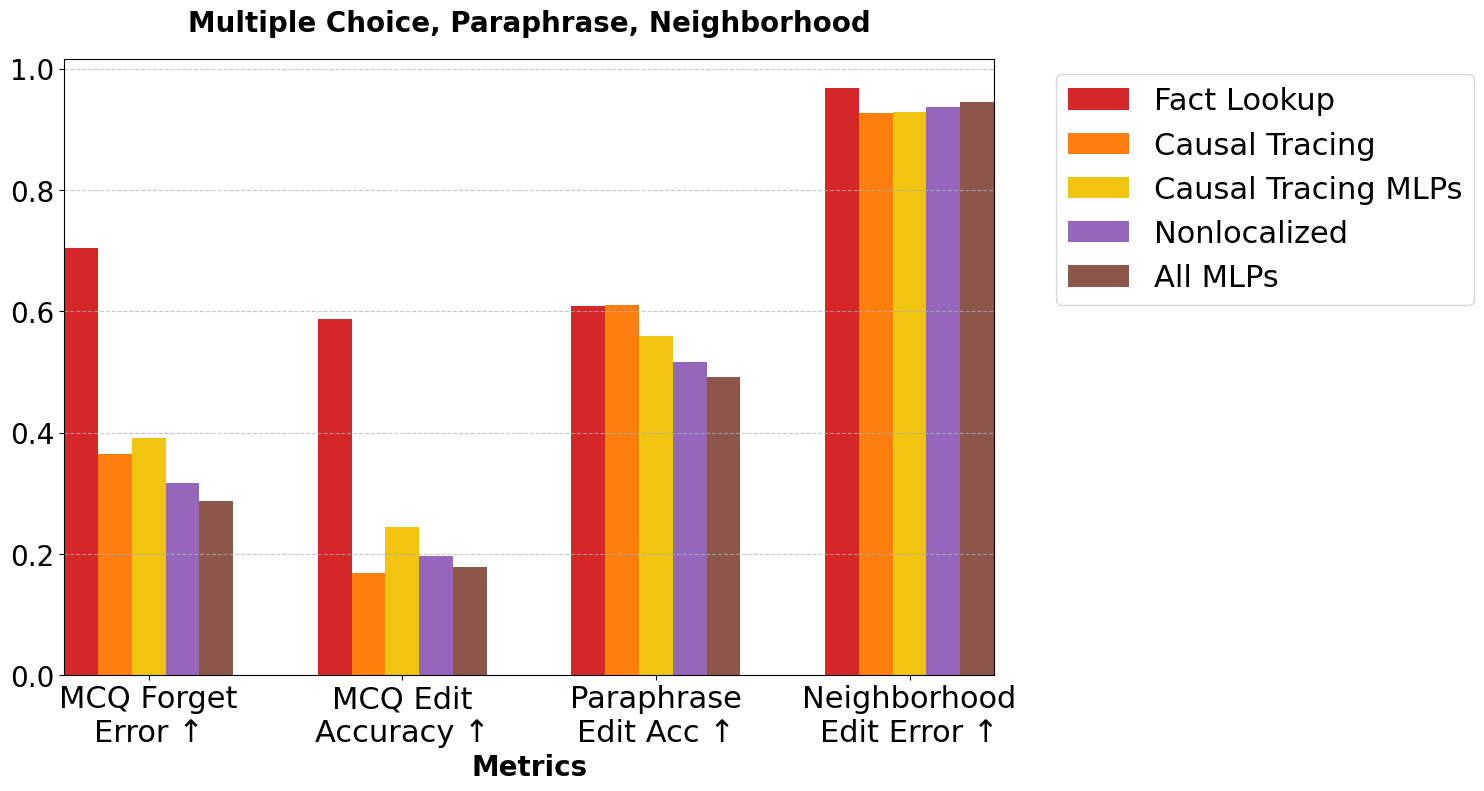}
        \caption{\CounterFactsequentialedit}
        \label{fig:counterfact_sequential_bar}
    \end{subfigure}
\caption{Bar charts showing results of MCQ, Paraphrase, and Neighborhood prompt evaluations, averaged over all three model types. For both \textbf{(a)} \CounterFactedit and \textbf{(b)} \CounterFactsequentialedit, \LT localization has the most robust edit accuracy measured by MCQ and Paraphrase. \LT localization editing also does not incorrectly generalize to Neighborhood prompts. Sequential editing is slightly more robust than nonsequential editing in MCQ when comparing between \CounterFactedit and \CounterFactsequentialedit bars.}
    \label{fig:counterfact_barcharts}
\end{figure}

\paragraph{CounterFact Results}
\Cref{fig:counterfact_barcharts} illustrates the robustness of \LT editing in the MCQ and Paraphrase evaluations.
Edited models using \LT localization answer less frequently with the original, incorrect information (\textit{MCQ Forget Error}) and more frequently provide the new, edited answer.
Furthermore, the \textit{Neighborhood Edit Error} highlights that other localization methods exhibit slightly more pronounced side effects,
inadvertently editing unintended, semantically similar facts.

Interestingly, sequential editing displays marginally greater robustness than nonsequential editing in MCQ when comparing between \CounterFactedit and \CounterFactsequentialedit bars. 
This observation supports an approach that editing large sets of facts can be made more effective by partitioning the set and applying edits sequentially.  
We present results comparing all localizations across each prompt robustness evaluation in \cref{app:adv_prompting}, with the largely consistent conclusions.

\begin{figure}[t]
    \centering
    \begin{subfigure}[b]{0.4\textwidth}
        \centering
        \includegraphics[height=4.25cm,keepaspectratio]{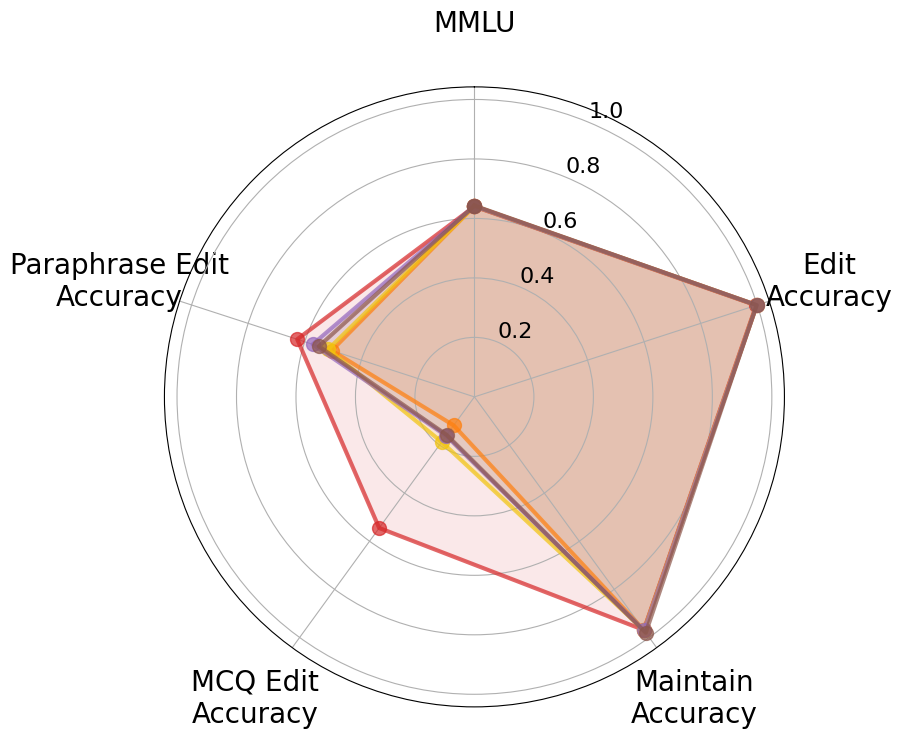}
        \caption{\CounterFactedit}
        \label{fig:counterfactedit}
    \end{subfigure}
    \hfill
    \begin{subfigure}[b]{0.58\textwidth}
        \centering
        \includegraphics[height=4.25cm,keepaspectratio]{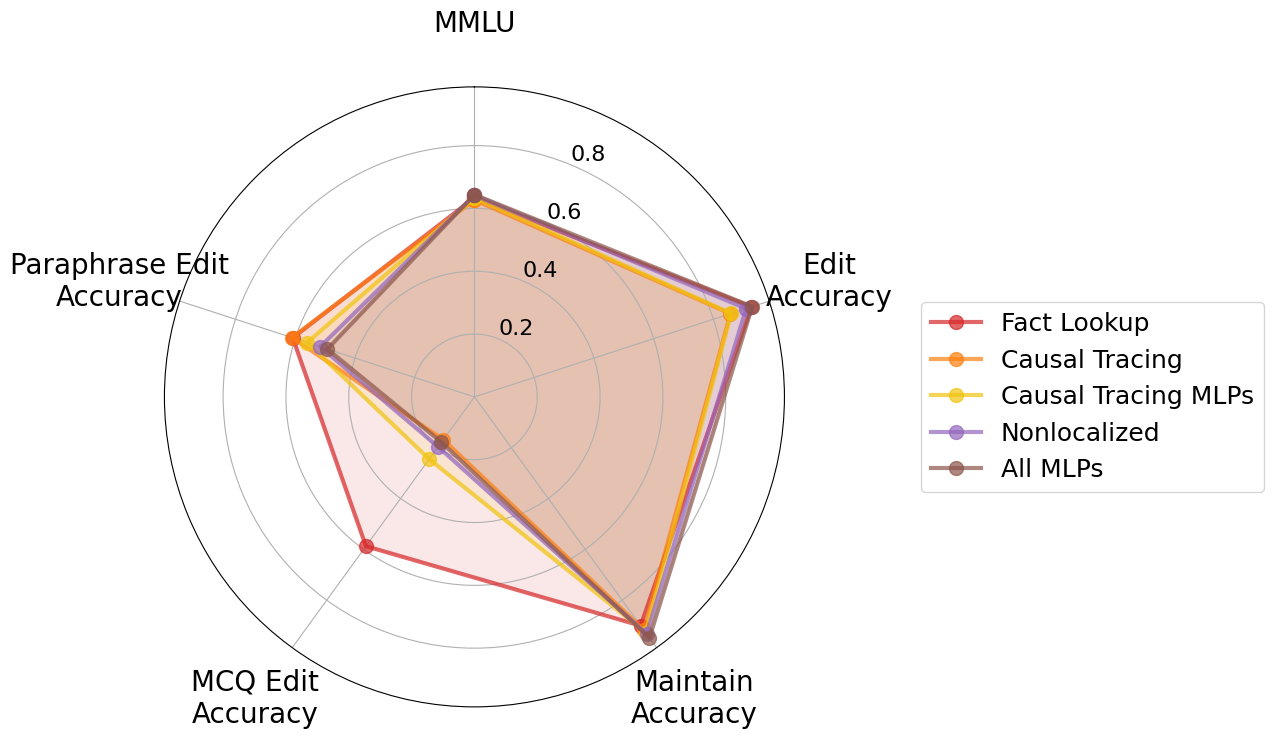}
        \caption{\CounterFactsequentialedit}
        \label{fig:sequentialedit}
    \end{subfigure}
    \caption{Spider plots illustrating the advantages of \LT for editing CounterFact across prompting evaluations, averaged over all three model types. \textbf{(Left)} The \CounterFactedit plot shows that \LT localization leads to editing that is the most robust against MCQ prompting and Paraphrasing. \textbf{(Right)} The \CounterFactsequentialedit plot shows that \LT localization is the most robust against MCQ prompting.}
    \label{fig:counterfact_spiderplots}
\end{figure}

\subsection{Adversarial Relearning Evaluation} \label{sec:adversarial_relearning}

We measure the ability of our models to withstand adversarial relearning, both to address the scenario in which adversaries may have fine-tuning access and as a measure for the quality of editing. We replicate the methodology of \citet{deeb2024unlearningmethodsremoveinformation}, splitting our forget sets in two independent halves, retraining with half of the ground truth labels, and evaluating on the other half. 
This methodology aims to discern whether the editing technique successfully removed the underlying factual association or merely obfuscated its direct retrieval while leaving it potentially susceptible to recovery when doing partial retraining.

We retrain with a rank-512 LoRA across all linear modules, with details available in \cref{app:advrelearn}. In this section, we focus on the \Sportathleteedit task, as in the other tasks it was either too easy to relearn (\Sportedit) or too hard to relearn any performance (all CounterFact tasks) across all localizations. Relearning isn't a valid evaluation for \Sportedit because the facts are not independent, and models should reasonably generalize from relearning on half the basketball athletes to answering with basketball on the other half of the basketball athletes. We show results from Counterfact and from all localizations in \cref{app:advrelearn}.

\paragraph{Sports Results}

\begin{figure}[h!]
\includegraphics[width=.98\linewidth]{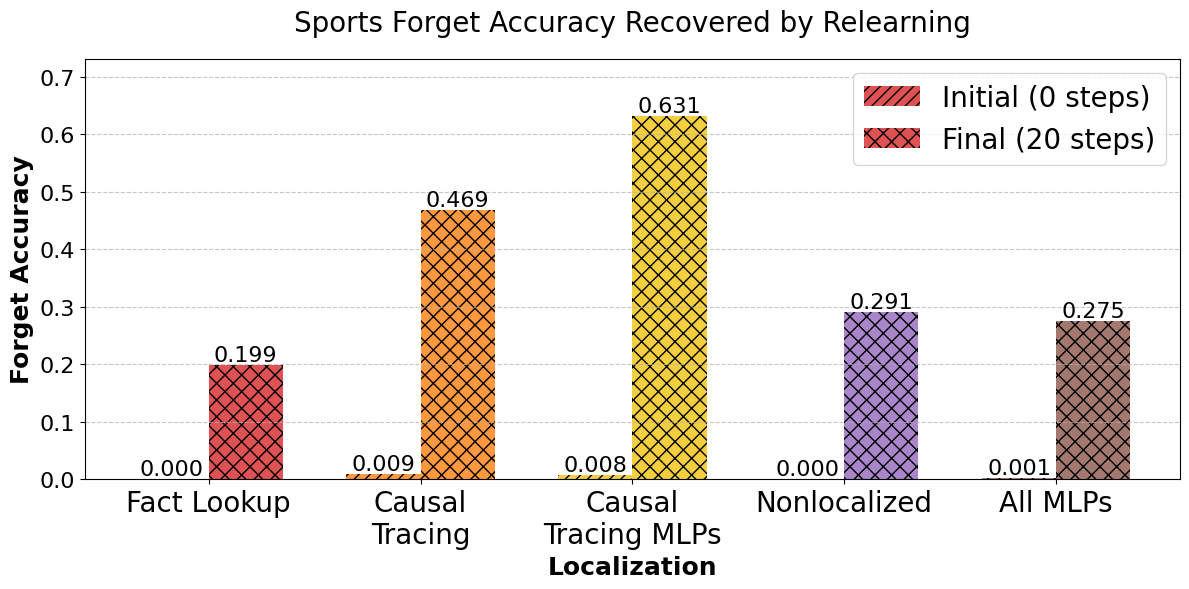}
\caption{Relearning recovers the least accuracy on the forget set using \LT localizations. Relearning recovers significant accuracy on the original forget set in \OT localizations (Causal Tracing and Causal Tracing MLPs).}
\label{fig:sports_relearning}
\end{figure}

Our adversarial relearning experiments, as depicted in \cref{fig:sports_relearning}, reveal that retraining on a subset of the original ``forgotten'' data can recover a significant portion, as much as 63\%, of the supposedly forgotten information when using \OT methods like Causal Tracing and Causal Tracing MLPs.  
This suggests that these methods may simply mask direct retrieval of this information, leaving the model susceptible to this information recovery through retraining.  
In contrast, \LT localization exhibits greater resilience to such adversarial relearning, with only about 20\% of the forgotten information recovered.  
This indicates that \LT localization may be more effective in targeting and removing the underlying knowledge, making it harder to recover through retraining.

\newpage
\subsection{Latent Knowledge Analysis} \label{sec:probing}
In this section, we provide more evidence of our hypothesis that \LT unlearning targets the source of intermediate latent knowledge. We analyze the \Sportathleteedit task again here because the ground truth and the edited answers vary between one of only three possibilities. 

We train logistic regression models (probes) \citep{alain2018understanding} on prompt activations following every model layer to predict the correct ground truth sport on the maintained set of athletes. This is possible because there are only three possible sports, so we can train binary classification probes for each sport and take the maximum classification over the sports. This is meant to discover internal representations of the true sport the model believes the answer to be: then, we apply these probes on the activations of the forget set of athletes.

We present graphs averaged over models and over 16 and 64 facts. Because layer is on the x-axis and models have different numbers of layers, averaging performance at a layer over multiple models isn't perfectly valid, so we also present probing graphs for each individual model and all localizations in \cref{app:latentknowledge}.

\paragraph{Sports Results}
\begin{figure}[h!] \centering
    \includegraphics[width=.98\linewidth]{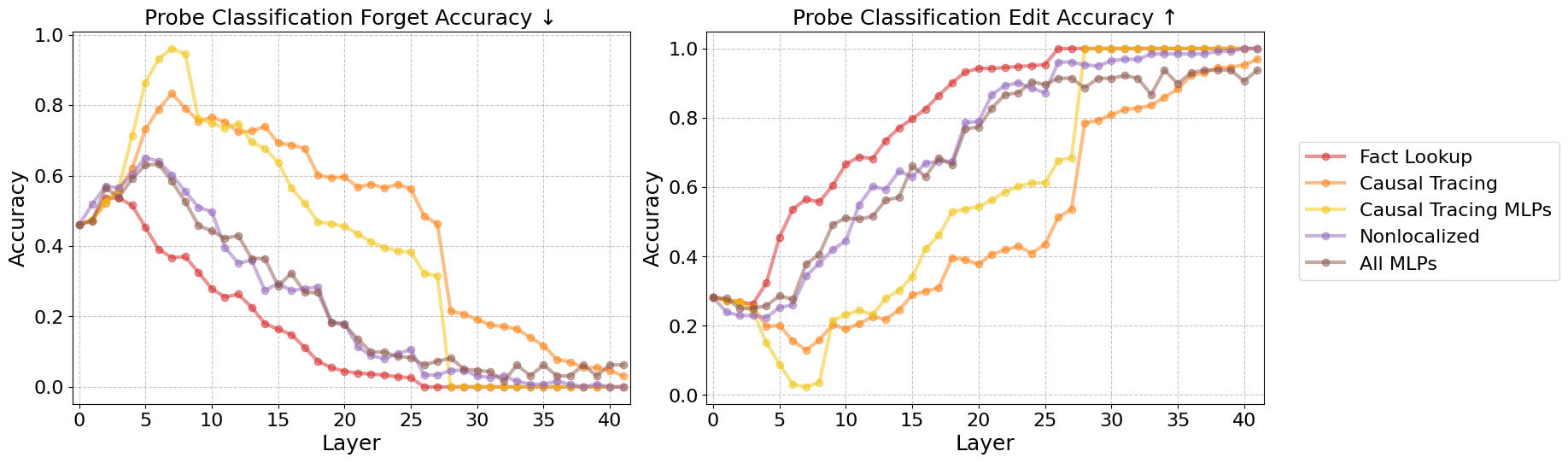}
    \caption{Linear probes applied to the forget set, classifying model activations after various layers. \textbf{(Left)} The line graph shows that some localizations still represent almost completely correct forget set knowledge in early layers, especially \OT, while \LT localizations represent this original knowledge the least. 
    \textbf{(Right)} The line graph shows that \LT localizations represent the edited rather than original answer earlier and more consistently throughout layers than any other localization.}
    \label{fig:ft-dlk-athletes-injection}
\end{figure}
In \cref{fig:ft-dlk-athletes-injection}, the probes on \LT consistently predict the forget answer less and the edit answer more than in any other localization, especially in early layers. Furthermore, the \LT probe classifications for the most part monotonically converge from their nonzero starting accuracy to 0 (for forget accuracy) and 1 (for edit accuracy). 

Every other localization has much higher peak probe classification forget accuracy in the early layers, especially the \OT localizations which have peak classification forget accuracy of almost 100\%. This strongly suggests that these models still significantly represent the ground truth answer rather than the edit answer in early layers.

\subsection{The Role of Parameter Count}\label{sec:param_count}
In this section, we perform weight-masking to quantify the size of edits with different localizations, as well as to investigate which factual mechanisms are targeted when editing with different localizations.
We employ a weight masking technique involving training a binary differentiable mask over individual weights of the model within the localized components, inspired by weight pruning/masking work \citep{bayazit2023discovering, panigrahi2023taskspecific}. In this case, no weight updates are being performed. Rather, the mask turns a subset of the weights to zero.

\paragraph{Controlling for Parameter Count} Although we already standardize the number of trainable parameters in most localizations, we additionally investigate if \LT editing is better than other localization techniques when controlling for the exact number of parameters that are masked.  

We perform weight masking on the \Sportunlearn, \Sportathleteedit, and \CounterFactedit tasks. Detailed results are reported in \cref{sec:weight_masking}. We find that when controlling for the size of the localization \LT is consistently more robust when subject to our suite of evaluations.

\paragraph{Other Localizations Affect the Extraction Mechanism}
After training weight masks, we analyze the proportion of each mechanism (fact lookup, attribution extraction) that is masked by each localization's weight mask in \cref{sec:mechanism_modification_analysis}. We demonstrate that \OT methods and nonlocalized editing all modify a higher proportion of the extraction head/MLP parameters than the fact lookup mechanism parameters, supporting our claim that OT methods target extraction mechanisms rather than the fact lookup mechanisms needed for robustness.

\section{Discussion} \label{sec:discussion}

Recent work by \citet{hase2023does} argued that localization is not useful for model editing. 
Our findings demonstrate that the relationship between localization and fact editing/unlearning is more nuanced, and reveals that not all localization techniques are equal. %

Our work evaluates the efficacy of different localization methods for modifying factual associations. We demonstrate clear benefits of localization for editing robustness through localized fine-tuning combined with \LT interpretability. 

In \cref{sec:probing,sec:param_count} and \cref{sec:mechanism_modification_analysis}, we provide evidence that \OT and baseline approaches fail to be robust because they target easily-localizable and high direct logit importance extraction components that transform existing latent factual knowledge to the desired output format. This can fail to generalize to different input and output formats and does not target the source of knowledge in the model: other input/output formats can allow alternative mechanisms to extract this knowledge, and relearning can repair the original mechanisms to recover accuracy. 
In contrast, \LT mechanistic understanding allows us to target editing at the sites where knowledge is sourced, which robustly prevents that information from entering the latent stream in any format. 

Our work also suggests unlearning/editing as a potential testbed for different interpretability methods, which might sidestep the inherent lack of ground truth in interpretability \citep{templeton2024scaling}. We hope our work provides a framework for evaluating localizations and explanations.

\subsubsection*{Acknowledgments}
We thank Stephen Casper, Aengus Lynch, Max Li, and Peter Hase for their advice with unlearning methods, interpreting results, and discussing future directions. 
We also thank Daniel M. Roy, Amr Khalifa, Eleni Triantafillou, Fabien Roger, and Neel Nanda for feedback on various drafts of this work. This project used compute provided by the Center for AI Safety, and part of this project was completed during the ML Alignment \& Theory Scholars Program.

\newpage

\bibliography{iclr2025_conference}
\bibliographystyle{iclr2025_conference}

\newpage
\appendix

\section{Appendix}
\label{appendix}

\subsection{Sports Unlearning Results}
\label{app:sports-unlearning-results}

For unlearning on the \Sportunlearn task, we use a loss function $$L = \lambda_1 L_{\text{forget}} + \lambda_2 L_{\text{retain}} + \lambda_3 L_{\text{SFT}},$$ where $L_{\text{forget}}$ is an unlearning loss on the $D_{forget}$ subset of facts we want to forget, $L_{\text{retain}}$ is a cross-entropy loss on the remaining facts, and $L_{\text{SFT}}$ is a cross-entropy loss on the Pile dataset \citep{gao2020pile}. The unlearning loss $L_{\text{forget}}$ we use is the $\log(1-p)$ measure (where p is the probability of the correct sport) from \citet{mazeika2024harmbench} due to its empirical stability and fewer side effects: vanilla gradient ascent more strongly incentivizes the model to have significantly lower logprobs than wouldn't be encountered in a model that has not been trained on the factual association, and it detracts from model maintenance of $L_{\text{retain}}$ and $L_{\text{SFT}}$. 

We present results on using various localizations on the \Sportunlearn task in \cref{tab:ft-sports-evals}. The \LT localization allows unlearning to be more robust to the MCQ prompt format while maintaining performance on the MMLU dataset. 

\begin{table}[h]\small
\caption{Localized fine-tuning accuracy on standard evaluations: unlearning all basketball athletes and retaining all other facts.}

\centering
\vskip 0.15 in
\begin{center}
\begin{scriptsize}
\begin{sc}
\begin{tabular}{lcccr}
\hline
\toprule
Localization        & Forget $\downarrow$         & Retain $\uparrow$         & MCQ $\downarrow$            & MMLU $\uparrow$          \\
\midrule
Attrib. Patching     & \textbf{0.000}      & \textbf{1.000}       & 0.767          & 0.602          \\
Causal Tracing     & 0.201      & 0.998       & 0.849          & 0.611          \\
\LT & 0.002      & 0.995       & \textbf{0.110}          & \textbf{0.613} \\
Random & 0.952               & 0.980                & 0.822          & 0.612          \\
All-MLPs & \textbf{0.000} & 0.994 & 0.279 & 0.606 \\
Nonlocalized   & \textbf{0.000}               & 0.985                & 0.196 & 0.595          \\
\bottomrule
\end{tabular} \label{tab:ft-sports-evals}
\end{sc}
\end{scriptsize}
\end{center}
\vskip -0.1in
\end{table}

\subsection{\LT Interpretability Analysis}
\subsubsection{Sports Facts}
\label{app:gemma_mi}
\begin{figure}[h]
\centering
\begin{minipage}{0.49\textwidth}

\includegraphics[width=7cm]{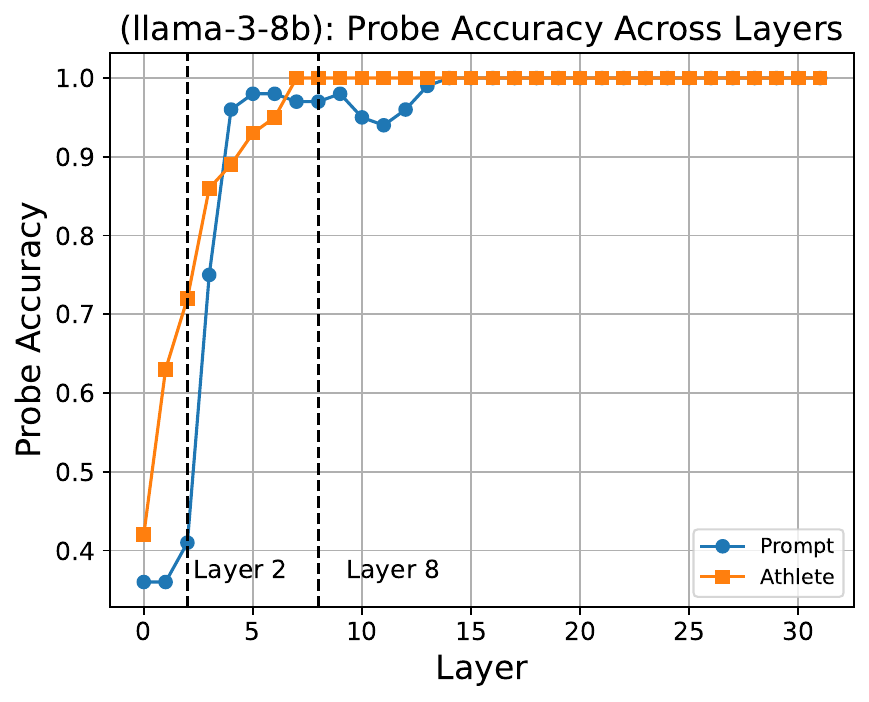}
\caption{Accuracy of probes predicting the correct sport across layers for different models.}
\label{fig:localize-probing}
\end{minipage}
\end{figure}

We redo analysis from \citet{nanda2023factfinding} on Gemma-7B, Gemma-2-9B, and Llama-3-8B. We train logistic regression models ("probes") to predict the correct sport given the internal representation of the model at a layer. We find that probes predicting the correct sport increase in accuracy significantly in layers 2 through 7 in Gemma-7B and 2 through 8 for Gemma-2-9B and Llama-3-8B (\cref{fig:localize-probing}). 

Unlike \citet{nanda2023factfinding}, however, we find attention heads past layer 2 that impact the linear representation of attributes and thus could potentially be important for fact lookup. However, because they could likely play a variety of other different roles such as token concatenation, following the findings of \citet{geva2023dissecting, nanda2023factfinding} that MLPs do primary factual representation enrichment, in this work we only consider the MLPs as our localization.

\begin{figure}[h!]
\centering
\begin{minipage}{0.42\textwidth}
  \centering
\includegraphics[width=7cm]{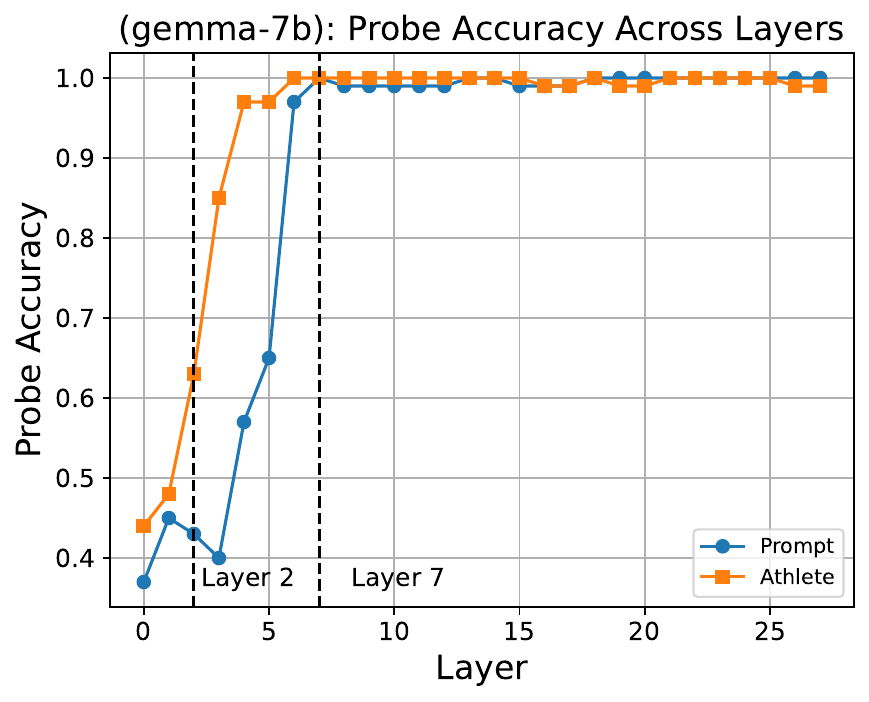}
\end{minipage}%
\hfill
\begin{minipage}{0.49\textwidth}
  \centering
\includegraphics[width=7cm]{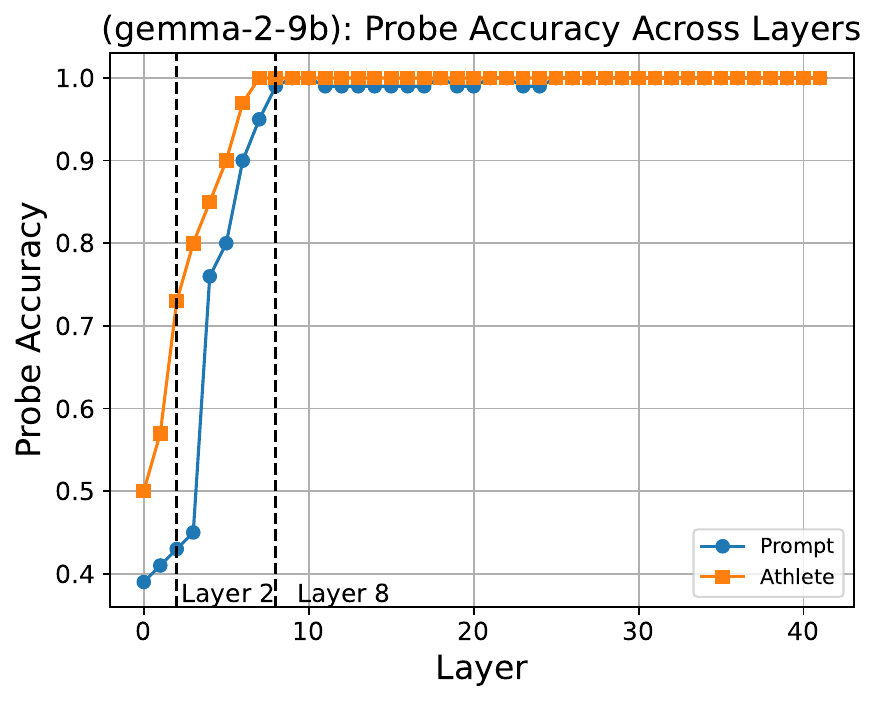}
\end{minipage}
\end{figure}

\newpage
\subsubsection{CounterFact}

\begin{figure}[h!]
\centering
\begin{minipage}{0.32\textwidth}
  \centering
  \includegraphics[width=\textwidth]{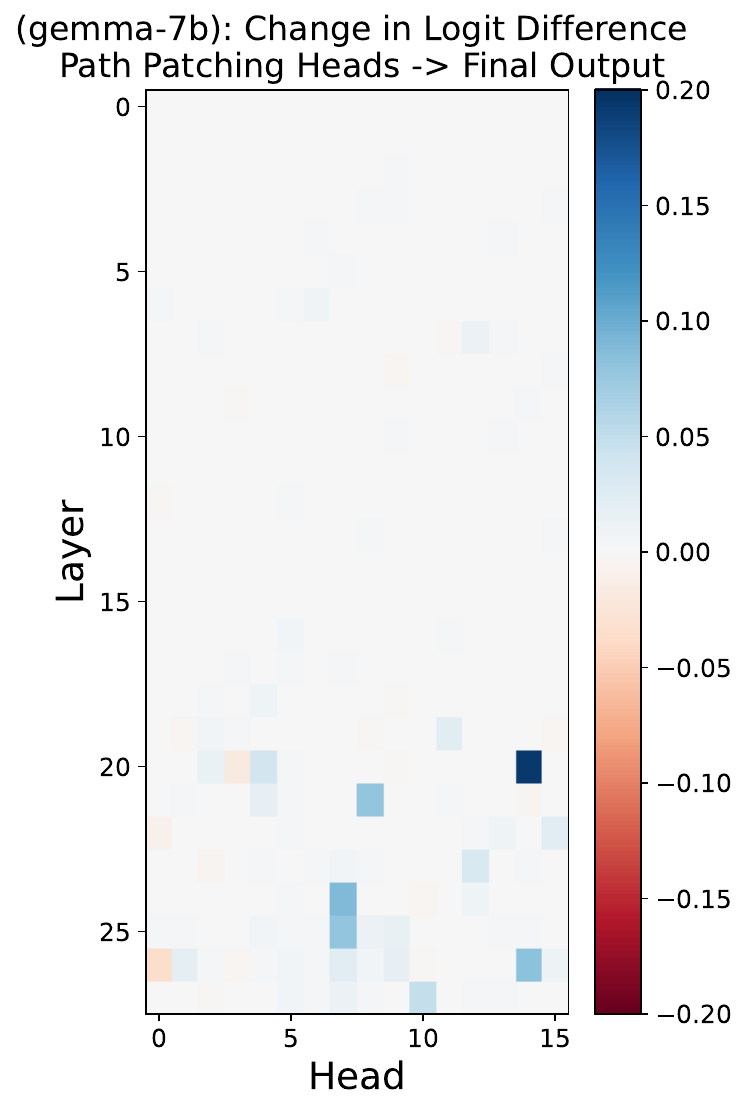}
\end{minipage}%
\hfill
\begin{minipage}{0.32\textwidth}
  \centering
  \includegraphics[width=\textwidth]{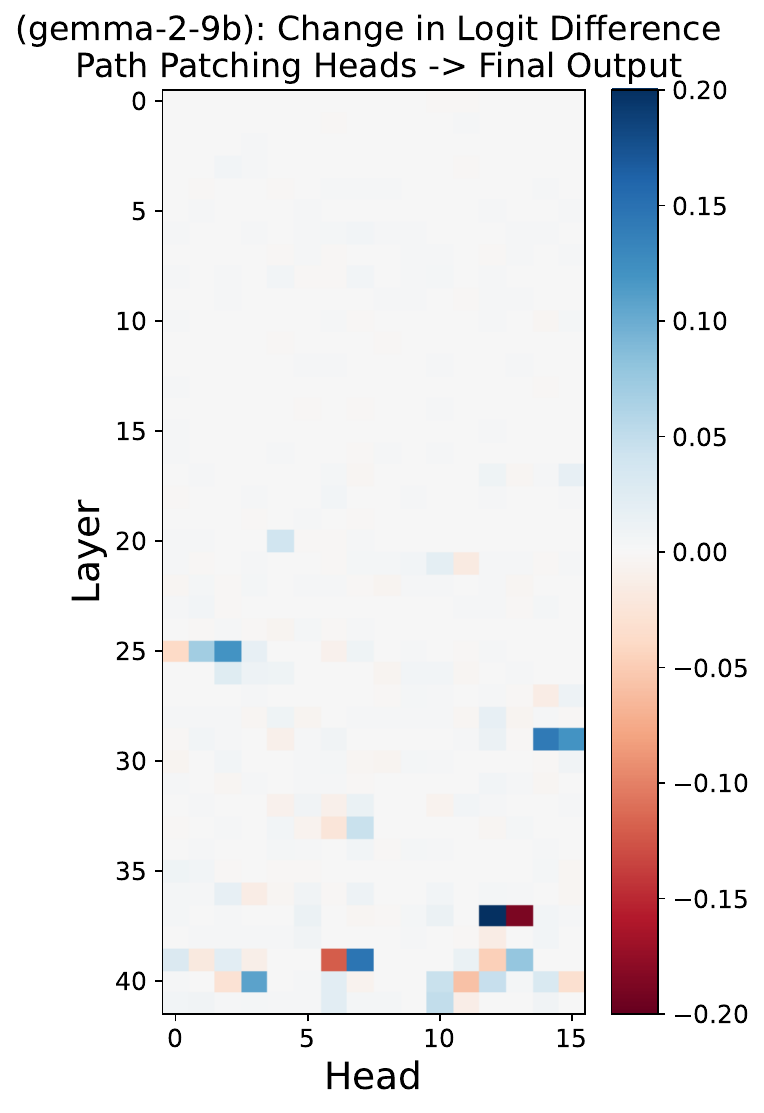}
\end{minipage}%
\hfill
\begin{minipage}{0.32\textwidth}
  \centering
  \includegraphics[width=\textwidth]{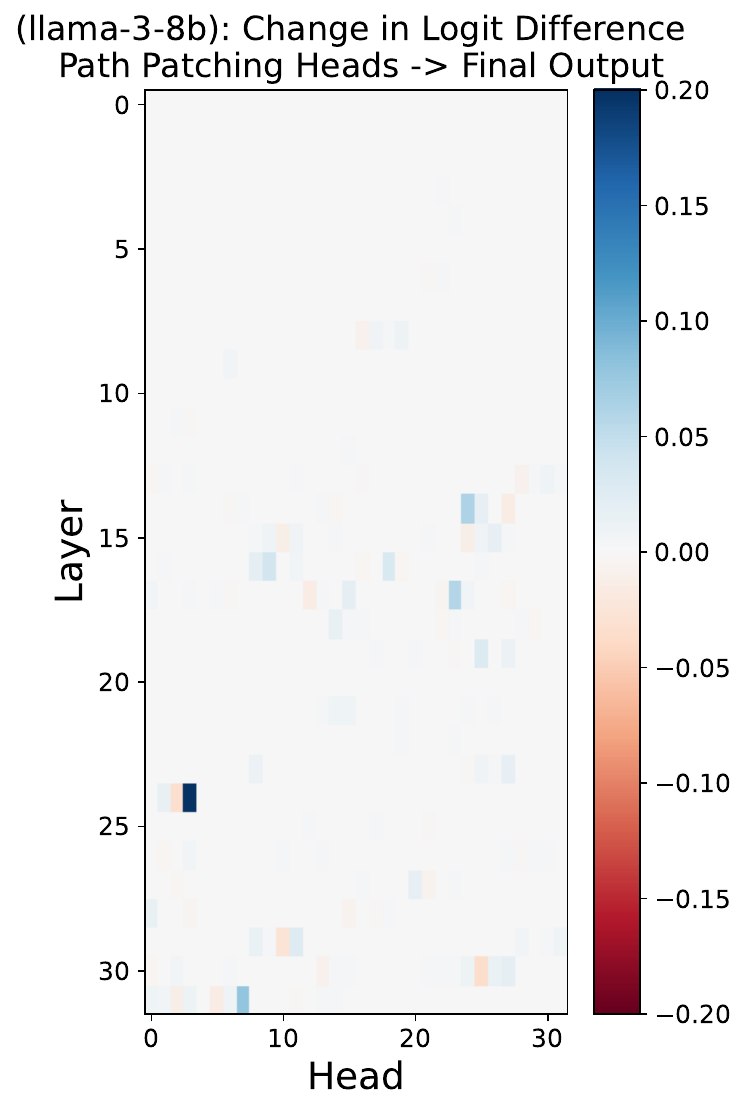}
\end{minipage}
\caption{Change in logit difference when patching attention heads to the output.}
\label{fig:counterfact-extraction}
\end{figure}

We repeat analysis from \citet{nanda2023factfinding} and \citet{geva2023dissecting} on Gemma-2-9B. We first measure the effect on the difference in logits between correct and incorrect answers of facts when patching the direct path of attention heads and MLPs to the final output, shown in \cref{fig:counterfact-extraction}. An attention head or MLP will have a large effect on the logit difference if it is important in moving the factual information to the last token position or decoding it into the correct answer. We call these components part of the "fact extraction mechanism", and aim to find the source of the factual information moved by this mechanism.

\begin{figure}[h!]
\centering
\begin{minipage}{0.45\textwidth}

\includegraphics[width=7cm]{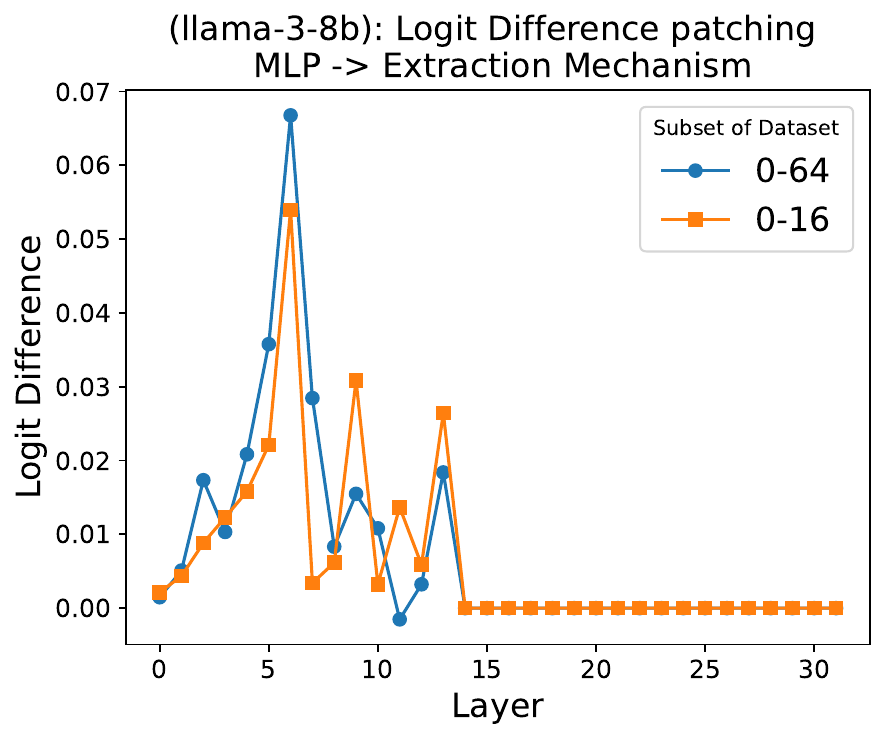}
\caption{Logit difference when patching MLPs to the extraction mechanism found above for different models.}
\label{fig:counterfact-enrichment}
\end{minipage}
\end{figure}

To find this source, we patch the outputs of MLPs to this "fact extraction mechanism" and measure the resultant change in logit difference (\cref{fig:counterfact-enrichment}). An MLP would cause a large change in logit difference if it caused relevant representations to form that are then moved by the "fact extraction heads" to increase the probability of the correct output. We provide the logit differences for all 64 facts along with just the first 16 facts, and see that the logit differences are similar across the dataset splits. We take the MLPs with the highest change ($> 0.02$) and include them in our FLU localization of CounterFact.

\begin{figure}[h!]
\centering
\begin{minipage}{0.49\textwidth}
  \centering
\includegraphics[width=7cm]{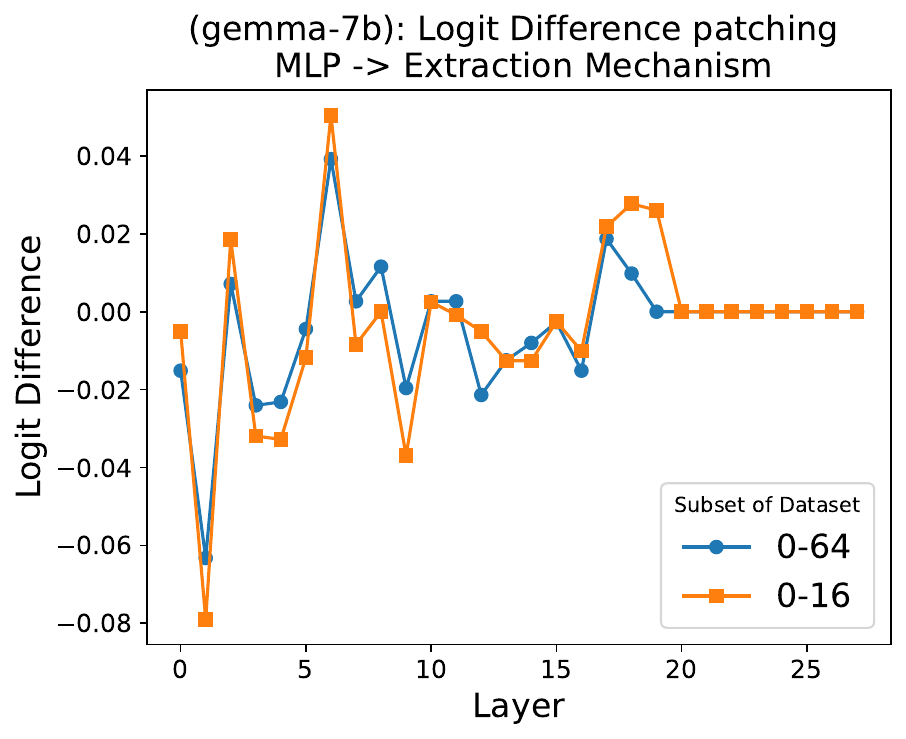}
\end{minipage}%
\hfill
\begin{minipage}{0.49\textwidth}
  \centering
\includegraphics[width=7cm]{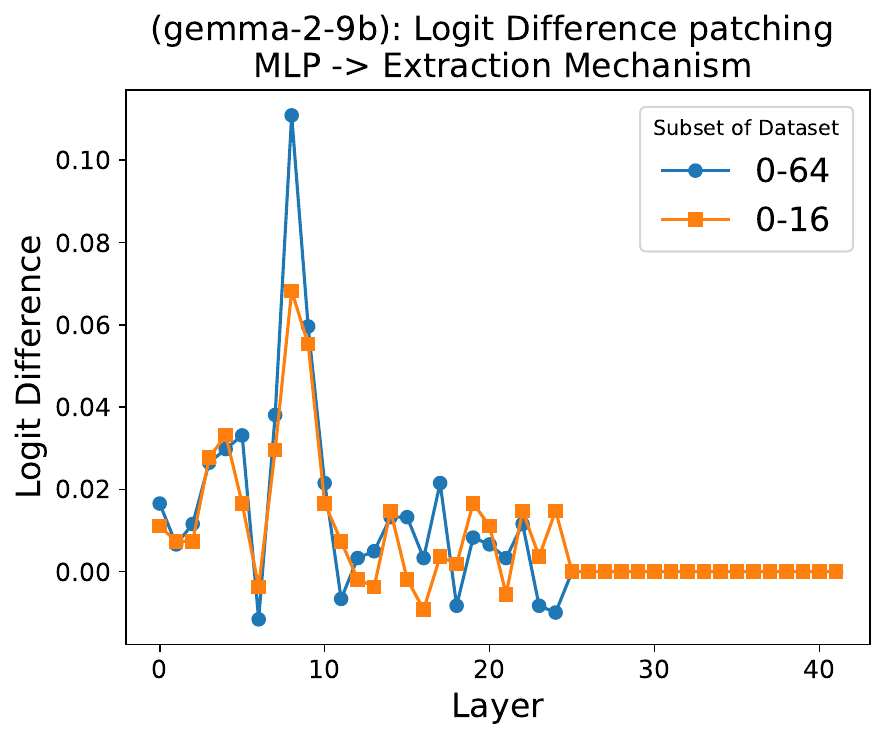}
\end{minipage}
\end{figure}
\vspace{-1pt}

\FloatBarrier
\subsection{\OT Selected Components} \label{app:ot_components}
\begin{figure}[h!]
\centering
\begin{minipage}{\textwidth}
  \centering
\includegraphics[width=9cm]{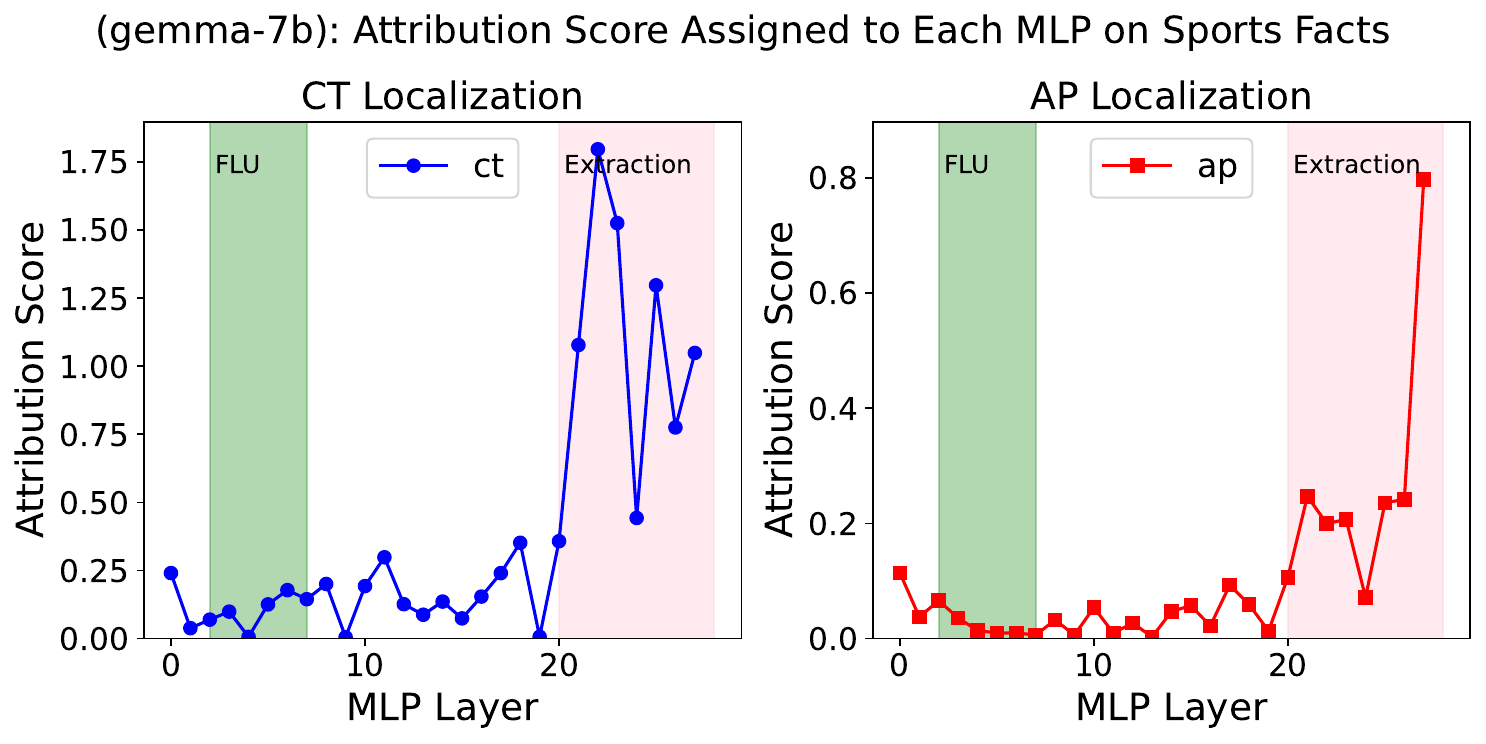}
\caption{Attribution scores on MLPs on sports facts for Gemma-7B.}
\label{fig:gemma_7b_sports_ap_mlp_scores}
\end{minipage}%
\end{figure}

\begin{figure}[h!]
\centering
\begin{minipage}{\textwidth}
  \centering
\includegraphics[width=9cm]{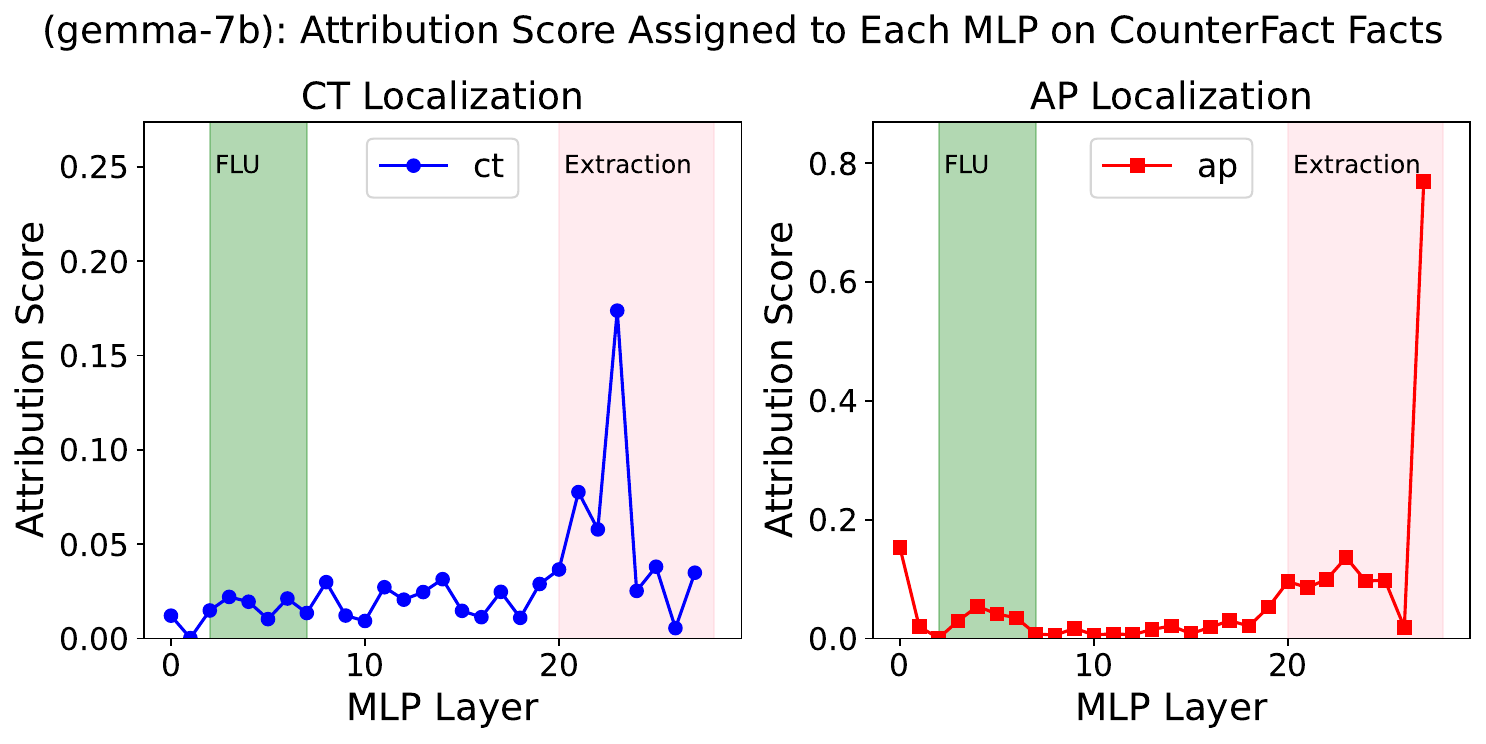}
\caption{Attribution scores on MLPs on CounterFact facts for Gemma-7B.}
\label{fig:gemma_7b_counterfact_ap_mlp_scores}
\end{minipage}%
\end{figure}

What MLPs do the automated \OT methods localize? We explore the attribution scores of the automated localization methods (causal tracing and attribution patching) on the MLPs to see if automated localization methods can detect the \LT mechanism. In \cref{fig:gemma_7b_sports_ap_mlp_scores,fig:gemma_7b_counterfact_ap_mlp_scores}, for Gemma-7B, we see that both CT and AP localizations target the later layer MLPs instead of the \LT mechanism (\cref{fig:gemma_7b_sports_ap_mlp_scores}). 

\begin{figure}[h!]
\centering
\begin{minipage}{\textwidth}
  \centering
\includegraphics[width=9cm]{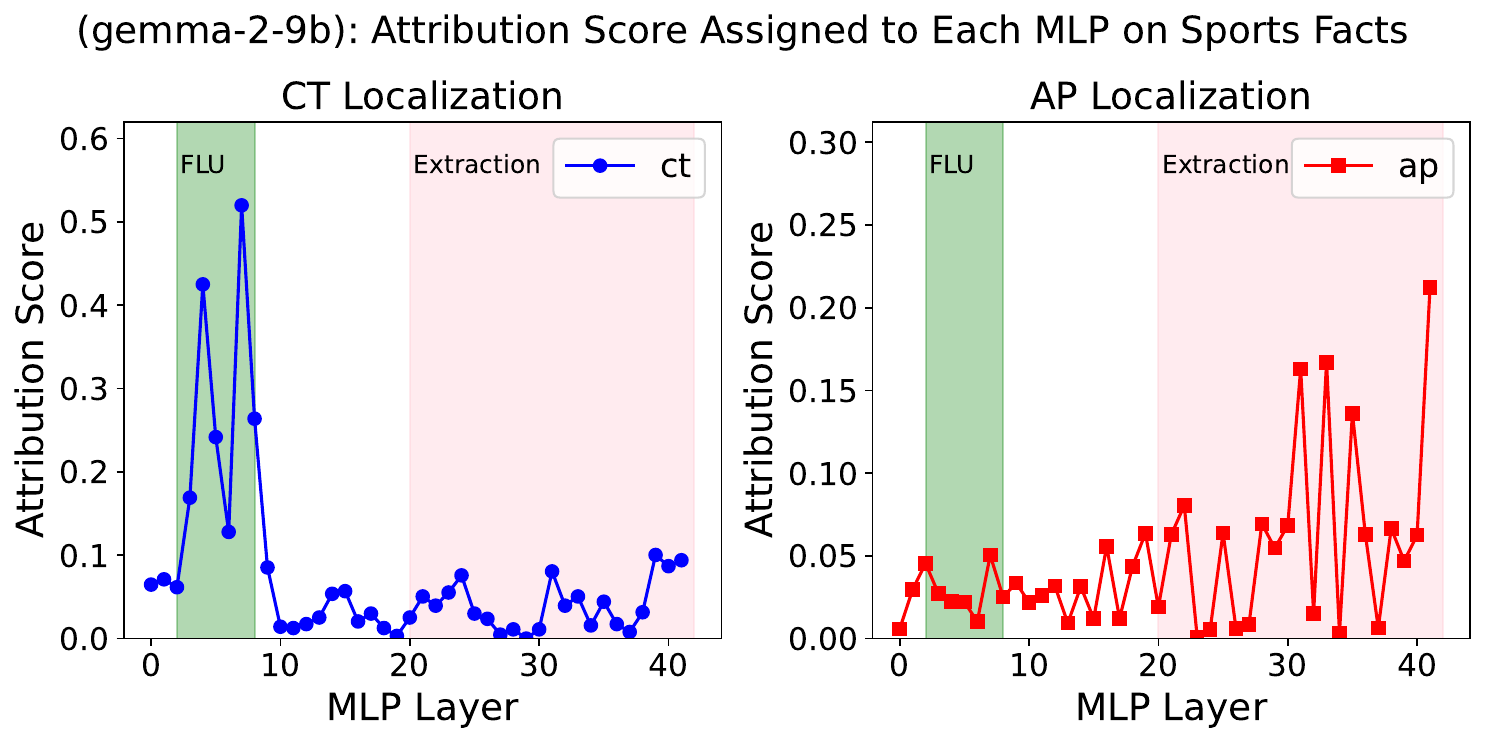}
\caption{Attribution scores on MLPs on sports facts for Gemma-2-9B.}
\label{fig:gemma_9b_sports_ap_mlp_scores}
\end{minipage}%
\end{figure}

\begin{figure}[h!]
\centering
\begin{minipage}{\textwidth}
  \centering
\includegraphics[width=9cm]{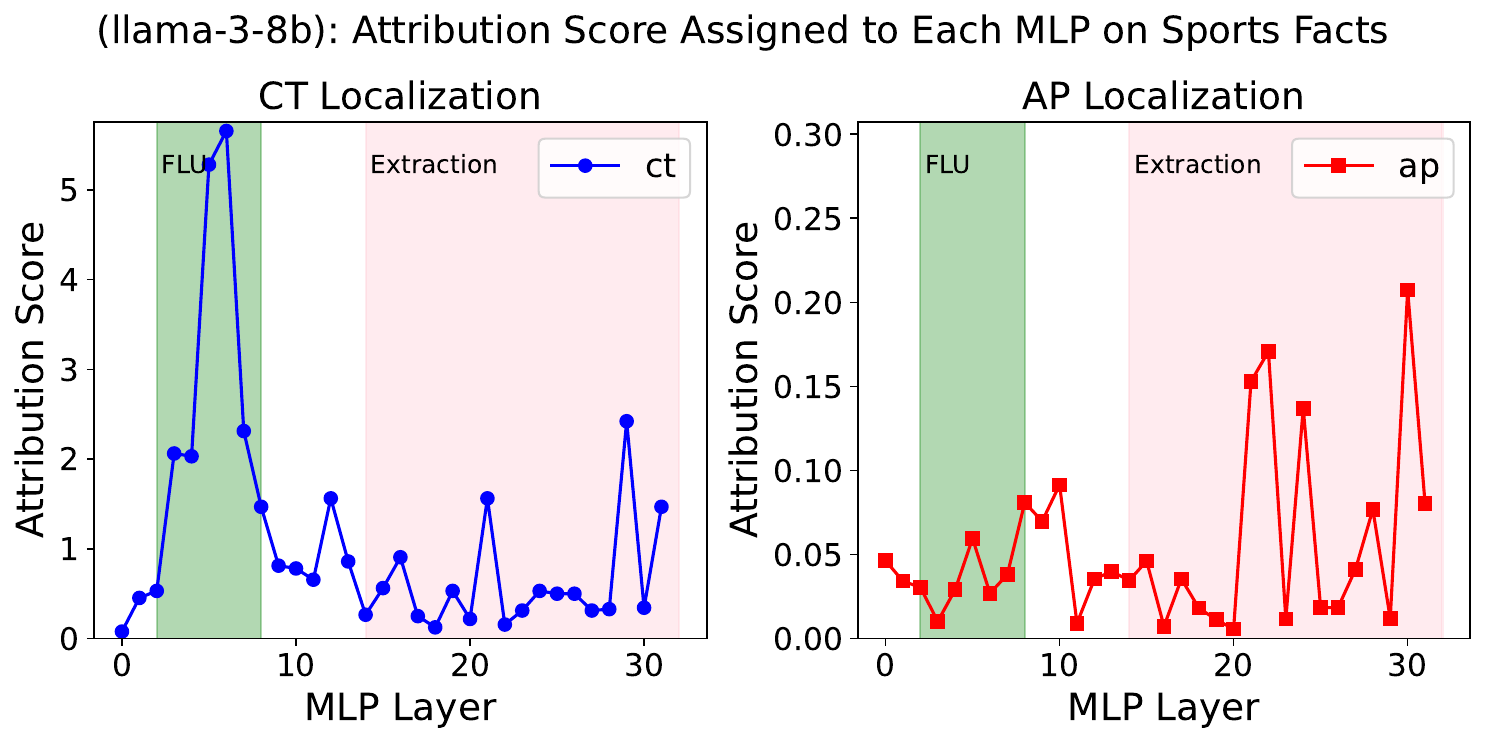}
\caption{Attribution scores on MLPs on sports facts for Llama-3-8B.}
\label{fig:llama_8b_sports_ap_mlp_scores}
\end{minipage}%
\end{figure}

For AP localization, this trend continues with Gemma-2-9B (\cref{fig:gemma_9b_sports_ap_mlp_scores}) and Llama-3-8B (\cref{fig:llama_8b_sports_ap_mlp_scores}). However, CT localization does highlight some of the early layer MLPs that are in the \LT mechanism, especially for Gemma-2-9B.

\begin{figure}[h!]
\centering
\begin{minipage}{\textwidth}
  \centering
\includegraphics[width=9cm]{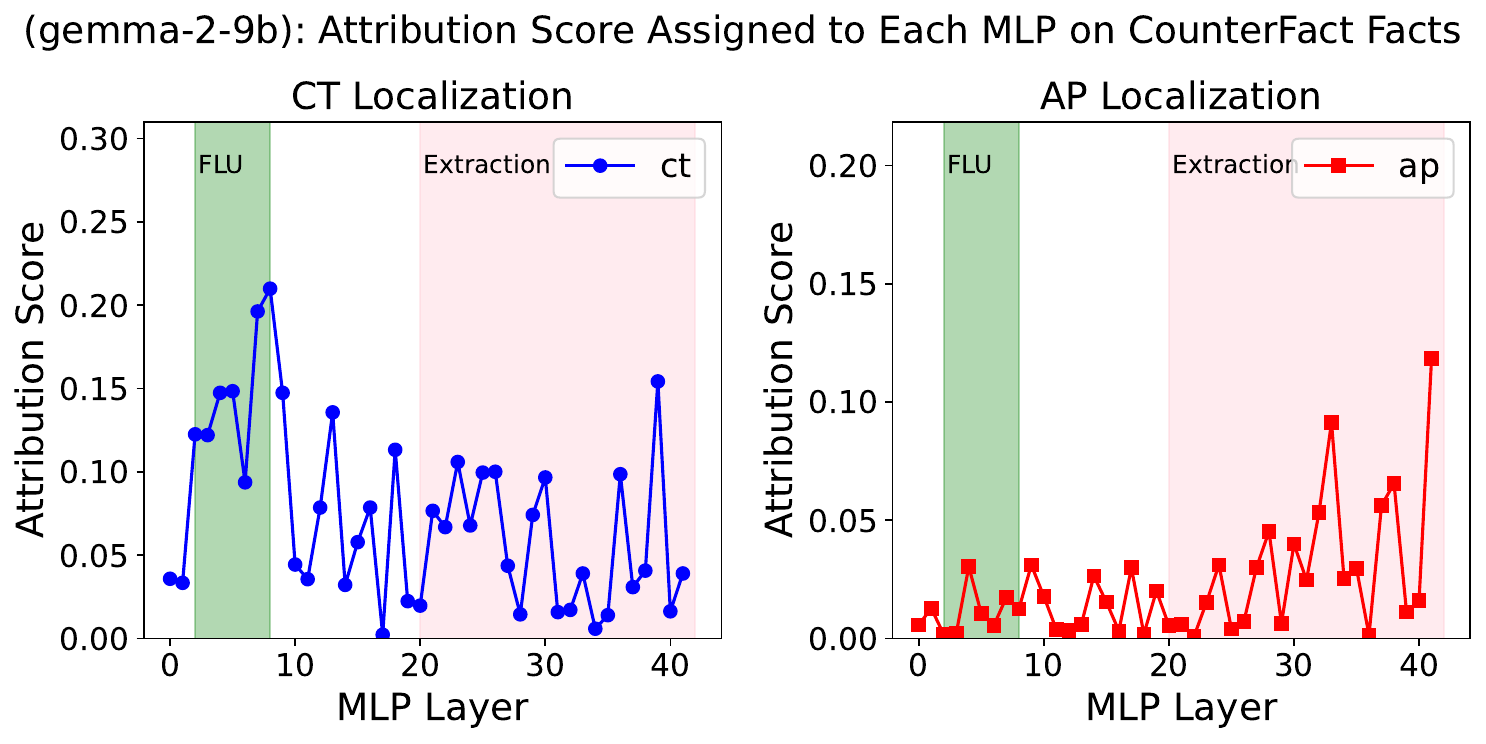}
\caption{Attribution scores on MLPs on CounterFact facts for Gemma-2-9B.}
\label{fig:gemma_9b_counterfact_ap_mlp_scores}
\end{minipage}%
\end{figure}

\begin{figure}[h!]
\centering
\begin{minipage}{\textwidth}
  \centering
\includegraphics[width=9cm]{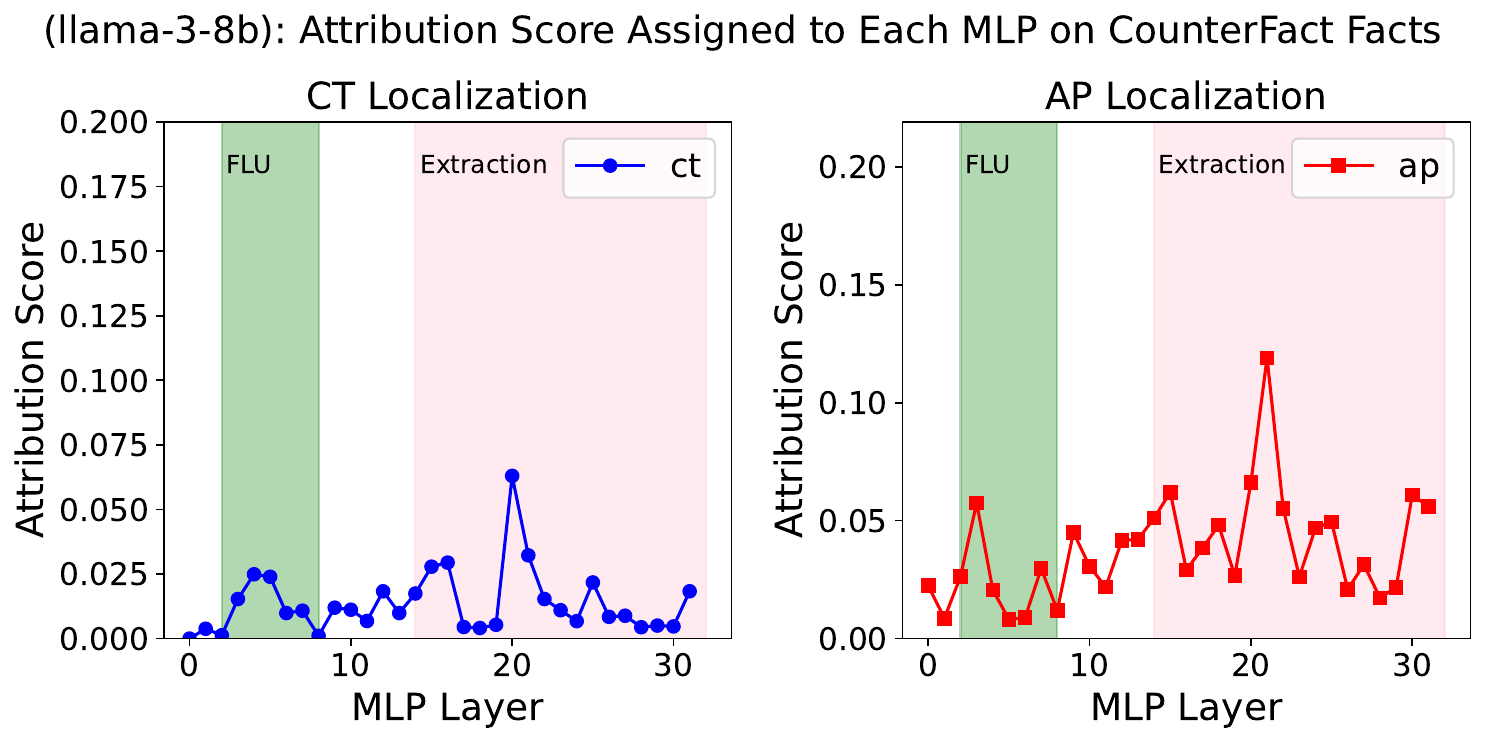}
\caption{Attribution scores on MLPs on CounterFact facts for Llama-3-8B.}
\label{fig:llama_8b_counterfact_ap_mlp_scores}
\end{minipage}%
\end{figure}

We repeat this analysis on CounterFact in \cref{fig:gemma_7b_counterfact_ap_mlp_scores,fig:gemma_9b_counterfact_ap_mlp_scores,fig:llama_8b_counterfact_ap_mlp_scores}. Again we see AP localizations in particular
assign higher scores to later-layer MLPs, and CT only highlights \LT components on Gemma-2-9B, localizing other extraction layers on the other models.

\subsection{Weight Masking} \label{sec:weight_masking}

In this section we employ weight masking to quantify the size of weight updates needed to unlearn/edit facts, for more direct comparisons. Our loss function $L = \lambda_1 L_{\text{forget}} + \lambda_2 L_{\text{retain}} + \lambda_3 L_{\text{SFT}}$ + $\lambda_4 L_{\text{reg}}$ now includes an L1 regularization term to control the sparsity. We empirically evaluate how a learned binary mask over individual weights of the localized components can produce editing/unlearning, and vary the size of this mask. ``Manual Interp" refers to the \LT localization technique for all the following results in this section. 

\subsubsection{Unlearning Sports}
\begin{figure}[h!] \centering 
    \includegraphics[width=0.45\linewidth]{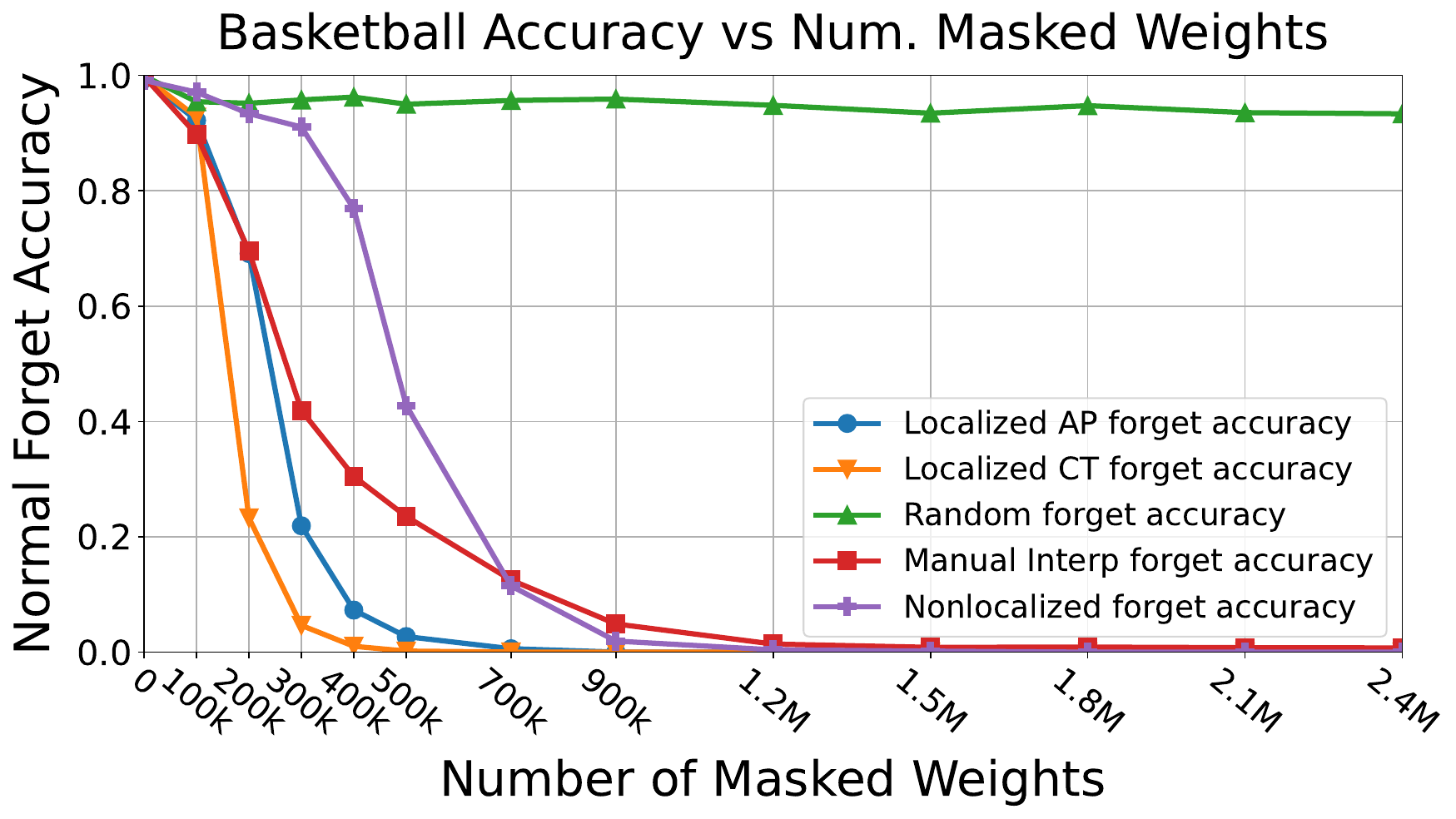}
    \includegraphics[width=0.45\linewidth]{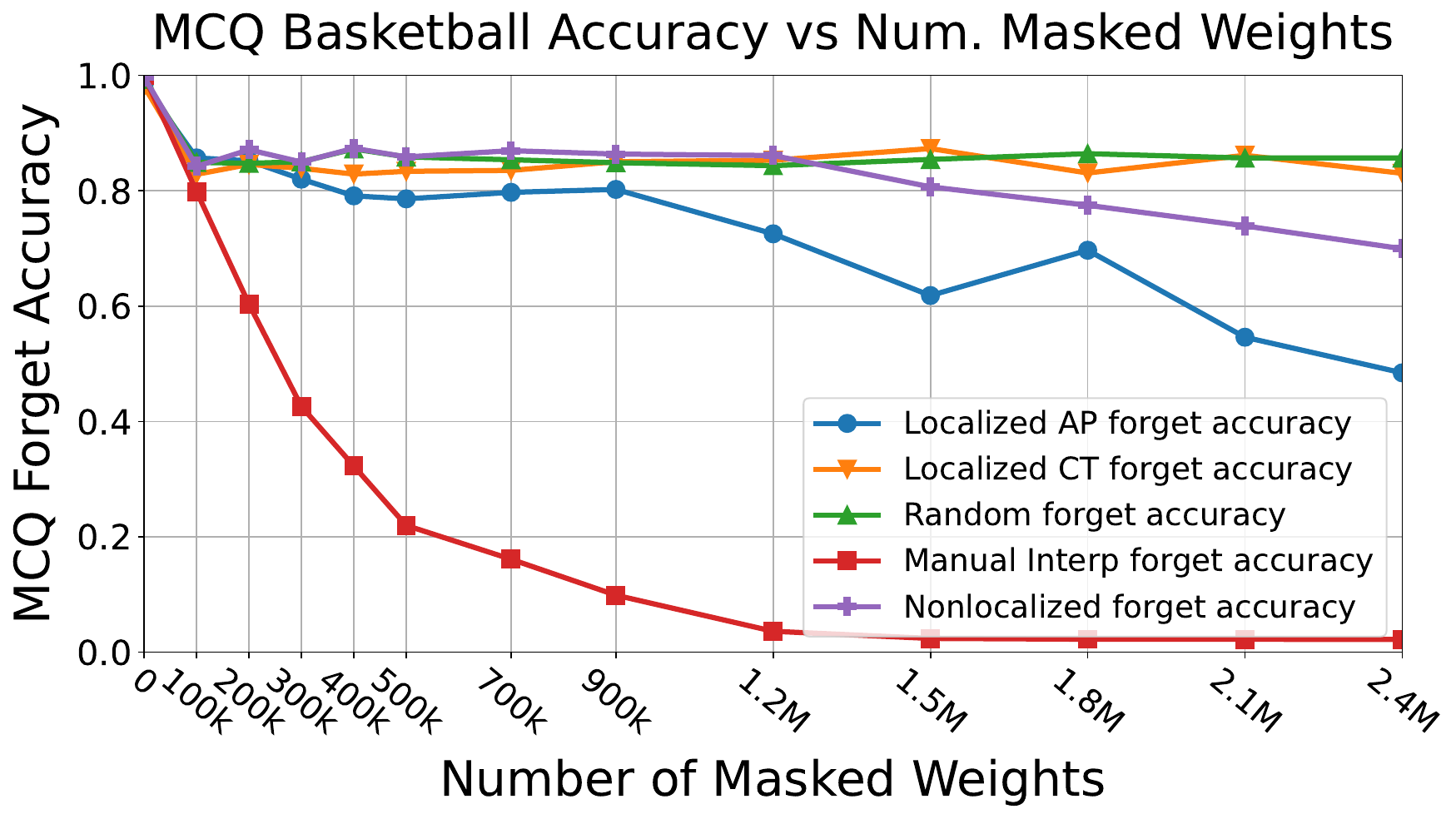}
    \caption{\textbf{(Left)} Testing the models' unlearning of basketball athletes against the number of weights masked. \textbf{(Right)} Testing the models' unlearning of basketball athletes against the number of weights masked, in the MCQ prompt format.}
    \label{fig:wm-forget-acc-sport}
\end{figure}

We show standard evaluations across a sweep of discretization thresholds, which directly corresponds to the size of the model edit. \cref{fig:wm-forget-acc-sport} shows the accuracy on the forget and retain sets for unlearning basketball across different edit sizes. Here, we see all methods being effective in unlearning basketball facts while retaining all other facts. While AP and CT localizations cause the model to have zero accuracy on the in-distribution set with much fewer masked weights needed, when checking for generalization using a multiple-choice format we clearly see that only manual localization has successfully generalized the unlearning of basketball facts (\cref{fig:wm-forget-acc-sport}, right).

\begin{figure}[t]
\centering
\includegraphics[width=0.4\textwidth]{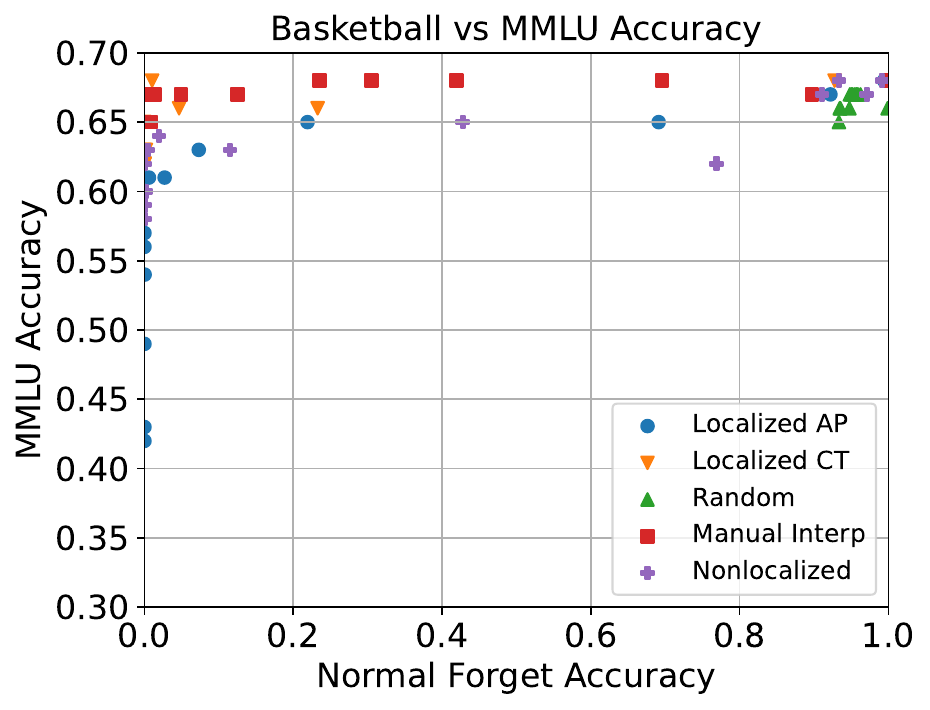}
\includegraphics[width=0.4\textwidth]{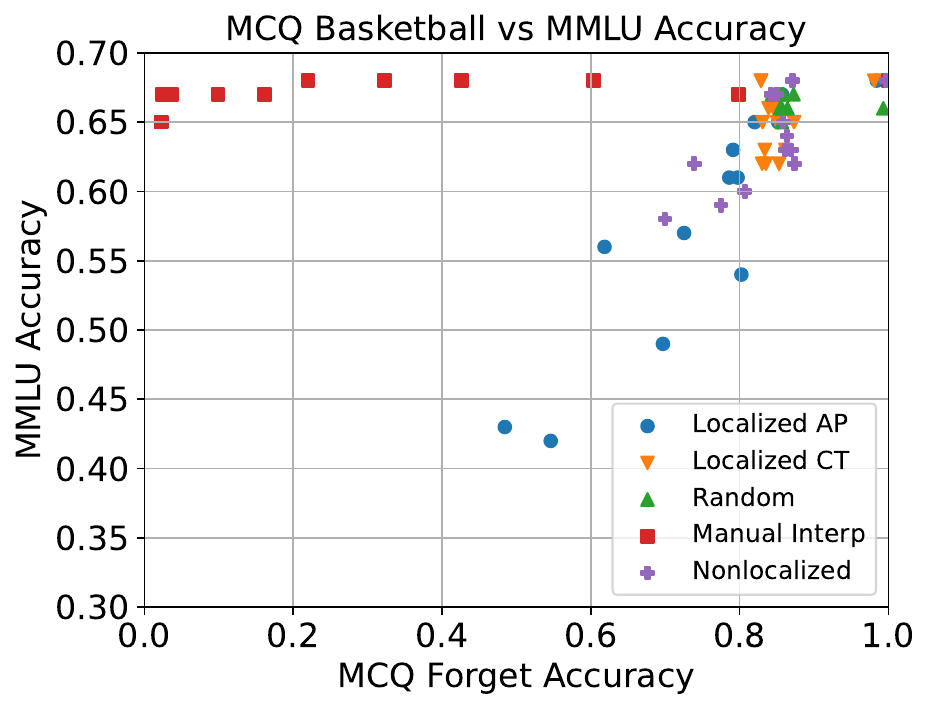}
\caption{Unlearning basketball facts. \textbf{(Left)} Measuring MMLU and forget set performance across different discretization thresholds.
\textbf{(Right)} Measuring MMLU and MCQ forget set performance across
different discretization thresholds.}
\label{fig:wm-normal-mmlu-sport}
\end{figure}

\begin{figure}[h]
  \centering
    \includegraphics[width=0.45\linewidth]{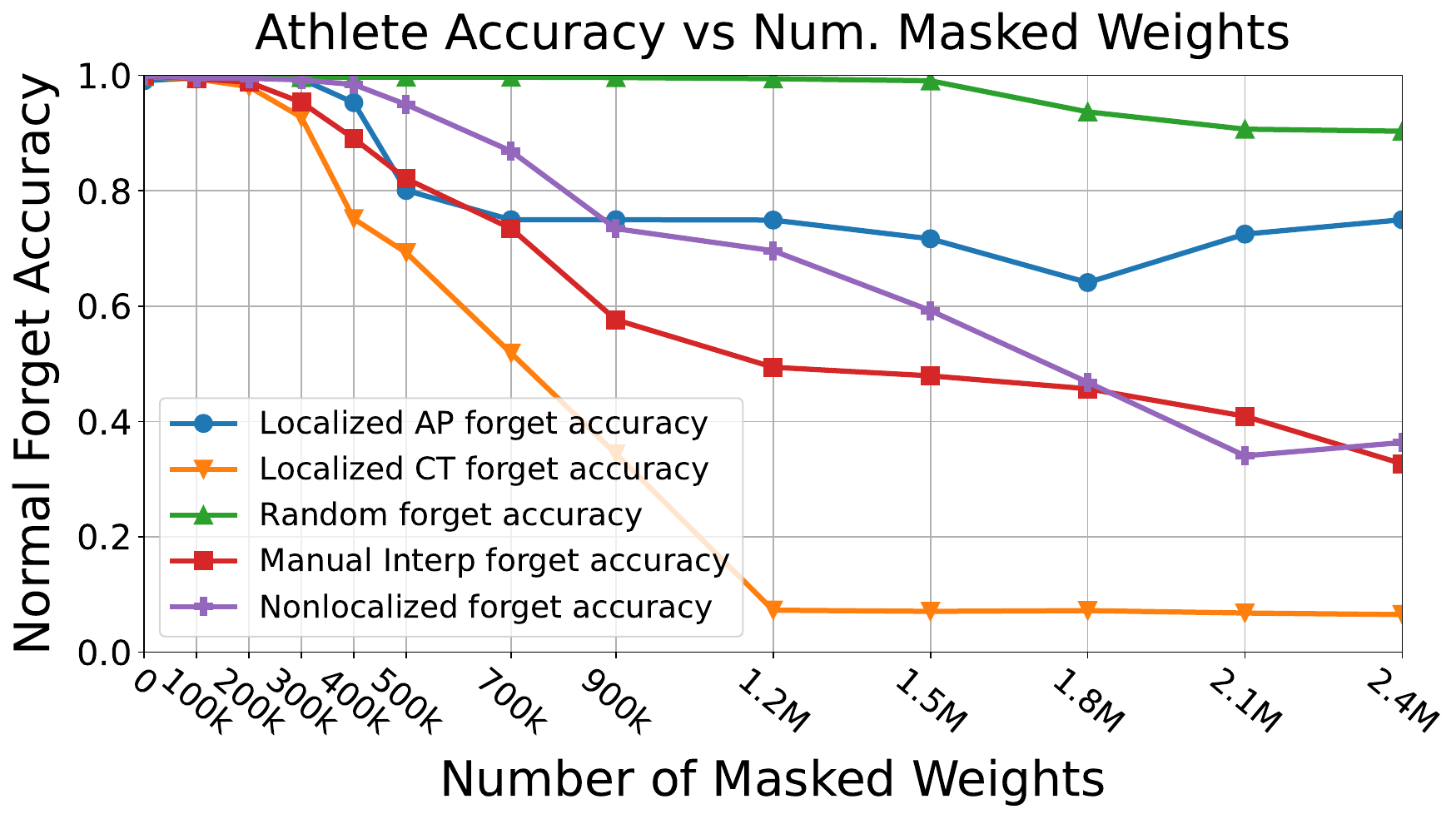}
    \includegraphics[width=0.45\linewidth]{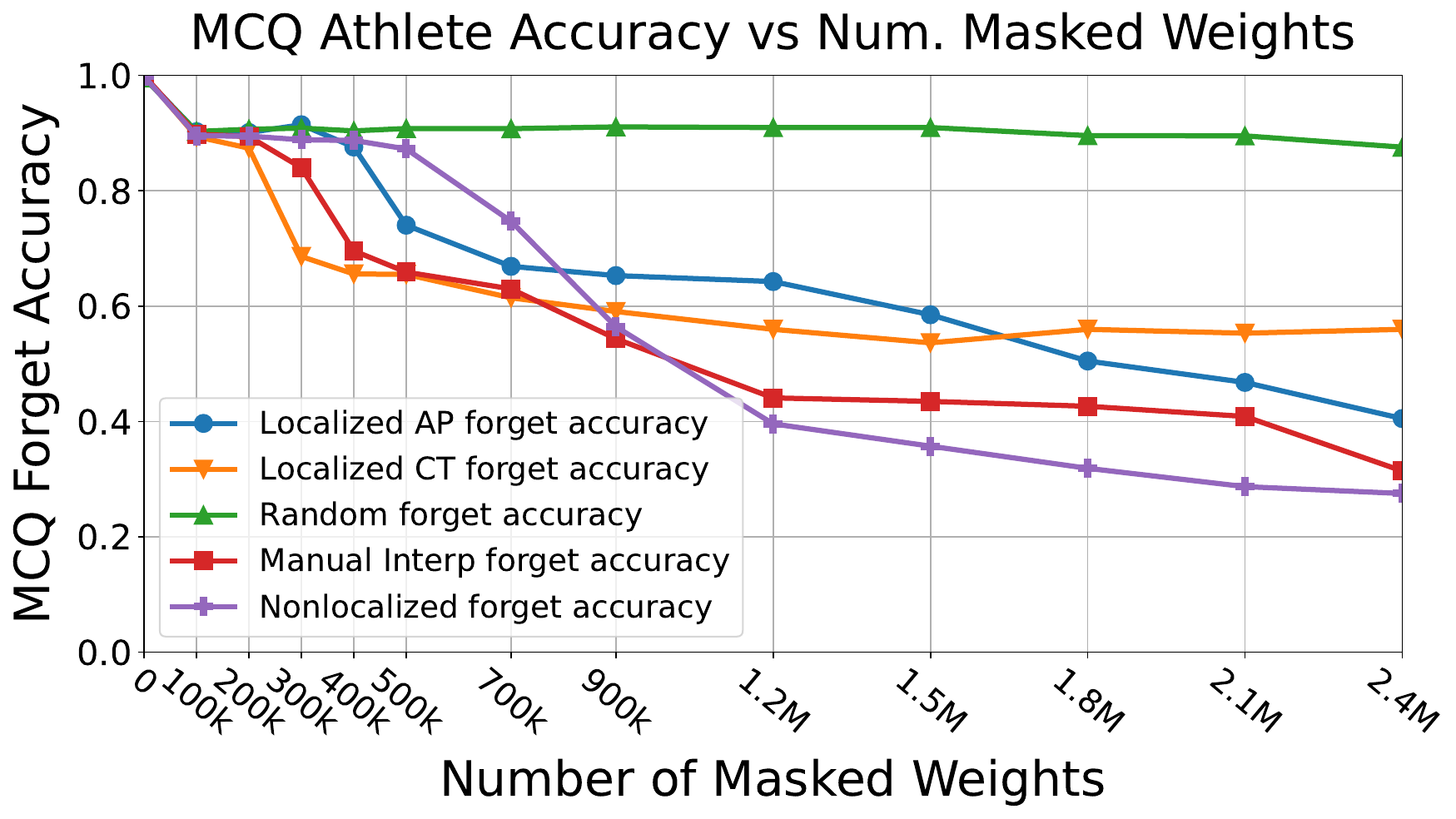}
    \caption{Editing subset of athletes. \textbf{(Left)} Measuring accuracy on the forget set.  \textbf{(Right)} Measuring accuracy on the forget set in the MCQ prompt format.}
    \label{fig:wm-forget-athlete}
\end{figure}

We find similar results when testing for performance degradation on MMLU (because we have to evaluate many model variations, we use a smaller MMLU test set from \citet{polo2024tinybenchmarks}). While all localized methods perform well when evaluated normally (\cref{fig:wm-normal-mmlu-sport}, left), \cref{fig:wm-normal-mmlu-sport} (right) shows manual localization generalizes for minimizing loss of MMLU capabilities while unlearning sports facts in the MCQ format compared to the other methods.

\subsubsection{Editing Athletes}

\begin{figure}[h!]
  \centering
    \includegraphics[width=0.45\textwidth]{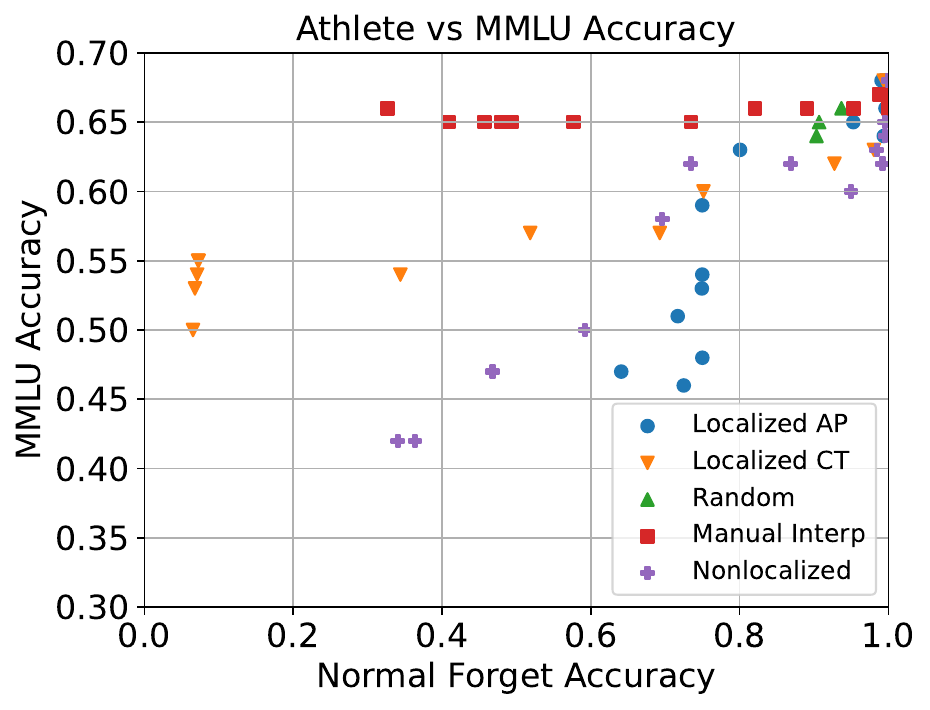}
    \includegraphics[width=0.45\textwidth]{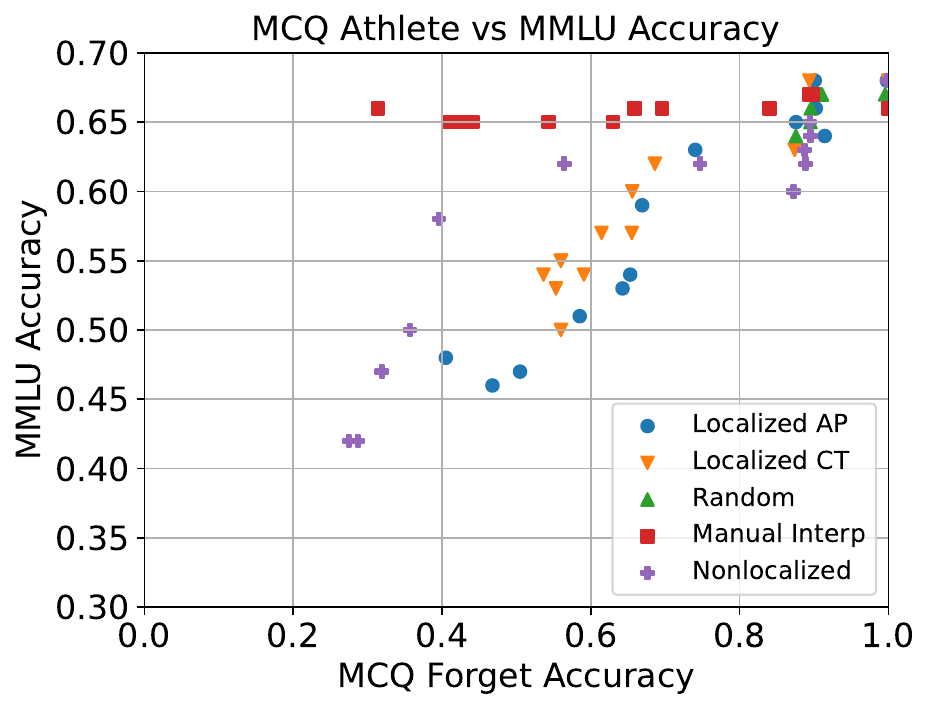}
    \caption{Editing subset of athletes. \textbf{(Left)} Measuring MMLU and forget set performance across different discretization thresholds. \textbf{(Right)} Measuring MMLU and MCQ forget set performance across different discretization thresholds.}
    \label{fig:wm-normal-mmlu-athlete}
\end{figure}

For editing the subset of athletes, \cref{fig:wm-forget-athlete} shows that causal tracing localization causes the model to have 0\% accuracy on the forget set, and \LT and nonlocalized editing cause the model to have near guessing rate (33\%) accuracy. However, only manual localization minimizes loss of capabilities while editing the athlete subset (\cref{fig:wm-normal-mmlu-athlete}).

Furthermore, no other method completely generalizes this unlearning to the MCQ prompt format (\cref{fig:wm-forget-athlete}), and manual localization remains superior in minimizing loss of capabilities while unlearning the athlete subset (\cref{fig:wm-normal-mmlu-athlete}, right).

\subsubsection{Editing CounterFact}
\begin{figure}[h!] \centering 
    \includegraphics[width=0.45\linewidth]{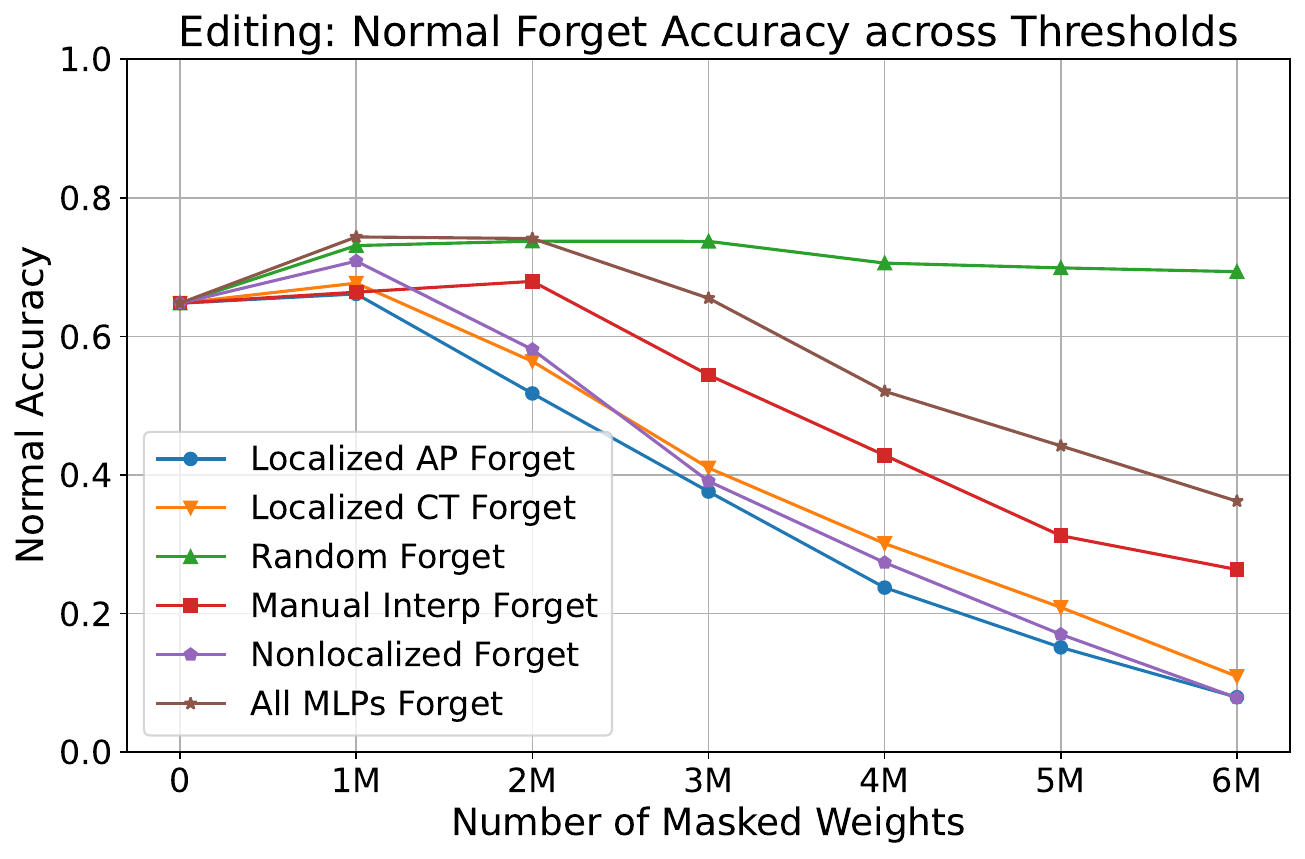}
    \includegraphics[width=0.45\linewidth]{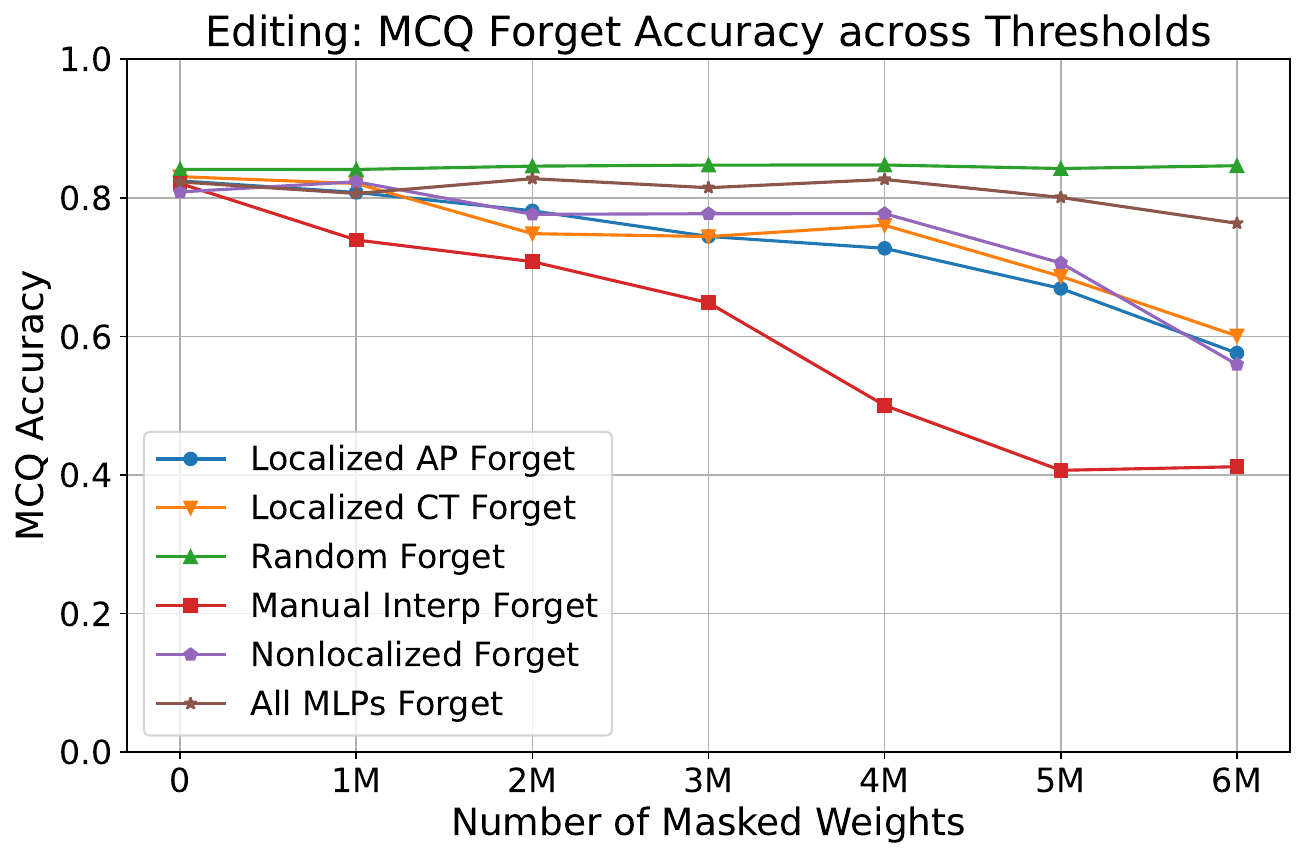}
    \caption{Editing CounterFact facts. \textbf{(Left)} Testing models accuracy on the normal forget set. \textbf{(Right)} Testing the models' accuracy in the MCQ prompt format.}
    \label{fig:wm-forget-acc-counterfact}
\end{figure}

We find similar results for editing on the CounterFact dataset. However, we find minimal difference in MMLU accuracy in all methods at all numbers of masked weights. Thus, we instead report the maintain and forget accuracies of these facts at different discretization thresholds in \cref{fig:wm-forget-acc-counterfact}. 

\begin{figure}[h!]
\centering
\begin{minipage}{0.45\textwidth}
  \centering
\includegraphics[width=0.9\linewidth]{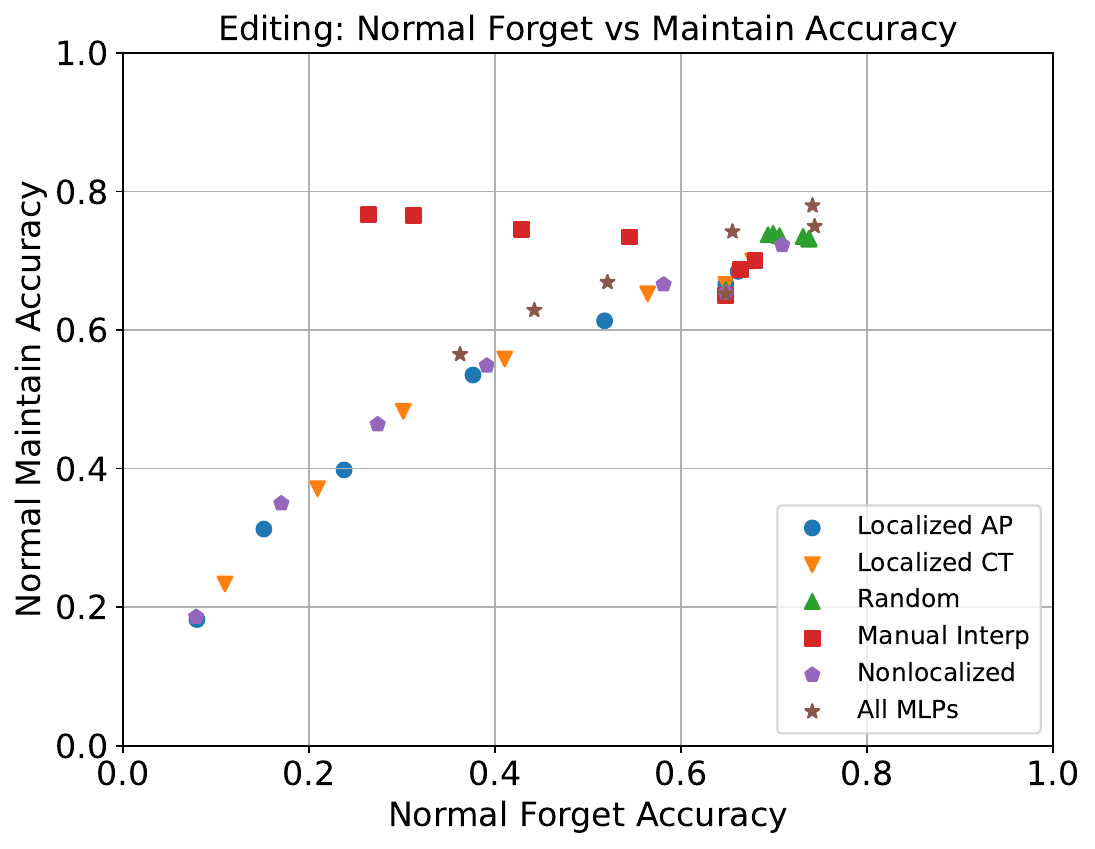}
\caption{Accuracy on normal forget set vs on the maintain set across localizations and discretization thresholds.}
\label{fig:counterfact_normal_forget_maintain}
\end{minipage}%
\hfill
\begin{minipage}{0.45\textwidth}
  \centering
\includegraphics[width=0.9\linewidth]{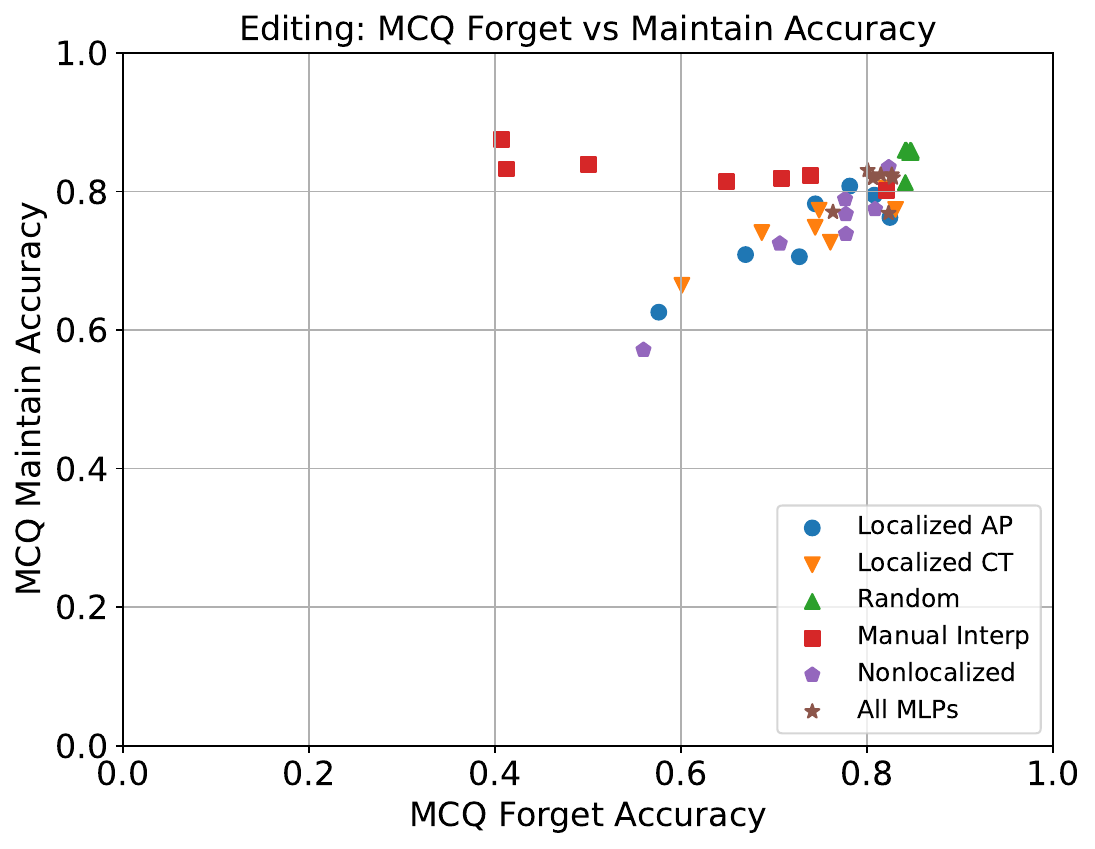}
\caption{Accuracy on multiple choice input vs on the maintain set across localizations and discretization thresholds.}
\label{fig:counterfact_mcq_forget_maintain}
\end{minipage}
\end{figure}

Additionally, we report a comparison of all localizations across discretization thresholds for normal and MCQ forget sets in \cref{fig:counterfact_normal_forget_maintain} and \cref{fig:counterfact_mcq_forget_maintain}. We see that \LT outperforms all other methods of localization in preserving maintain accuracy while decreasing forget accuracy.

\begin{figure}[h!]
\centering
\begin{minipage}{0.45\textwidth}
  \centering
\includegraphics[width=0.9\linewidth]{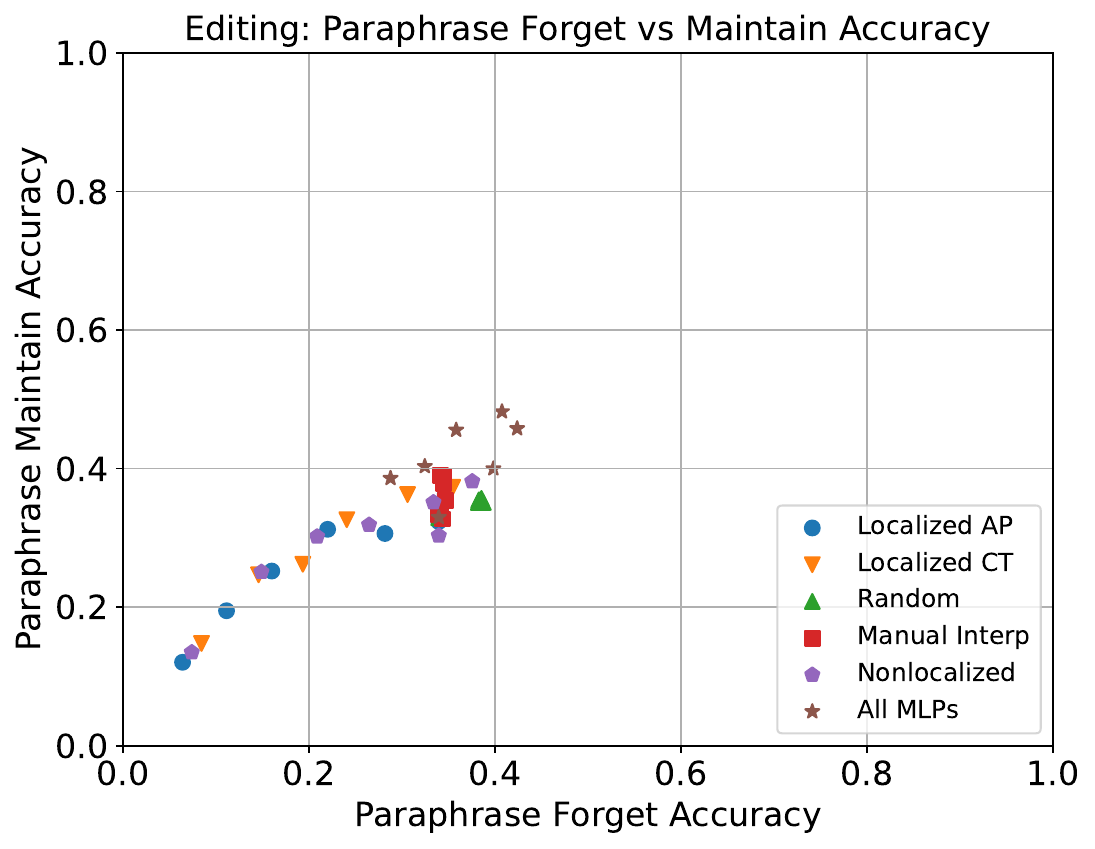}
\caption{Accuracy on paraphrased input vs on the maintain set across localizations and discretization thresholds.}
\label{fig:counterfact_paraphrase}
\end{minipage}%
\hfill
\begin{minipage}{0.45\textwidth}
  \centering
\includegraphics[width=0.9\linewidth]{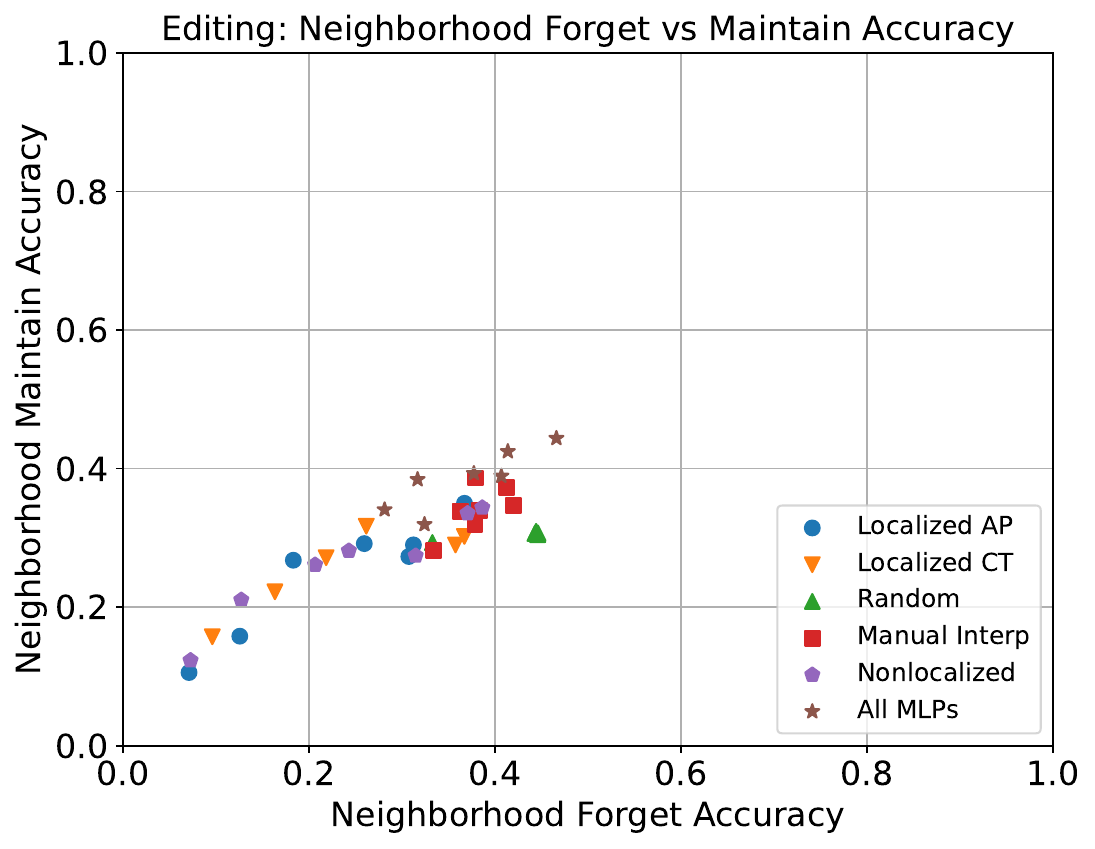}
\caption{Accuracy on "neighborhood" input vs on the maintain set across localizations and discretization thresholds.}
\label{fig:counterfact_neighborhood}
\end{minipage}
\end{figure}

We perform additional adversarial analysis of accuracies across different discretization thresholds. We report the "paraphrase" and "neighborhood" adversarial results in \cref{fig:counterfact_paraphrase} and \cref{fig:counterfact_neighborhood}, but find no significant results. 

\subsection{Mechanism Weight Analysis} \label{sec:mechanism_modification_analysis}
We analyze the actual components localized by each localization type and our baselines, for the CounterFact editing task. We seek to demonstrate that the \OT localizations and baselines target extraction mechanisms rather than just the \LT mechanisms. 

First, in \cref{tab:mechanism_weights}, we compare the parameter counts of the part of each mechanism that is present in each localization. \cref{tab:mechanism_weights} shows that causal tracing and attribution patching both have the potential to modify a considerable proportion of the extraction heads and extraction MLPs.

\begin{table}[h]
\centering
\caption{Comparison of total parameters of each mechanism that are present in each localization, for editing 16 facts from CounterFact}
\vskip 0.15 in
\begin{center}
\begin{scriptsize}
\begin{sc}
\begin{tabular}{lccc}
\hline
\toprule
Localization              & Extraction Heads & Extraction MLPs & Fact Lookup \\
\midrule
Total                       & 27,448,320                         & 1,027,604,480                     & 1,130,364,928 \\
\hline
Attrib. Patching            & 13,724,160 (50.0\%)                & 616,562,688 (60.0\%)              & 102,760,448 (9.1\%) \\
Causal Tracing              & 8,234,496 (30.0\%)                 & 308,281,344 (30.0\%)              & 411,041,792 (36.4\%) \\
\LT              & 0                          & 0                         & 1,130,364,928 (100.0\%) \\
All-MLPs                    & 0                          & 1,027,604,480 (100.0\%)           & 1,130,364,928 (100.0\%) \\
Nonlocalized                & 27,448,320 (100.0\%)               & 1,027,604,480 (100.0\%)           & 1,130,364,928 (100.0\%) \\
\bottomrule
\end{tabular}
\label{tab:mechanism_weights}
\end{sc}
\end{scriptsize}
\end{center}
\vskip -0.1in
\end{table}

Then, in \cref{tab:masked_mechanism_weights}, we compare the proportion of each mechanism that is masked when using a localized weight mask and discretizing to about 6 million weights. This is one approximate metric for how much each mechanism is modified by the localized editing. \cref{tab:masked_mechanism_weights} demonstrates that attribution patching, causal tracing, and nonlocalized editing all modify a higher proportion of the extraction head/MLP weights than the fact lookup mechanism weights.

\begin{table}[h]
\centering
\caption{Comparison of parameters of each mechanism that are masked by a trained weight mask, discretized to about 6 million weights}
\vskip 0.15 in
\begin{center}
\begin{scriptsize}
\begin{sc}
\begin{tabular}{lccccc}
\hline
\toprule
Localization Type          & Extraction Heads            & Extraction MLPs             & Fact Lookup \\
\midrule
\textbf{Total (baseline)}  & 27,448,320 (100\%)                  & 1,027,604,480 (100\%)               & 1,130,364,928 (100\%) \\
\hline
Attrib. Patching           & 165,300 (0.60\%)                    & 1,479,877 (0.14\%)                  & 1,385,198 (0.12\%) \\
Causal Tracing             & 30,828 (0.11\%)                     & 1,491,040 (0.15\%)                  & 1,424,059 (0.13\%) \\
\LT             & 0 (0.0\%)                           & 0 (0.0\%)                           & 6,248,039 (0.55\%) \\
All-MLPs                   & 0 (0.0\%)                           & 1,378,744 (0.13\%)                  & 1,663,772 (0.15\%) \\
Nonlocalized               & 358,918 (1.3\%)                     & 1,198,211 (0.12\%)                  & 1,174,939 (0.10\%) \\
\bottomrule
\end{tabular}
\label{tab:masked_mechanism_weights}
\end{sc}
\end{scriptsize}
\end{center}
\vskip -0.1in
\end{table}

This supports our argument that \OT methods target high logit-diff extraction mechanisms, rather than the fact lookup mechanisms that enrich the latent stream with the correct attributes, which decreases the robustness of edits/unlearning. It is important to note that since our \LT localization is based on our discovered mechanisms, this does not serve as an evaluation of \LT (since by definition \LT localization will only localize the \LT mechanism), but rather only of causal tracing and attribution patching.

\subsection{Hyperparameters}
\label{app:hyperparams}

Across all tasks except \CounterFactsequentialedit and all models, we fine tune using 50 iterations of batch size 4 with 16 accumulation steps, using an AdamW optimizer ~\citep{kingma2017adam} with 0 weight decay and a cosine annealing scheduler. For Gemma-2-9b, we are forced to use an 8-bit optimizer to fit our training in the memory of 1 GPU. We find that the optimal learning rate is quite sensitive to the localization used and the edit task, so we first sweep over learning rates to find reasonable learning rates. We sweep over the learning rates of 2e-6, 5e-6, 1e-5, 2e-5, 5e-5, and 1e-4, training models over 50 iterations with all $\lambda$s set to 1. We also tune the $\lambda_1$ parameter associated with the $L_{\text{injection}}$ cross entropy loss. We don't tune the other $\lambda$ parameters because they are all maintenance losses, and setting them to 1 works sufficiently to maintain performance across almost all setups. 

For the \CounterFactsequentialedit task, we use the same hyperparameters from the \CounterFactedit task and we train for 100 total iterations rather than 50, using 25 for each subset of 16 facts. We choose 25 iterations because it balances between being half the number of iterations we typically use per subset of that size, and also double the number of steps overall as we use in \CounterFactedit.

To avoid leaking evaluation information through this sweep process, we optimize learning rate for the objective of $(1 - \textit{Forget Set Ground Truth Accuracy}) + \textit{Forget Set Edit Accuracy} + \textit{Maintain Set Ground Truth Accuracy} + \textit{Pile Accuracy}$, avoiding any of our robustness metrics (one could view this sweep process as simply another part of the training process, since we only use train-time information). We run sweeps for Causal Tracing, Manual Fact Lookup, and No Localization. We then use the hyperparameters from Causal Tracing for Attribution Patching and Random (which all localize to MLPs and attention components), we use the hyperparameters from Fact Lookup for Random MLPs, Causal Tracing MLPs, and Attribution Patching MLPs (all localize to the same number of MLPs), and we use the hyperparameters from No Localization for All MLPs (which have the largest number of active parameters).

\subsubsection{Sample Hyperparameter Sweep}
In \cref{tab:gemma_lr_sweep_16_cfact}, we show the full results of one sweep, optimizing learning rate for editing 16 facts from CounterFact. Especially for No Localization, some learning rates fail to edit in the correct answer with high accuracy, or fail to maintain accuracy on the maintain set. In \cref{tab:gemma_fc_sweep_16_cfact}, we see that editing results are not particularly sensitive to the coefficient used with the injection cross entropy loss.
\begin{table}[h]
\scriptsize
\centering
\caption{Gemma-7b learning rate sweep, editing 16 CounterFact facts.}
\begin{tabular}{lccccc}
\toprule
& Pile Accuracy $\uparrow$ & Forget Accuracy $\downarrow$ & Edit Accuracy $\uparrow$ & Maintained Accuracy $\uparrow$ & Overall Score $\uparrow$ \\
\midrule
\multicolumn{6}{l}{\textit{\LT}} \\
\midrule
LR 0.0001 & 0.488 & 0.000 & 1.000 & 0.698 & 3.186 \\
LR 1e-05 & 0.513 & 0.000 & 1.000 & 0.975 & 3.488 \\
LR 2e-05 & 0.542 & 0.000 & 1.000 & 0.950 & 3.492 \\
LR 2e-06 & 0.499 & 0.007 & 0.961 & 0.869 & 3.323 \\
LR 5e-05 & 0.520 & 0.000 & 1.000 & 0.822 & 3.342 \\
LR 5e-06 & 0.528 & 0.000 & 0.999 & 0.980 & \textbf{3.507} \\
\midrule
\multicolumn{6}{l}{\textit{Localized CT}} \\
\midrule
LR 0.0001 & 0.462 & 0.001 & 0.999 & 0.697 & 3.157 \\
LR 1e-05 & 0.513 & 0.000 & 1.000 & 0.984 & 3.496 \\
LR 2e-05 & 0.507 & 0.000 & 1.000 & 0.961 & 3.467 \\
LR 2e-06 & 0.540 & 0.036 & 0.921 & 0.849 & 3.274 \\
LR 5e-05 & 0.488 & 0.000 & 1.000 & 0.846 & 3.334 \\
LR 5e-06 & 0.537 & 0.000 & 1.000 & 0.982 & \textbf{3.519} \\
\midrule
\multicolumn{6}{l}{\textit{Nonlocalized}} \\
\midrule
LR 0.0001 & 0.062 & 0.032 & 0.557 & 0.094 & 1.680 \\
LR 1e-05 & 0.520 & 0.000 & 1.000 & 0.892 & 3.412 \\
LR 2e-05 & 0.479 & 0.001 & 0.998 & 0.807 & 3.284 \\
LR 2e-06 & 0.529 & 0.000 & 1.000 & 0.982 & 3.510 \\
LR 5e-05 & 0.046 & 0.034 & 0.710 & 0.092 & 1.815 \\
LR 5e-06 & 0.536 & 0.000 & 1.000 & 0.988 & \textbf{3.524} \\
\bottomrule
\end{tabular} \label{tab:gemma_lr_sweep_16_cfact}
\label{tab:performance_metrics}
\end{table}

\begin{table}[h]
\scriptsize
\centering
\caption{Gemma-7b inject loss coefficient sweep, editing 16 CounterFact facts.}
\begin{tabular}{lccccc}
\toprule
& Pile Accuracy $\uparrow$ & Forget Accuracy $\downarrow$ & Edit Accuracy $\uparrow$ & Maintained Accuracy $\uparrow$ & Overall Score $\uparrow$ \\
\midrule
\multicolumn{6}{l}{\textit{\LT}} \\
\midrule
FC 0.1 & 0.538 & 0.000 & 0.998 & 0.990 & \textbf{3.525} \\
FC 0.2 & 0.510 & 0.000 & 0.999 & 0.981 & 3.491 \\
FC 0.5 & 0.508 & 0.000 & 0.999 & 0.985 & 3.493 \\
FC 1 & 0.510 & 0.000 & 0.999 & 0.973 & 3.483 \\
FC 2 & 0.532 & 0.000 & 1.000 & 0.984 & 3.515 \\
FC 5 & 0.534 & 0.000 & 1.000 & 0.985 & 3.518 \\
\midrule
\multicolumn{6}{l}{\textit{Localized CT}} \\
\midrule
FC 0.1 & 0.532 & 0.004 & 0.995 & 0.983 & 3.506 \\
FC 0.2 & 0.535 & 0.001 & 0.998 & 0.986 & \textbf{3.518} \\
FC 0.5 & 0.524 & 0.000 & 0.999 & 0.974 & 3.497 \\
FC 1 & 0.519 & 0.000 & 1.000 & 0.988 & 3.507 \\
FC 2 & 0.536 & 0.000 & 1.000 & 0.979 & 3.514 \\
FC 5 & 0.518 & 0.000 & 1.000 & 0.975 & 3.493 \\
\midrule
\multicolumn{6}{l}{\textit{Nonlocalized}} \\
\midrule
FC 0.1 & 0.525 & 0.001 & 0.998 & 0.969 & 3.492 \\
FC 0.2 & 0.562 & 0.000 & 0.999 & 0.982 & \textbf{3.543} \\
FC 0.5 & 0.524 & 0.000 & 1.000 & 0.962 & 3.486 \\
FC 1 & 0.531 & 0.000 & 1.000 & 0.980 & 3.511 \\
FC 2 & 0.514 & 0.000 & 1.000 & 0.978 & 3.492 \\
FC 5 & 0.528 & 0.000 & 1.000 & 0.975 & 3.503 \\
\bottomrule
\end{tabular} \label{tab:gemma_fc_sweep_16_cfact}
\label{tab:performance_metrics_fc}
\end{table}

\subsubsection{All Hyperparameters Used}
\Cref{tab:optimal_lrs} has all learning rates used and \cref{tab:optimal_fc_coefs} has all injection loss coefficients used.

\begin{table}[h!]
\scriptsize
\centering
\caption{Optimal learning rates for different models, task types, and localizations.}
\begin{tabular}{@{}lllll@{}}
\toprule
\textbf{Model} & \textbf{64 athletes to random sport} & \textbf{Basketball Athletes to Golf} & \textbf{16 CounterFact facts} & \textbf{64 CounterFact facts} \\ \midrule
\multicolumn{5}{l}{\textbf{Gemma}} \\ \midrule
\LT   & $1 \times 10^{-5}$   & $2 \times 10^{-5}$   & $5 \times 10^{-6}$   & $1 \times 10^{-5}$ \\
Localized CT    & $1 \times 10^{-5}$   & $5 \times 10^{-6}$   & $5 \times 10^{-6}$   & $2 \times 10^{-5}$ \\
Nonlocalized    & $2 \times 10^{-6}$   & $5 \times 10^{-6}$   & $5 \times 10^{-6}$   & $2 \times 10^{-6}$ \\ \midrule
\multicolumn{5}{l}{\textbf{Gemma 2}} \\ \midrule
\LT   & $1 \times 10^{-5}$   & $5 \times 10^{-5}$   & $5 \times 10^{-5}$   & $2 \times 10^{-5}$ \\
Localized CT    & $5 \times 10^{-5}$   & $5 \times 10^{-5}$   & $5 \times 10^{-5}$   & $2 \times 10^{-5}$ \\
Nonlocalized    & $2 \times 10^{-5}$   & $1 \times 10^{-5}$   & $5 \times 10^{-6}$   & $5 \times 10^{-6}$ \\ \midrule
\multicolumn{5}{l}{\textbf{Llama 3}} \\ \midrule
\LT   & $5 \times 10^{-5}$   & $5 \times 10^{-5}$   & $1 \times 10^{-4}$   & $5 \times 10^{-5}$ \\
Localized CT    & $1 \times 10^{-4}$   & $5 \times 10^{-5}$   & $2 \times 10^{-5}$   & $5 \times 10^{-5}$ \\
Nonlocalized    & $2 \times 10^{-5}$   & $1 \times 10^{-5}$   & $2 \times 10^{-5}$   & $1 \times 10^{-5}$ \\ \bottomrule
\end{tabular} \label{tab:optimal_lrs}
\label{tab:learning_rates}
\end{table}

\begin{table}[h!]
\scriptsize
\centering
\caption{Optimal inject loss coefficients for different models, task types, and localizations.}
\begin{tabular}{@{}lllll@{}}
\toprule
\textbf{Model} & \textbf{64 athletes to random sport} & \textbf{Basketball Athletes to Golf} & \textbf{16 CounterFact facts} & \textbf{64 CounterFact facts} \\ \midrule
\multicolumn{5}{l}{\textbf{Gemma}} \\ \midrule
\LT   & 5.0   & 5.0   & 0.1   & 2.0 \\
Localized CT    & 0.1   & 1.0   & 0.2   & 1.0 \\
Nonlocalized    & 0.2   & 1.0   & 0.2   & 1.0 \\ \midrule
\multicolumn{5}{l}{\textbf{Gemma 2}} \\ \midrule
\LT   & 1.0   & 5.0   & 2.0   & 0.5 \\
Localized CT    & 5.0   & 5.0   & 2.0   & 2.0 \\
Nonlocalized    & 5.0   & 5.0   & 1.0   & 1.0 \\ \midrule
\multicolumn{5}{l}{\textbf{Llama 3}} \\ \midrule
\LT   & 5.0   & 0.2   & 1.0   & 0.2 \\
Localized CT    & 2.0   & 1.0   & 0.1   & 2.0 \\
Nonlocalized    & 2.0   & 0.1   & 2.0   & 0.2 \\ \bottomrule
\end{tabular} \label{tab:optimal_fc_coefs}
\label{tab:forget_loss_coef}
\end{table}

\subsection{Evaluation Details}

\subsubsection{Details on Standard Prompt Evaluations}
\label{app:standard_prompting}
\begin{figure}[h!] \centering
    \includegraphics[width=.98\linewidth]{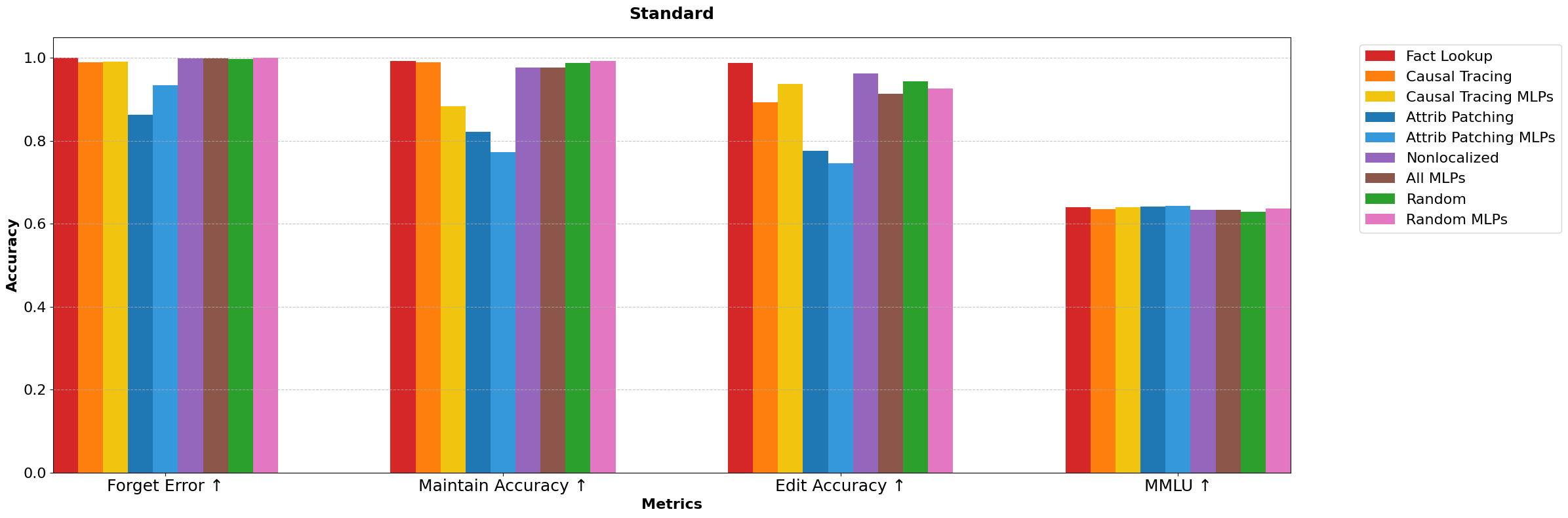}
    \caption{Standard Prompting results for \Sportathleteedit, across all localizations.}
    \label{fig:app-standard-prompting-sportsathleteedit}
\end{figure}

\begin{figure}[h!] \centering
    \includegraphics[width=.98\linewidth]{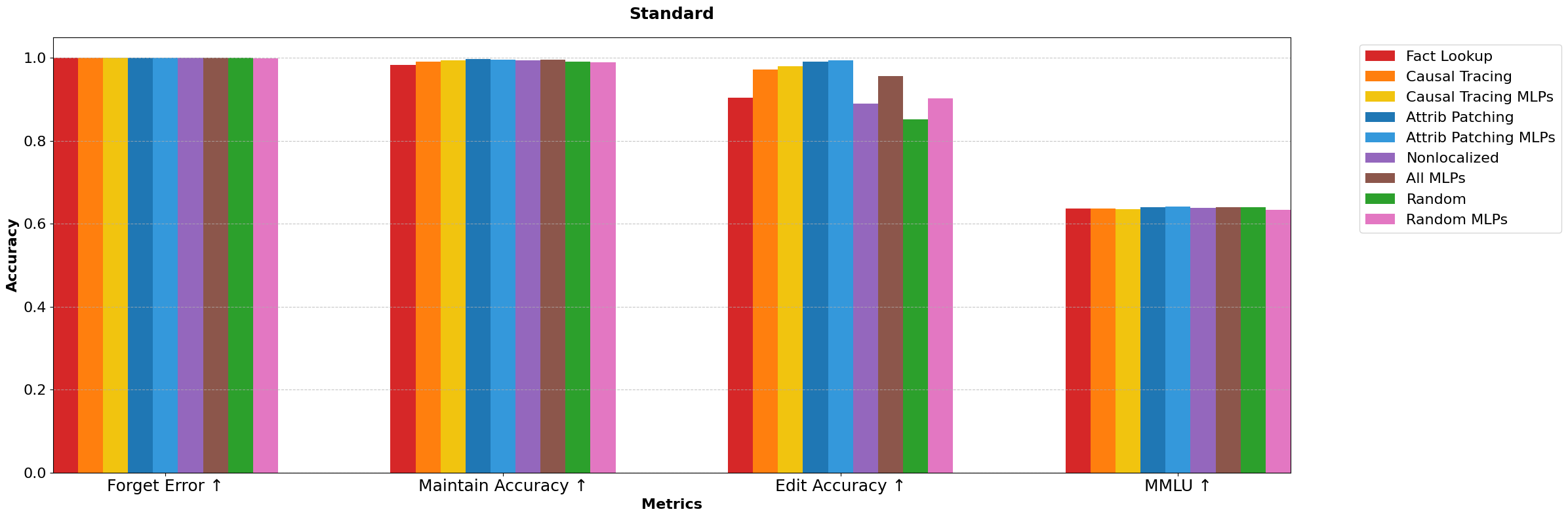}
    \caption{Standard Prompting results for \Sportedit, across all localizations.}
    \label{fig:app-standard-prompting-sportsedit}
\end{figure}

\begin{figure}[h!] \centering
    \includegraphics[width=.98\linewidth]{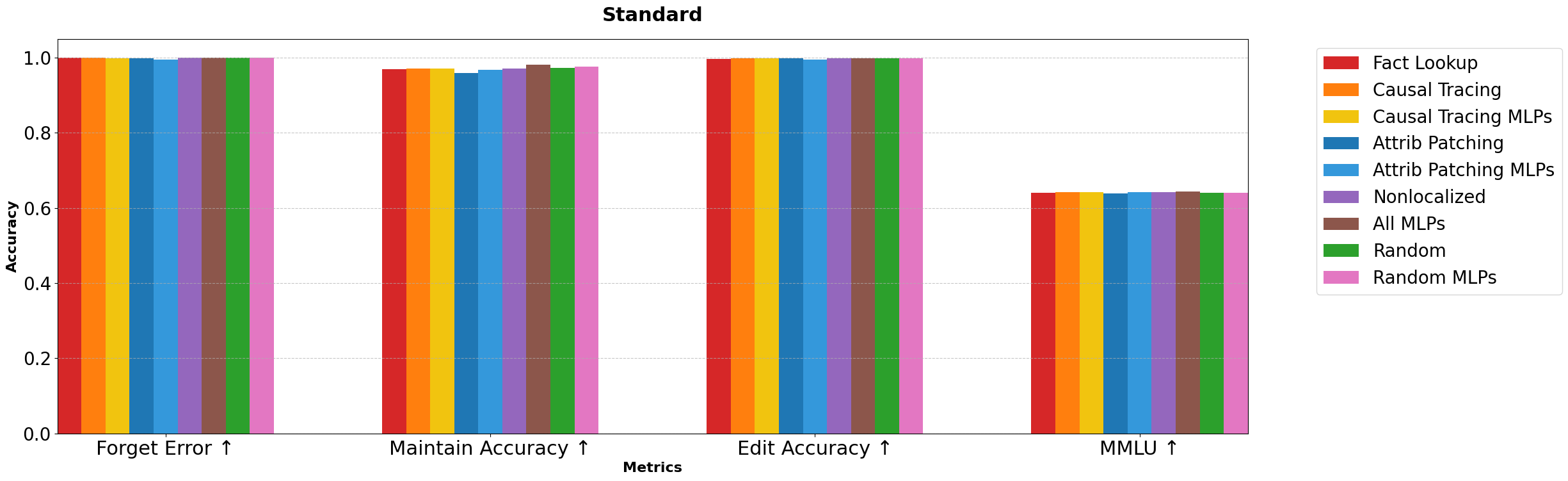}
    \caption{Standard Prompting results for \CounterFactedit, across all localizations.}
    \label{fig:app-standard-prompting-counterfactedit}
\end{figure}

\begin{figure}[h!] \centering
    \includegraphics[width=.98\linewidth]{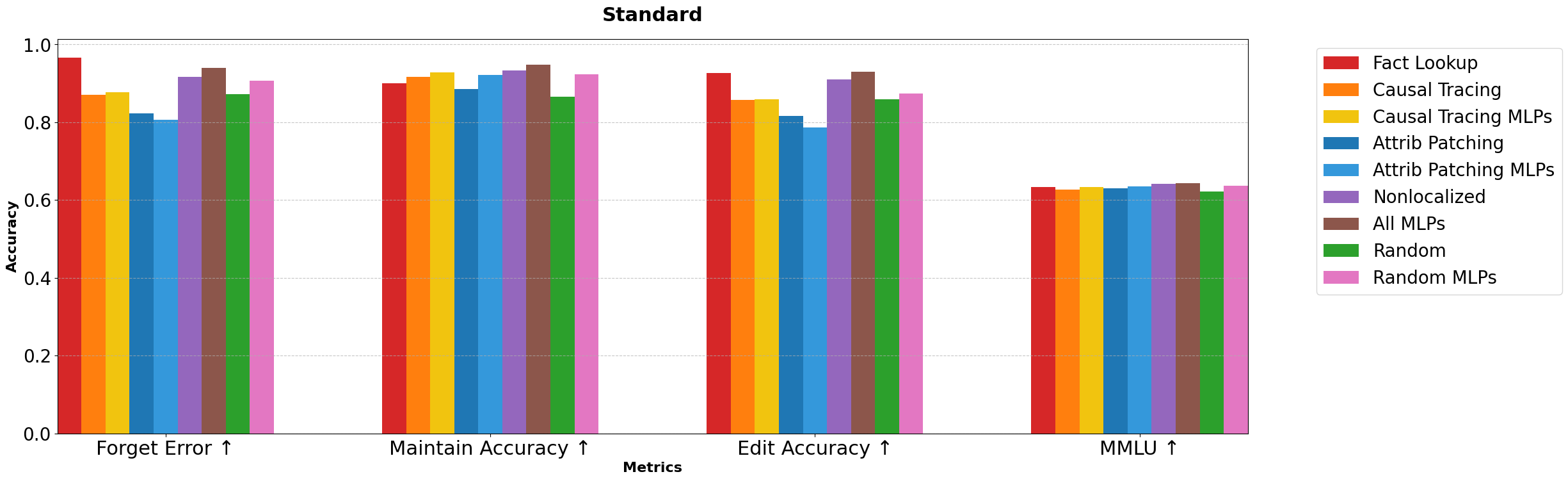}
    \caption{Standard Prompting results for \CounterFactsequentialedit, across all localizations.}
    \label{fig:app-standard-prompting-counterfactsequentialedit}
\end{figure}

We report standard metrics of Forget Error, Edit Accuracy, and Maintain Accuracy, in the same prompt format that the models were trained on. These metrics are optimized by the loss, so we expect all localizations to do almost perfectly on these evaluations. \Cref{fig:app-standard-prompting-sportsathleteedit,fig:app-standard-prompting-sportsedit,fig:app-standard-prompting-counterfactedit,fig:app-standard-prompting-counterfactsequentialedit} show that localizations perform approximately equivalently on these basic evaluations across all tasks. 

\subsubsection{Details on Adversarial Prompt Evaluations}
\label{app:adv_prompting}
\begin{figure}[h!] \centering
    \includegraphics[width=.98\linewidth]{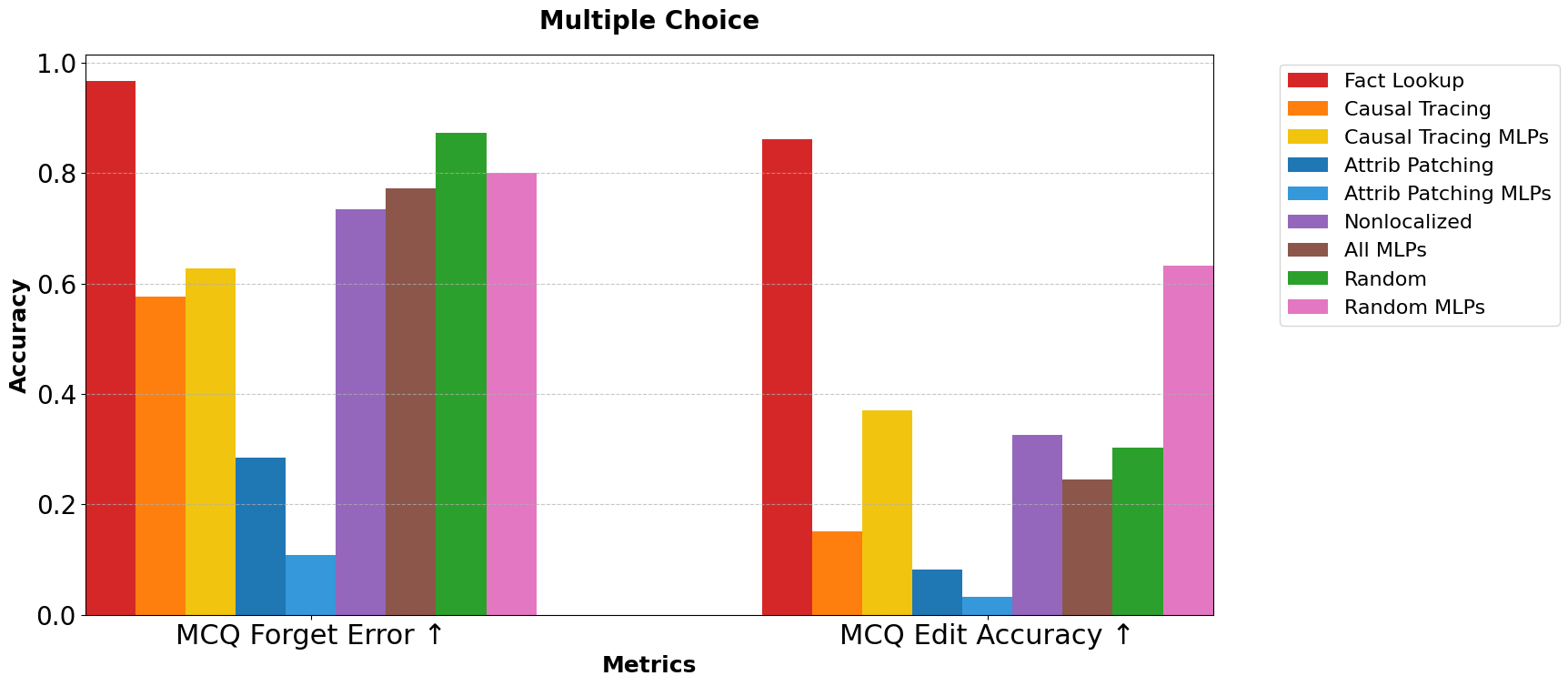}
    \caption{Adversarial Prompting results for \Sportathleteedit, across all localizations.}
    \label{fig:app-adversarial-prompting-sportsathleteedit}
\end{figure}

\begin{figure}[h!] \centering
    \includegraphics[width=.98\linewidth]{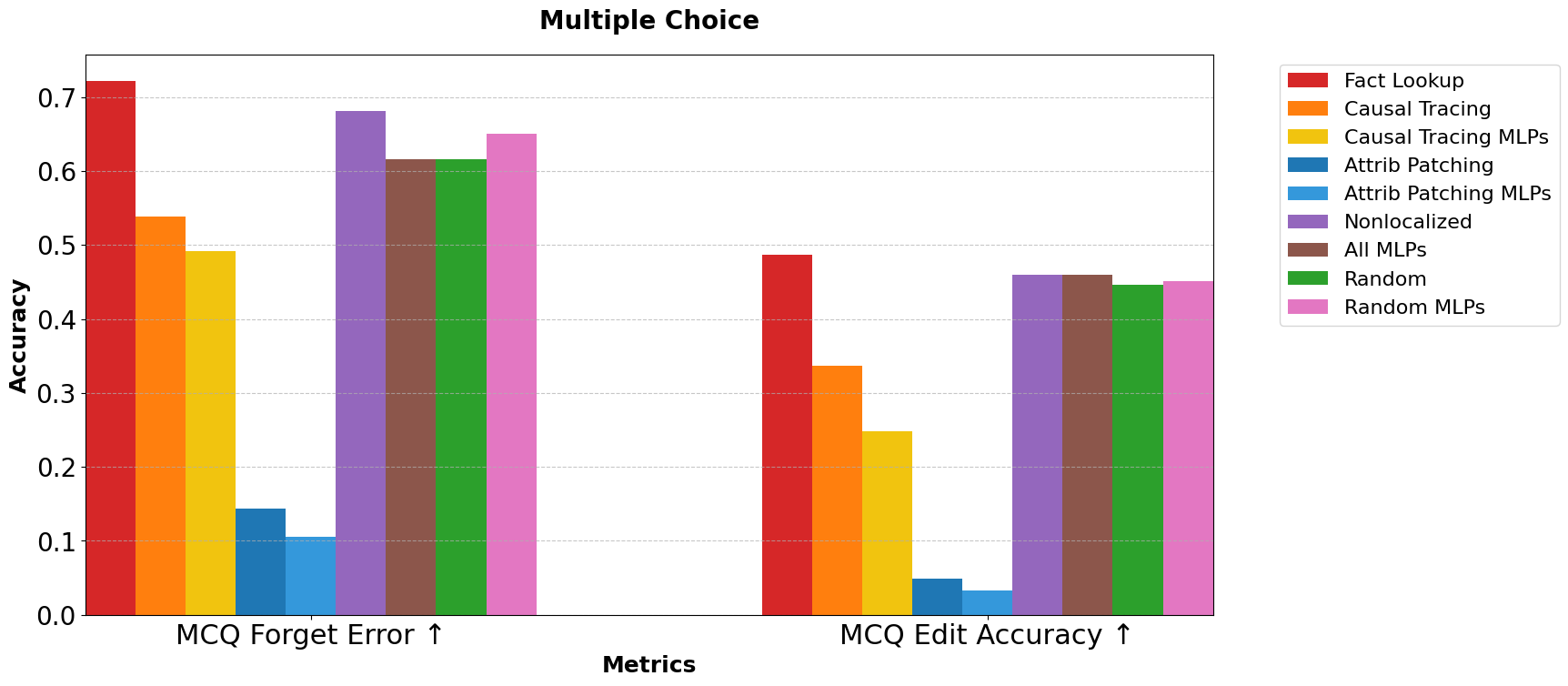}
    \caption{Adversarial Prompting results for \Sportedit, across all localizations.}
    \label{fig:app-adversarial-prompting-sportsedit}
\end{figure}

\begin{figure}[h!] \centering
    \includegraphics[width=.98\linewidth]{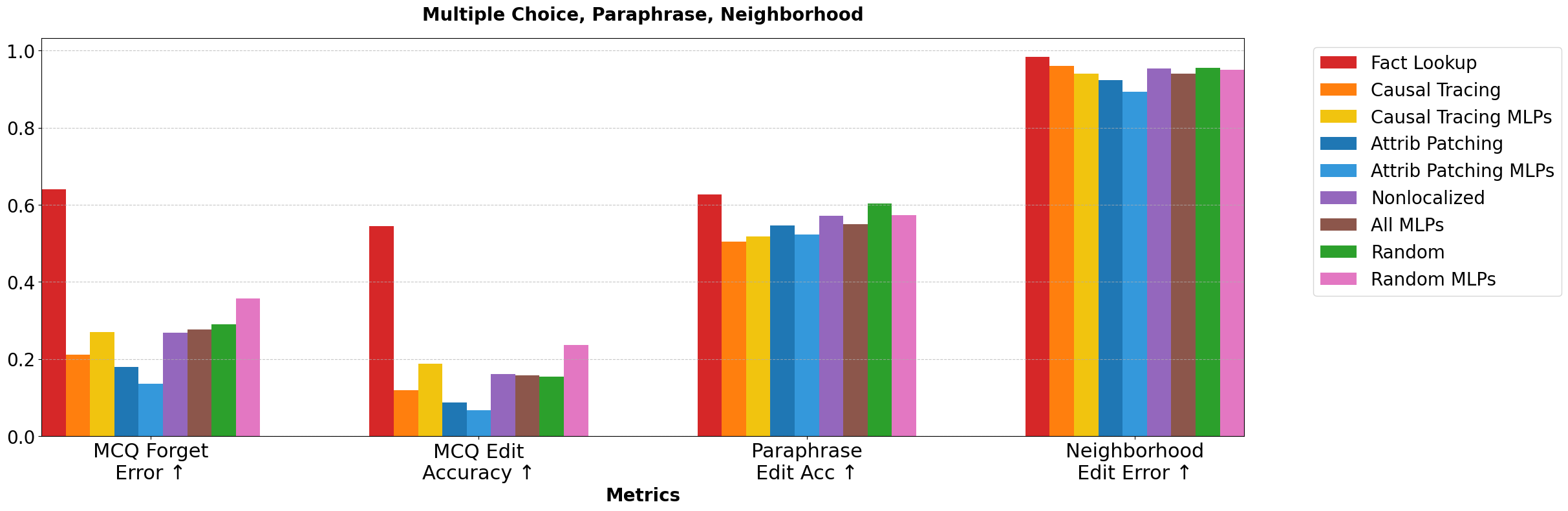}
    \caption{Adversarial Prompting results for \CounterFactedit, across all localizations.}
    \label{fig:app-adversarial-prompting-counterfactedit}
\end{figure}

\begin{figure}[h!] \centering
    \includegraphics[width=.98\linewidth]{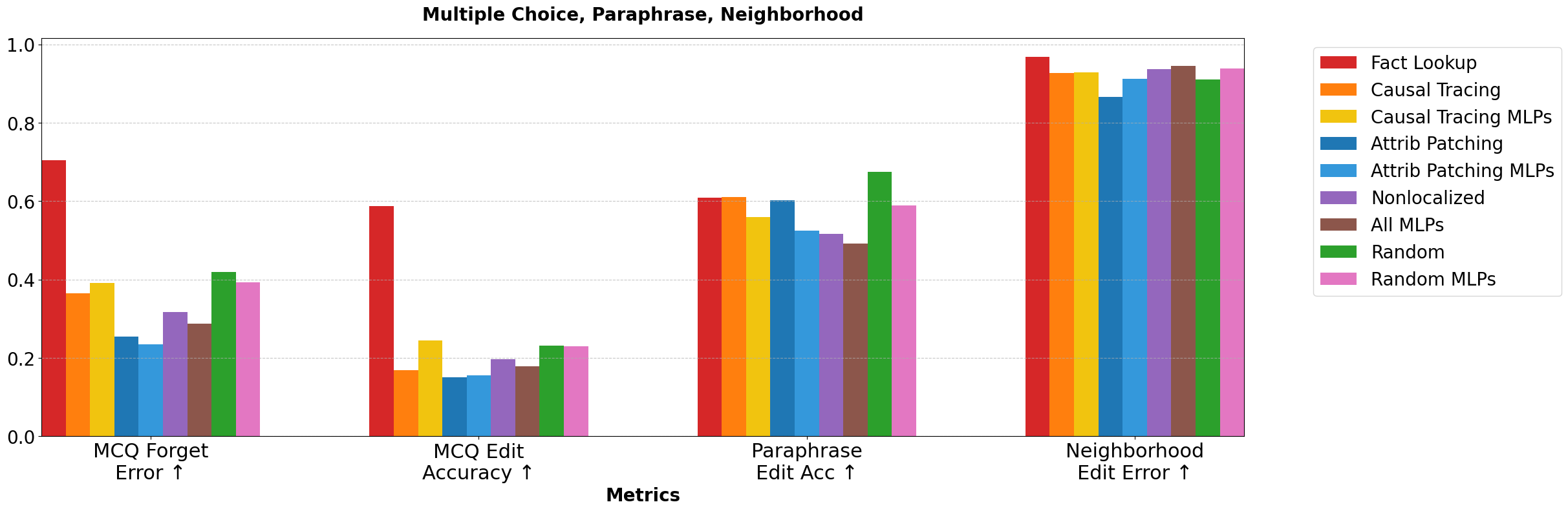}
    \caption{Adversarial Prompting results for \CounterFactsequentialedit, across all localizations.}
    \label{fig:app-adversarial-prompting-counterfactsequentialedit}
\end{figure}

We report the adversarial prompt evaluations from \cref{sec:prompting} across all localizations. \Cref{fig:app-adversarial-prompting-sportsathleteedit,fig:app-adversarial-prompting-sportsedit,fig:app-adversarial-prompting-counterfactedit,fig:app-adversarial-prompting-counterfactsequentialedit} all show that \LT localization is more robust in MCQ compared to every other localization (significantly stronger for CounterFact). \Cref{fig:app-adversarial-prompting-counterfactedit,fig:app-adversarial-prompting-counterfactsequentialedit} show that \LT localization is optimal in Paraphrase and Neighborhood in all cases except for Paraphrase compared to the Random localization in \CounterFactsequentialedit.

\subsubsection{Details on Adversarial Relearning}
\label{app:advrelearn}
We retrain the model for 20 iterations with cross-entropy on half of the forget set (along with a standard retain and SFT loss), adding up all losses with loss coefficient 1. 

\begin{figure}[h!] \centering
    \includegraphics[width=.98\linewidth]{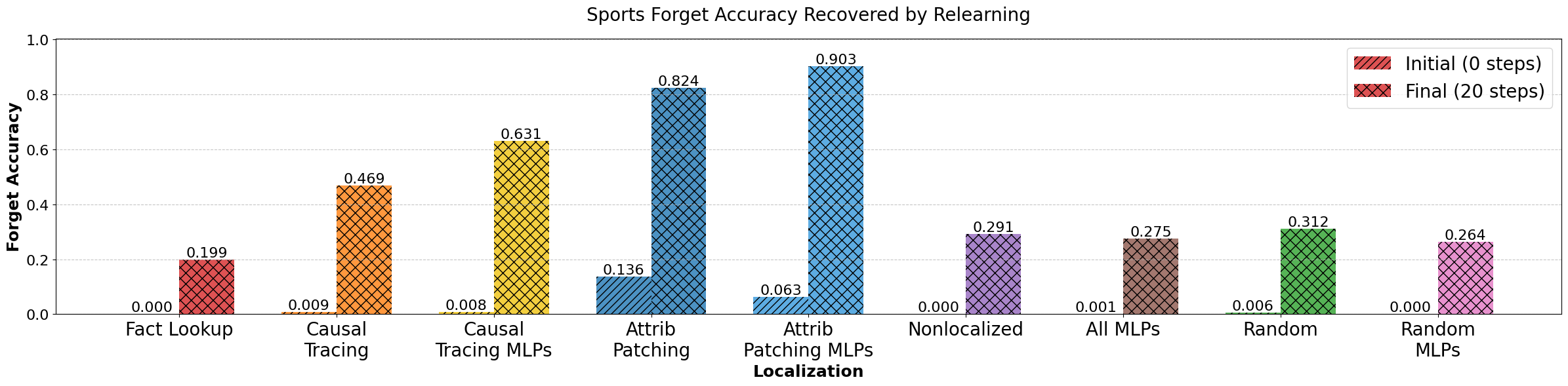}
    \caption{Relearning results for \Sportathleteedit, across all localizations.}
    \label{fig:app-relearning-sportsathleteedit}
\end{figure}

We present relearning results for all localizations averaged over models. As shown in \cref{fig:app-relearning-sportsathleteedit}, the \LT localization remains optimal, although the baselines of Nonlocalized, All MLPs, Random, and Random MLPs are competitive.

\begin{figure}[h!] \centering
    \includegraphics[width=.98\linewidth]{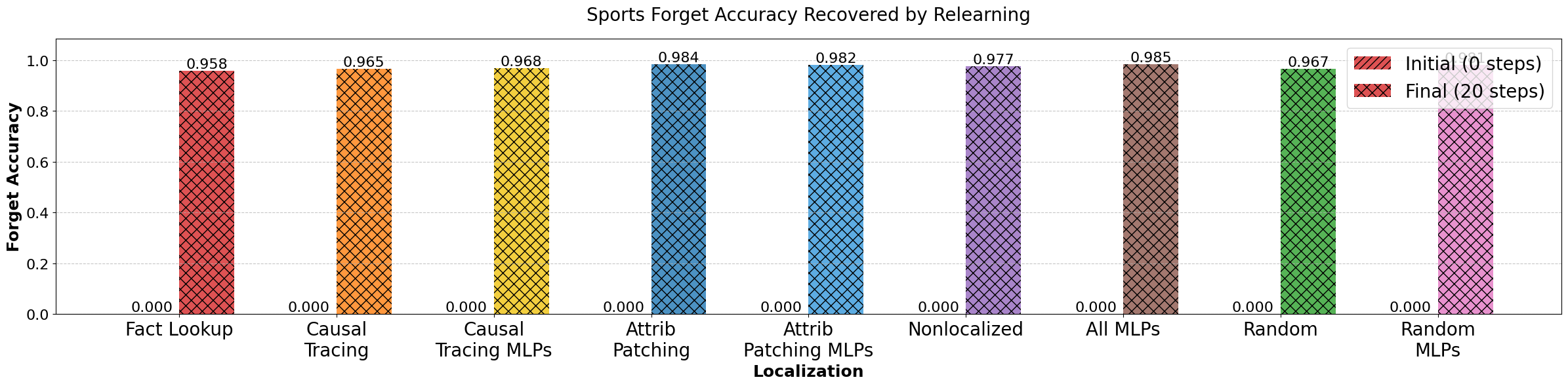}
    \caption{Relearning results for \Sportedit.}
    \label{fig:app-relearning-sportsedit}
\end{figure}

We also present relearning results on the other tasks. As mentioned in \cref{sec:adversarial_relearning}, since \Sportedit forget facts are not independent, we don't expect valid results from relearning. Thus, in \cref{fig:app-relearning-sportsedit}, we see that every localization regains 100\% editing accuracy.

\begin{figure}[h!] \centering
    \includegraphics[width=.98\linewidth]{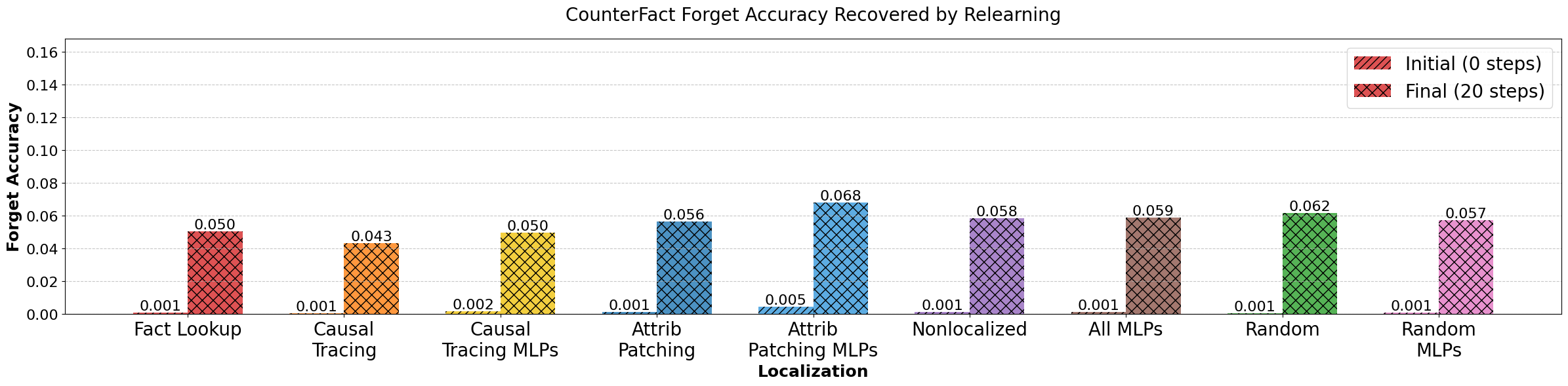}
    \caption{Relearning results for \CounterFactedit.}
    \label{fig:app-relearning-counterfactedit}
\end{figure}

\begin{figure}[h!] \centering
    \includegraphics[width=.98\linewidth]{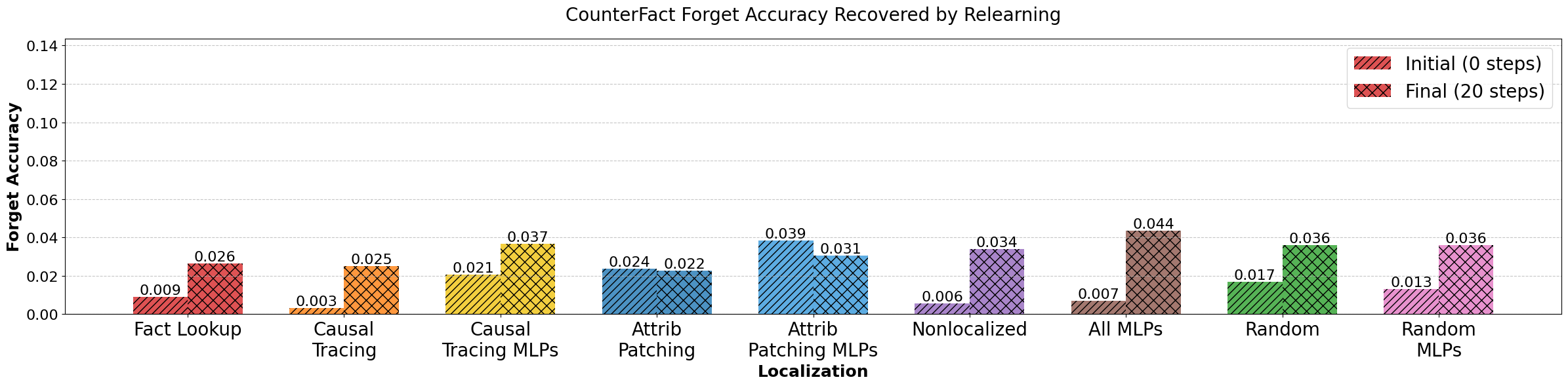}
    \caption{Relearning results for \CounterFactsequentialedit.}
    \label{fig:app-relearning-counterfactsequentialedit}
\end{figure}

On \CounterFactedit and \CounterFactsequentialedit, as shown in \cref{fig:app-relearning-counterfactedit,fig:app-relearning-counterfactsequentialedit}, none of the localizations relearn more than 7\% accuracy, suggesting adversarial relearning was not a sufficiently strong enough evaluation for these tasks. Regardless, \LT localization is either the most or second-most robust localization to relearning, although localizations don't differ by much.

\subsubsection{Details on Latent Knowledge}
\label{app:latentknowledge}
We present the probing classification accuracies for the three models separately here, as well as for all localizations we previously left out.
\begin{figure}[h!] \centering
    \includegraphics[width=.98\linewidth]{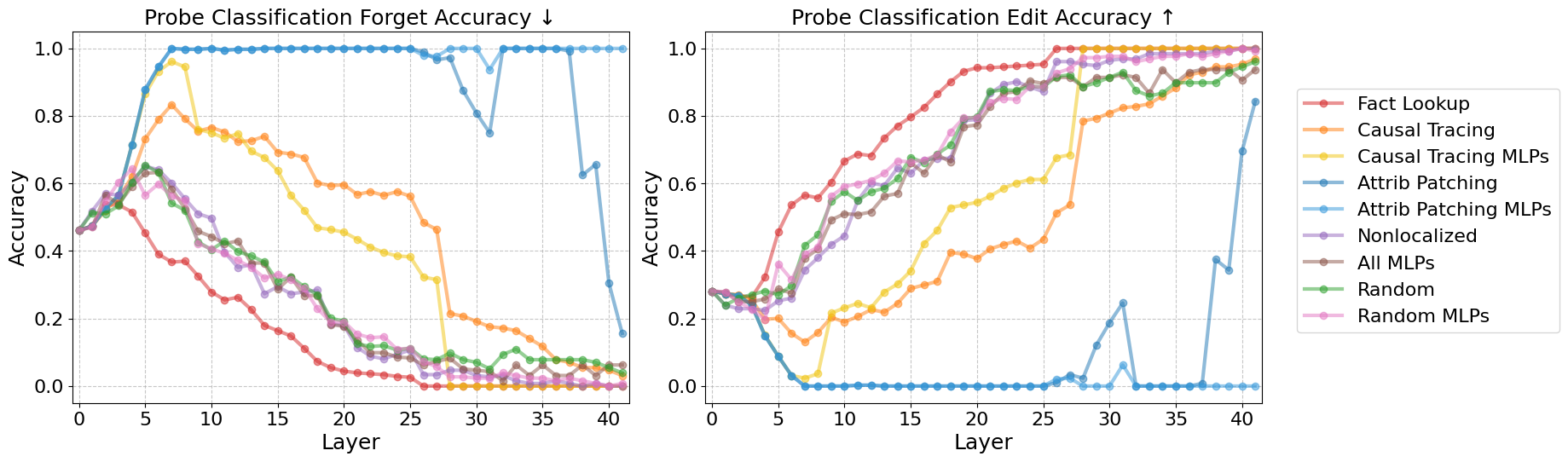}
    \caption{Linear probes applied to the forget set across all models, classifying model activations after various layers.}
    \label{fig:app-ft-dlk-athletes-injection-all}
\end{figure}

\begin{figure}[h!] \centering
    \includegraphics[width=.98\linewidth]{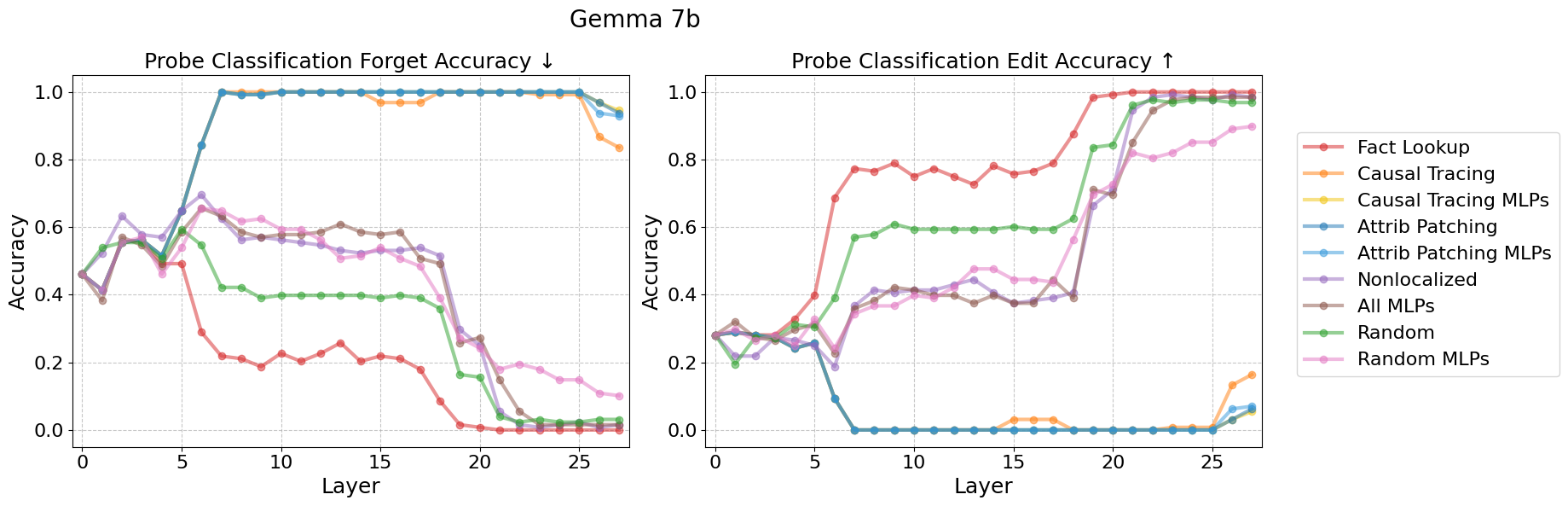}
    \caption{Linear probes applied to the forget set on Gemma-7B with 28 layers. }
    \label{fig:app-ft-dlk-athletes-injection-gemma}
\end{figure}

\begin{figure}[h!] \centering
    \includegraphics[width=.98\linewidth]{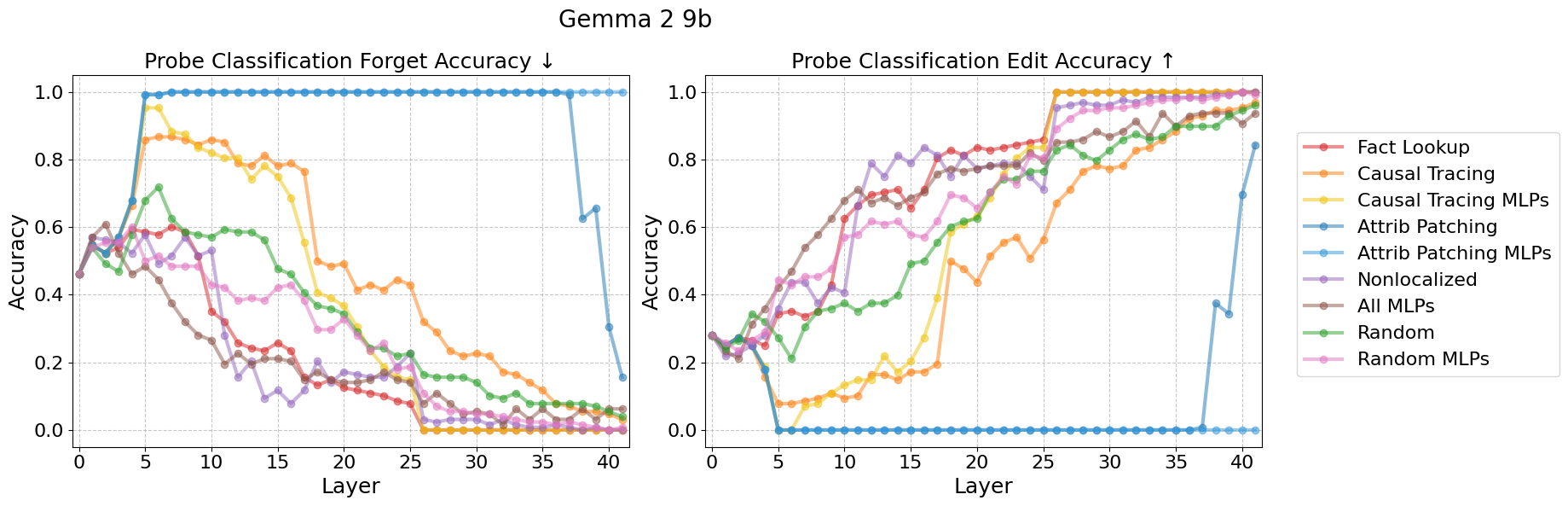}
    \caption{Linear probes applied to the forget set on Gemma-2-9b with 42 layers. }
    \label{fig:app-ft-dlk-athletes-injection-gemma2}
\end{figure}

\begin{figure}[h!] \centering
    \includegraphics[width=.98\linewidth]{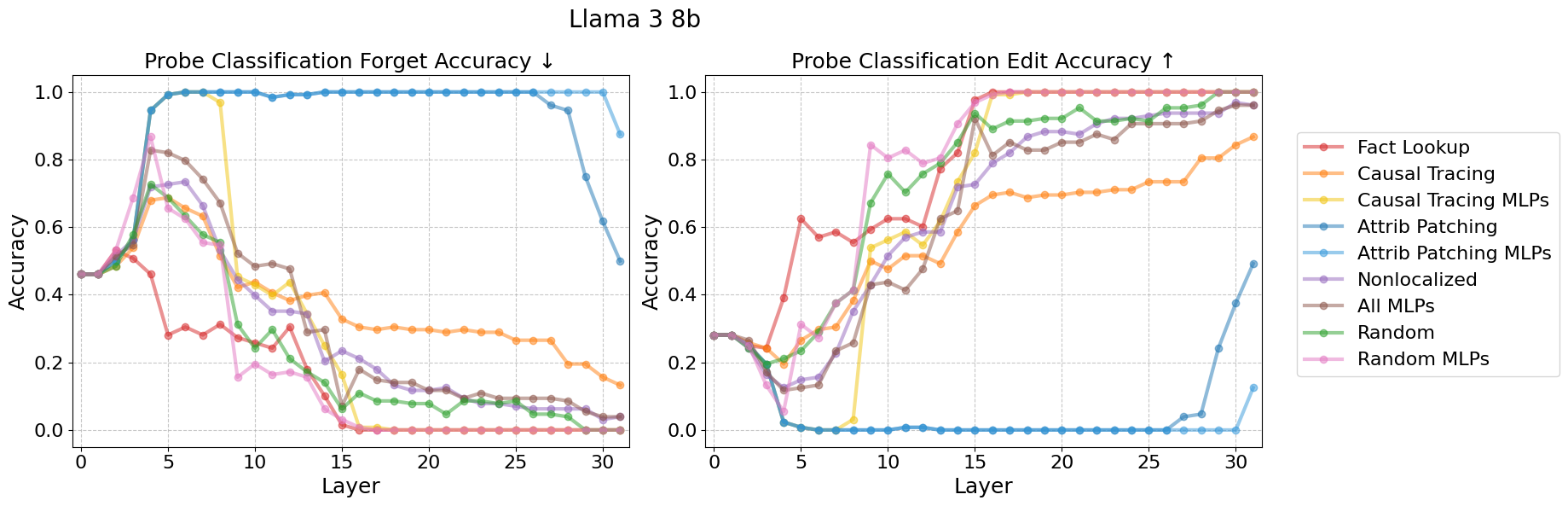}
    \caption{Linear probes applied to the forget set on Llama-3-8b with 32 layers. }
    \label{fig:app-ft-dlk-athletes-injection-llama3}
\end{figure}

In Gemma-7b and Llama-3-8b, \LT probing is the most monotonic and the best in the early layers, either steadily decreasing to 0 for forget accuracy or increasing to 1 for edit accuracy, with the least extreme peaks. In Gemma-2-9b, the Nonlocalized, All MLPs, and Random MLPs baselines are competitive with \LT. The other \OT localization, Attribution Patching, has 100\% probing forget accuracy across many layers, suggesting it represents the ground truth answer very clearly.

\subsection{Soft Prompt Evaluations}
\label{app:soft_prompts}
Because many localizations seem to be weak to prompting schemes, we attempt a simple adaptive attack of soft prompts, where we optimize the continuous embeddings at the end of the prompt to recover the correct answer on half of our forget set.  We then evaluate the model’s performance on the other half, with this soft prompt in place \citep{lester2021powerscaleparameterefficientprompt}. We average over evaluations from four soft prompts. Soft prompt evaluations can be considered to be a more narrow form of few-shot finetuning, that is closer to searching for prompts that recover the model’s knowledge.

\begin{figure}[h!] \centering
    \includegraphics[width=.98\linewidth]{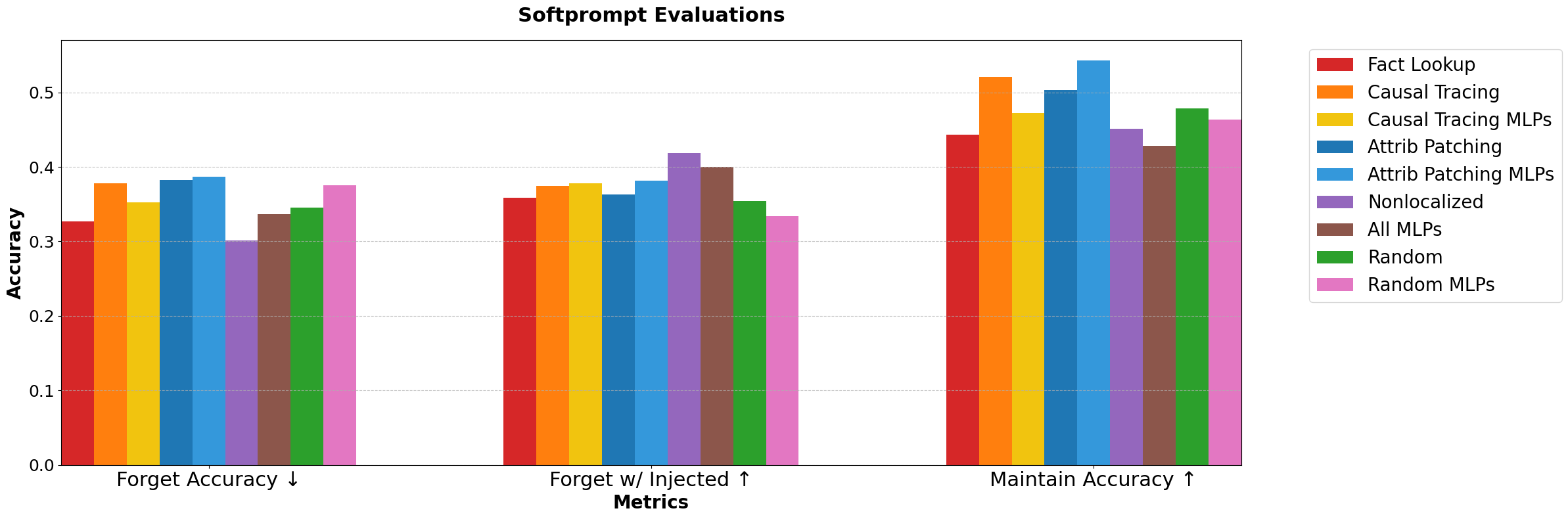}
    \caption{Metrics with soft prompts applied for \Sportathleteedit, averaged over all models.}
    \label{fig:app-softprompts-sportsathleteedit-all}
\end{figure}

\begin{figure}[h!] \centering
    \includegraphics[width=.98\linewidth]{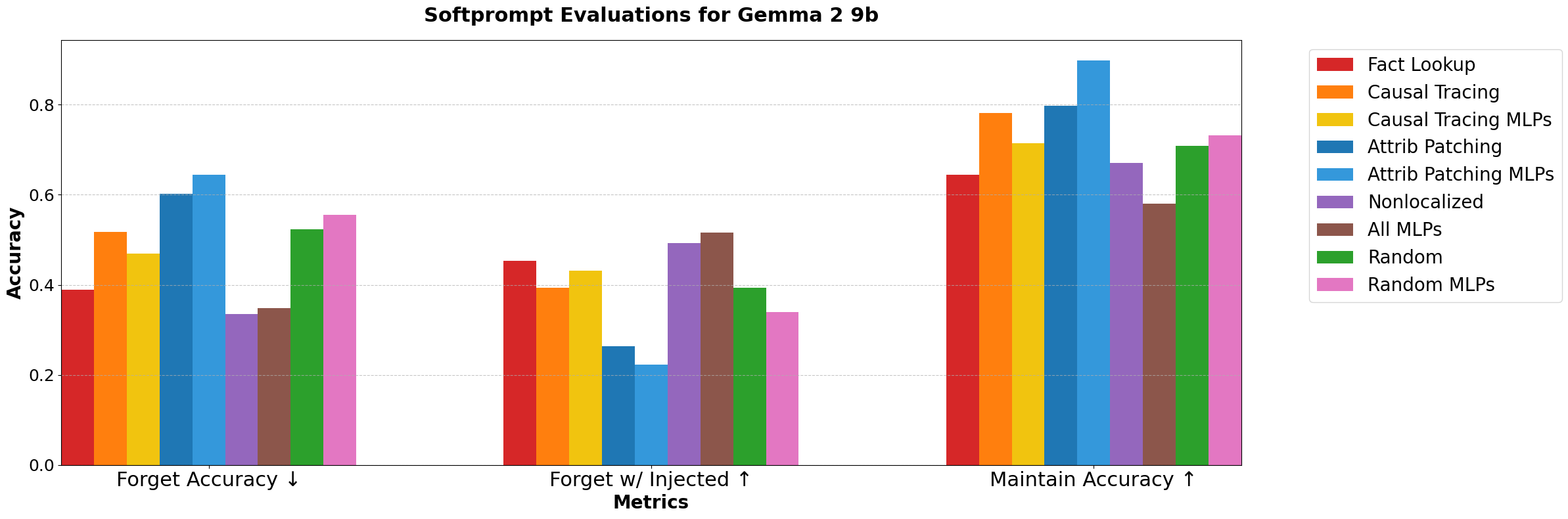}
    \caption{Metrics with soft prompts applied for \Sportathleteedit for Gemma-2-9b.}
    \label{fig:app-softprompts-sportsathleteedit-gemma2}
\end{figure}

\begin{figure}[h!] \centering
    \includegraphics[width=.98\linewidth]{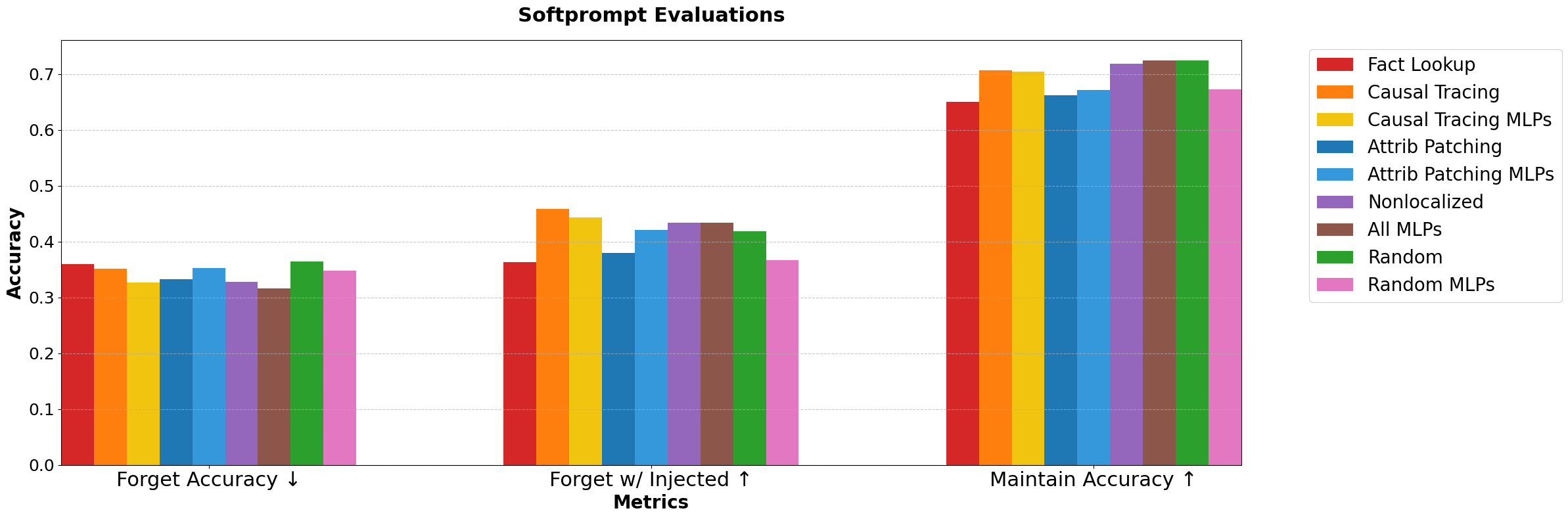}
    \caption{Metrics with soft prompts applied for \CounterFactedit, averaged over all models.}
    \label{fig:app-softprompts-counterfactedit-all}
\end{figure}

\begin{figure}[h!] \centering
    \includegraphics[width=.98\linewidth]{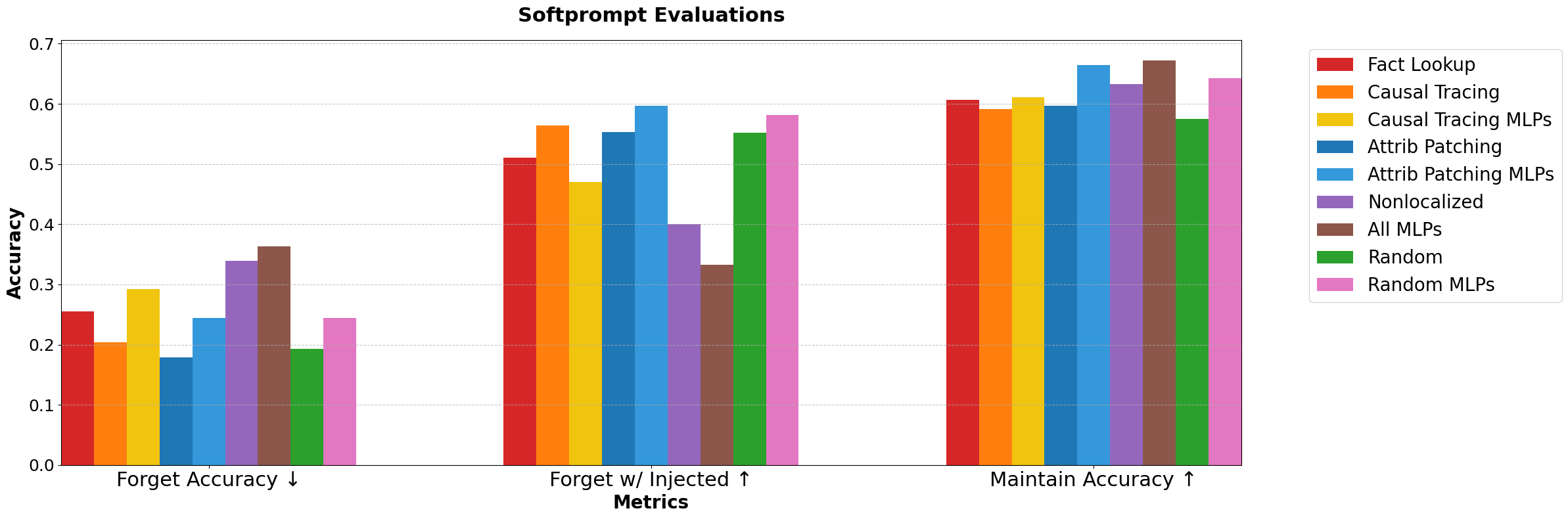}
    \caption{Metrics with soft prompts applied for \CounterFactsequentialedit, averaged over all models.}
    \label{fig:app-softprompts-counterfactsequentialedit-all}
\end{figure}

We find limited soft prompt success: across most tasks and models, we don't recover much held-out forget set accuracy. On the \Sportathleteedit, \CounterFactedit, and \CounterFactsequentialedit tasks, \cref{fig:app-softprompts-sportsathleteedit-all,fig:app-softprompts-counterfactedit-all,fig:app-softprompts-counterfactsequentialedit-all} show that all localizations don't significantly improve in Forget Accuracy over random chance, or are about equal between localizations, after soft prompts are applied. In \cref{fig:app-softprompts-sportsathleteedit-gemma2}, specifically for Gemma-2 on \Sportathleteedit we see some reasonable results with softprompts that are able to recover over 60\% Forget Accuracy on \OT localizations, while \LT, Nonlocalized, and All MLPs remain under 40\% Forget Accuracy.

\end{document}